%% file: main.tex
\documentclass[
	11pt,
	letterpaper,
	oneside,
]{book} 

\usepackage[
	font=footnotesize,
    labelfont=bf
]{caption}
\usepackage[main=english,spanish]{babel}
\usepackage[utf8]{inputenc}
\usepackage{textcomp}

\usepackage{amsmath}
\usepackage{mathtools}

\usepackage{float}
\usepackage{subfigure}

\usepackage[ruled]{algorithm2e}

\usepackage{color}
\usepackage{graphicx}
\usepackage{graphics}
\usepackage{soul}

\usepackage[draft=true]{hyperref} 

\usepackage{apacite}
\usepackage[sort&compress]{natbib}
\usepackage{bibentry}

\usepackage{url}
\usepackage{setspace} 
\usepackage{ragged2e} 

\usepackage{array} 
\newcolumntype{L}[1]{>{\raggedright\let\newline\\\arraybackslash\hspace{0pt}}m{#1}}
\newcolumntype{C}[1]{>{\centering\let\newline\\\arraybackslash\hspace{0pt}}m{#1}}
\newcolumntype{R}[1]{>{\raggedleft\let\newline\\\arraybackslash\hspace{0pt}}m{#1}}

\usepackage{booktabs}
\usepackage{pbox}
\usepackage{longtable}
\usepackage[paper=portrait,pagesize]{typearea}

\usepackage[final]{pdfpages}

\usepackage{fancyhdr}

\input{config.tex}   %

\usepackage[
    left=4.0cm,
    right=3.5cm,
    top=3.5cm,
    bottom=3.5cm
]{geometry}

\begin{document}

    \frontmatter
    	
    \input{template/titlepage.tex}

    \input{template/approval.tex} 


        \setcounter{page}{1}    

        \onehalfspacing
        
     \input{chapters/abstract.tex} 

        \singlespacing

        \pagestyle{fancy}
        \setlength\headheight{42pt} 
        \lhead{\includegraphics[width=1.3cm]{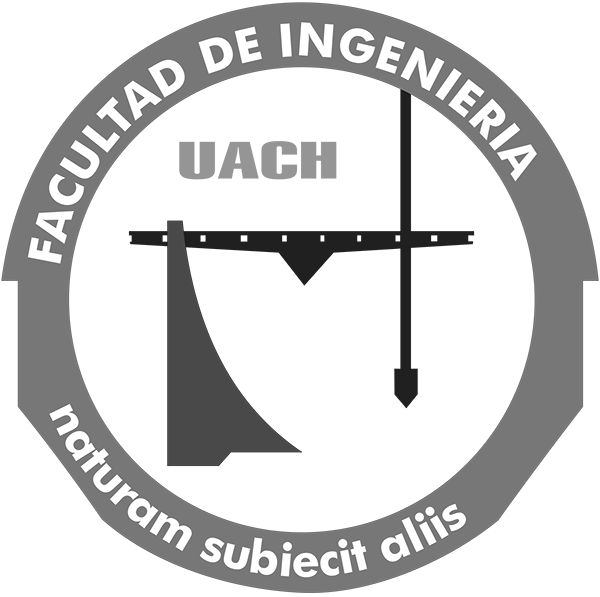}}   

        \iIndices

	\mainmatter
    	\doublespacing    
    
        
    \input{chapters/introduction.tex}
    \input{chapters/background.tex}
    \input{chapters/lit_rev.tex}

    \input{chapters/reorientation.tex}
    \input{chapters/potholes.tex}
    \input{chapters/profiling.tex}
    \input{chapters/risky.tex}
    \input{chapters/conclusions.tex}

    \backmatter
        \onehalfspacing
            
        \cleardoublepage
        \phantomsection
    	\addcontentsline{toc}{chapter}{References}        
        \renewcommand\bibname{References}

        \bibliographystyle{apacite}
        
	    \bibliography{references}

\end{document}

%% file: config.tex
\newcommand{\iIndices}{
     \tableofcontents
     \listoffigures
     \listoftables
}

\newcommand{\vTitle}{A Machine Learning Approach for Smartphone-based Sensing of Roads and Driving Style}
\newcommand{\vAuthor}{Manuel Ricardo Carlos Loya }
\newcommand{\vDegree}{Doctor of Engineering}
\newcommand{\vDate}{August 2019}
\newcommand{\vDissertationDirector}{Dr. Luis Carlos Gonz\'alez Gurrola}
\newcommand{\vCommittee}{\noindent
    Dr. Luis Carlos Gonz\'alez Gurrola\\
    Dr. Fernando Mart\'inez Reyes\\
    Dr. Graciela Mar\'ia de Jes\'us Ram\'irez Alonso\\
    Dr. Raymundo Cornejo García\\
    Dr. Manuel Montes y G\'omez
}


%% file: template/titlepage.tex
\newgeometry{
    top=2.5cm, 
    bottom=1cm, 
    left=3cm, 
    right=3cm
}

\thispagestyle{empty}

\begin{center}

    \LARGE{ \textbf{ \textsc{
        Universidad Autónoma de Chihuahua
    } } }    
    
    \vspace*{3mm}
    \Large{ \textbf{ \textsc{
        Facultad de Ingeniería
    } } }    
    
    \vspace*{4mm}
    \large{ \textbf{ \textsc{
        Secretaría de Investigación y Posgrado
    } } }  
    
    \vspace*{5mm}
    \line(1,0){300}
    
    \vspace*{1.5cm}
    \includegraphics{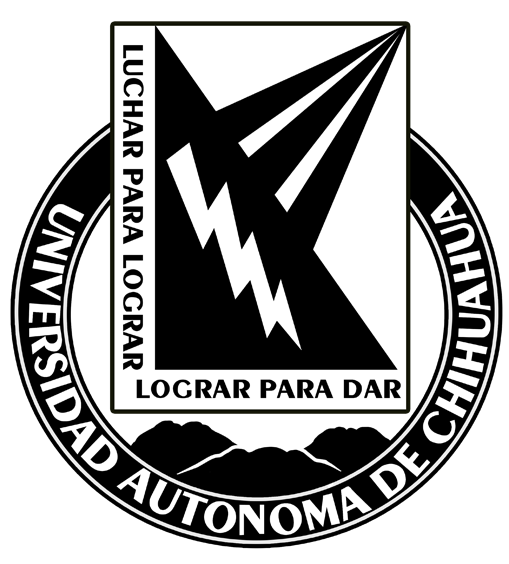}
    
    \vspace*{1cm}
    \huge{ \textbf{ \textsc{
        \vTitle
    } } }    
    
    \vspace*{1cm}
    \normalsize{ 
        \textsc{
            a dissertation submitted by
        } 
    }  
    
    \vspace*{5mm}
    \LARGE{ \textbf{ \textsc{
        \vAuthor
    } } }   
    
    \vspace*{7mm}
    \normalsize{
        \textsc{
            in fulfillment of the requirements for the degree of 
        } 
    }
    
    \vspace*{5mm}
    \Large{ \textbf{ \textsc{
        \vDegree
    } } }   
    
\end{center}

\vspace*{4cm}
\normalsize{ 
    \textsc{
        Chihuahua, Chih., Mexico.
        \hspace*{6cm}
        \vDate
    }
}

\restoregeometry

%% file: template/approval.tex
\newgeometry{
    top=3cm, 
    bottom=1cm, 
    left=6cm, 
    right=3cm
}

\newpage
\thispagestyle{empty}

\noindent
\justify
\vTitle. A dissertation submitted by \vAuthor in partial fulfillment of the requirements for the degree of \vDegree, has been verified and accepted by:

\vspace*{-1.9cm}
\hspace*{-3.7cm}
\includegraphics[keepaspectratio=true,height=2.5cm]{template/uach.png}
\vspace*{1cm}

\noindent\line(1,0){250} \\
\textbf{M.I. Javier González Cantú} \\
Director de la Facultad de Ingeniería \\

\vspace*{8mm}

\noindent\line(1,0){250} \\
\textbf{Dr. Alejandro Villalobos Aragón} \\
Secretario de Investigación y Posgrado \\

\vspace*{6mm}

\noindent\line(1,0){250} \\
\textbf{Dr. Alejandro Villalobos Aragón} \\
Coordinador Académico \\

\vspace*{6mm}

\noindent\line(1,0){250} \\
\textbf{\vDissertationDirector} \\
Dissertation Director \\

\vspace*{6mm}

\noindent
\textbf{\vDate} \\

\vspace*{6mm}

\noindent
Examination committee: \\

\vCommittee

\vspace*{1.5cm}

\hspace*{5.3cm}\fbox{
    \begin{minipage}{16em}
    \centering
    \textcopyright \hspace{1ex} All rights reserved\\
    \textbf{\vAuthor}\\
    Circuito Universitario Campus II,\\
    Chihuahua, Chih., México.\\
    \vDate
    \end{minipage}
}

\restoregeometry    

%% file: chapters/abstract.tex
\thispagestyle{plain}

\section*{Abstract}

Road transportation is of critical importance for a nation, having profound effects in the economy, the health and life style of its people. With the growth of cities and populations come bigger demands for mobility and safety, creating new problems and magnifying those of the past. New tools are needed to face the challenge, to keep roads in good conditions, their users safe, and minimize the impact on the environment.

This dissertation is concerned with road quality assessment and aggressive driving, two important problems in road transportation, approached in the context of Intelligent Transportation Systems by using Machine Learning techniques to analyze acceleration time series acquired with smartphone-based opportunistic sensing to automatically detect, classify, and characterize events of interest. 

Two aspects of road quality assessment are addressed: the detection and the characterization of road anomalies. For the first, the most widely cited works in the literature are compared and proposals capable of equal or better performance are presented, removing the reliance on threshold values and reducing the computational cost and dimensionality of previous proposals. For the second, new approaches for the estimation of pothole depth and the functional condition of speed reducers are showed. The new problem of pothole depth ranking is introduced, using a learning-to-rank approach to sort acceleration signals by the depth of the potholes that they reflect. 

The classification of aggressive driving maneuvers  is done with automatic feature extraction, finding characteristically shaped subsequences in the signals as more effective discriminants than conventional descriptors calculated over time windows. 

Finally, all the previously mentioned tasks are  combined to produce a robust road transport evaluation platform.

\newpage 
\thispagestyle{plain}

\begin{otherlanguage}{spanish}

\section*{Resumen}

El transporte terrestre es de importacia crítica para una nación, teniendo grandes efectos en la economía, salud, y estilo de vida de la gente. El crecimiento de ciudades y poblaciones crea mayores demandas de mobilidad y seguridad, originando nuevos problemas y acrecentando los del pasado. Por ello se requieren nuevas herramientas para  mantener los caminos en buen estado, sus usuarios seguros, y minimizar el impacto en el ambiente. 

Esta disertación trata la calidad de los caminos y la conducción agresiva, dos importantes problemas en el transporte terrestre, dentro del contexto de los Sistemas Inteligentes de Transporte, usando aprendizaje computacional para analizar series de tiempo adquiridas mediante sensado oportunista y así detectar, clasificar, y caracterizar automáticamente eventos de interés.

Se consideran dos aspectos de calidad de caminos: la detección y la caracterización de anomalías. Primero se comparan los trabajos más citados en la literatura, y se hacen propuestas capaces de igual o mejor desempeño, removiendo el uso de umbrales y reduciendo el costo computacional y dimensionalidad de las propuestas anteriores. Luego se tratan nuevas técnicas para la estimación de profundidad de baches y el estado de reductores de velocidad. Se presenta el nuevo enfoque de ordenado de baches, usando learning-to-rank para ordenar señales de aceleración respecto a la profundidad de los baches que reflejan.

La clasificación de eventos de conducción agresiva se realiza mediante extracción automática de características, encontrando que secuencias de las señales son indicadores más discriminativos que los descriptores convencionales. 

Finalmente, se combinan las tareas anteriores para producir una plataforma robusta para la evaluación del transporte terrestre.

\end{otherlanguage}

%% file: chapters/introduction.tex
\chapter{Introduction}

\section{Motivation}

The condition of roads is of critical importance for a nation. Safe roads that are well maintained have a positive influence on the quality of life and economic growth by increasing the mobility of people and goods, lowering transportation costs, and reducing pollution. A road transport infrastructure in bad condition not only affects ride quality, but also impacts the economy by reducing trade and general productivity, increasing vehicle operating costs, carbon footprint, and the number of accidents, which leads to injuries and loss of lives. Aging roads and larger traffic volumes increase both the need and the cost for infrastructure maintenance, and incorrect or insufficient repairs can double or triple the costs of both vehicle ownership and road upkeep. 

Road networks extend over millions of kilometers, with countries in the Organization for Economic Co-operation and Development (OECD) averaging over 500,000 km, and are frequently considered to represent as much as 5\% of the gross domestic product; when all road transport equipment and fuel are also considered, estimates become as high as 10\% or more \citep{piarc2014}. It has been seen that economic efficiency increases when the costs of transport is reduced but, unlike freight costs, automobile transport has been increasing during the last century and now occupies a significant part of the expenditures of households \citep{litman2018}. Bad road condition has a significant impact on the wear of vehicles \citep{bogsjo2009}, and delaying maintenance will not only increase the costs for vehicle owners but for the whole community because repairing a road in bad shape is four or more times more expensive than performing maintenance when it was in regular condition \citep{betanzo2008}.

Roads in good condition are only one factor in road transportation. The efficiency of transport is also linked to safety, which is affected by the regulations applied to roads and the capacity to enforce them, the driving skills and the mentality of the drivers. 

In the last years road accidents have increased, leaving tens of millions of people injured, disabled, or dead. These events are not only significant because of the great cost of emergency response and health care services, but because of the devastating implications and great grief suffered by the individuals, families, and whole communities that have to deal with the aftermath. Lack of safety standards and lax enforcement of traffic law, drivers that are distracted, fatigued, or under the influence of drugs or alcohol, speeding and careless driving, among other causes, have made road traffic injury the leading cause of death for people aged between five and twenty-nine years \citep{who2018}.

Early deployments of electronic technology  for the improvement of road transportation date from at least the 1960s, and very important planning and monitoring systems were implemented in the 1980s and 1990s, but it's in the last twenty years when computer science has entered the transportation world \citep{auer2016}. Precise and fast-flowing information for end-users and road administrators can improve the decisions and reduce the cost of the necessary actions to overcome the previously mentioned challenges \citep{unhabitat2016}. However, developing nations are the most affected by traffic problems, and are frequently less equipped and have the least resources to mitigate them. Therefore, technological solutions that can  mitigate the problems described above are of particular interest: inexpensive, automatic pavement evaluation for the identification of sections of damaged road, systems that can improve the coordination and prioritization of repairs, early warnings for vehicle drivers, insurance telematics that can shape rates to incentivize safe driving.

This dissertation is concerned with two main problems in road transportation: road quality assessment, with emphasis in road anomalies, and aggressive driving. Road anomalies refer to defects in roads (such as potholes, shoving, rutting, and blowups), or speed reducing elements (including speed bumps and humps), features of the road that are originated by deviations in its surface and affect the movement of the vehicles, the flow of traffic, and ride quality. Aggressive driving collectively covers driving maneuvers that are sudden and increase the risk of accident, actions such as harsh braking or acceleration, abrupt lane changing, evasive maneuvers, or producing an erratic trajectory. Specifically, the following tasks are addressed:

\begin{itemize}
    
    \item Accelerometer-based road anomaly detection. This problem consists on the inspection of acceleration time series to find anomalous subsequences that reflect the vibrations of the vehicle when it passes over a road anomaly. 
    
    \item Road anomaly characterization. It consists on estimating the depth of potholes (in cm), and the functional condition of speed reducers (i.e. if a speed bump or a line of metal bumps are, or not, in good condition as to force drivers to slow down), by analyzing acceleration time series.
    
    \item Aggressive driving detection. 
    Acceleration readings are inspected to find sequences of samples that reflect the performance of an aggressive driving maneuver by the driver of a vehicle.
    
    \item Aggressive driving classification. Once a subsequence of an acceleration time series has been found to contain an aggressive driving event, further analysis is performed to determine the specific type of maneuver performed.
    
\end{itemize}

These problems are addressed in the context of Intelligent Transportation Systems, by using machine learning techniques to automatically find a way to perform the tasks described above. Smartphones are used as opportunistic sensors, providing an ubiquitous and low cost data acquisition platform. 

\section{Aims and Objectives}

This study is focused on the analysis of techniques for the improvement of the automatic detection and characterization of events relevant to road transportation safety. Specifically, the creation of features that allow the application of machine learning algorithms to perform classification, regression, and ranking, over time series with high variability and high levels of noise.

To achieve these, the following objectives were established:

\begin{itemize}

	\item Analyze and explore previous approaches that deal with road anomaly detection, to establish the state of the art from the incohesive literature on the subject.
	
	\item Design a feature vector with high discriminative power, low computational cost, and low dimensionality for the detection of road anomalies to remove the need for threshold heuristics.
	
	\item Study and design machine learning strategies for a highly granular inference of the severity of potholes.
	
	\item Study the viability of condition assessment for speed reducers, to identify those that have lost effectiveness to perform their intended purpose.
	
	\item Study automatic feature extraction techniques based on subsequence regularization for the task of aggressive maneuver classification.
	
\end{itemize}

\section{Contributions}

In Chapter \ref{potholes}, the state of the art for accelerometer-based road anomaly detection is determined by directly confronting the methods proposed in the most widely cited works in the literature, over the same data and evaluated with the same methodology. This is the first time such comparison has been performed. 

A vector of twelve features that applies the detection criteria of the threshold based detectors was proposed as an alternative to the surveyed methods, obtaining superior results in terms of F-score, and being considered as equal or better after a critical difference analysis. In Chapter \ref{conclusions}, an alternative of similar functionality is presented: a vector with eight features for the separation of acceleration time series in three categories: those that reflect normal driving conditions, those in which an aggressive maneuver was present, and those capture while the vehicle passed over a road anomaly. The work on this topic found in in Chapter \ref{potholes} was published in \citet{carlos2018} and \citet{aragon2016}.

A feature vector containing time and frequency domain descriptors is presented in Chapter \ref{characterization} as an effective representation over which Machine Learning based techniques can be used for the characterization of road anomalies. Pothole depth estimation was achieved with 22\% relative error by means of regression, and 0.89 AUC was obtained for the classification of shallow and deep potholes. A new approach is also introduced: applying the learning to rank paradigm to obtain a function capable of sorting a list of time series reflecting potholes by their estimated depth. For this task, a Kendall tau coefficient of 0.34 was the best result.

The bag of words methodology was found to be an effective automatic feature extraction for the problems of detection and classification of aggressive driving maneuvers. As detailed in Chapter \ref{risky}, this representation can outperform state of the art proposals that rely on hand-crafted feature vectors. SAX and bag of features, two methodologies that share some similarities, were also evaluated to compare classification results with techniques based on subsequence similarity and conventional summarization, finding that the first have more discriminative power when it comes classifying specific types of event. These findings were published in \citet{carlos2018a}.

A pipeline that performs the tasks addressed in this dissertation is proposed and evaluated in Chapter \ref{conclusions}. It combines the findings and insights that resulted from this research, and serves as a guide for the creation of road transport evaluation platforms, managing to extract information about the state of the roads and the driving style of their users. 

\section{Products of this study}

The following lists present the articles in which parts of this dissertation have been published, submitted for publication, and/or presented in conferences, international as well as domestic.

\subsection{Peer Reviewed Journals}

\begin{itemize}

\item \textbf{Carlos, M. R.} González, L. C., Wahlström, J. J., Cornejo, R. \& Martínez, F. (2019). \textit{Becoming smarter at characterizing potholes and speed bumps from smartphone data -- introducing a second-generation problem}. (Manuscript under review in \textit{IEEE Transactions on Mobile Computing}) \textcolor{white}{\citet{carlos2018b}}

\item \textbf{Carlos, M. R.}, González, L. C. Wahlström, J. J., Ramírez, G. Martínez, F. \& Runger, G. (2019). How smartphone accelerometers reveal aggressive driving behavior -- The key is the representation. \textit{IEEE Transactions on Intelligent Transportation Systems}. (Advance online publication) doi: 10.1109/TITS.2019.2926639 \textcolor{white}{\citet{carlos2018a}}

\item \textbf{Carlos, M. R.}, Aragón, M. E., González, L. C. Escalante, H. J., \& Martínez, F. (2018, oct). Evaluation of detection approaches for road anomalies based on accelerometer readings -- addressing who's who. \textit{IEEE Transactions on Intelligent Transportation Systems,} 19(10), 3334-3343. doi: 10.1109/tits.2017.2773084 \textcolor{white}{\citet{carlos2018}}

\item González, L. C., Moreno, R., Escalante, H. J., Martínez, F. \& \textbf{Carlos, M. R.} (2017, nov). Learning roadway surface disruption patterns using the bag of words representation. \textit{IEEE Transactions on Intelligent Transportation Systems, 18}(11), 2916-2928. doi: 10.1109/tits.2017.2662483 \textcolor{white}{\citet{gonzalez2017}}

\end{itemize}

\subsection{Conferences}

\begin{itemize}

\item \textbf{Carlos, M. R.}, Martínez, F., Cornejo, R., \& González, L. C. (2017). Are android smartphones ready to locally execute intelligent algorithms? In \textit{Advances in soft computing} (pp. 15-25) Springer International Publishing. doi: 10.1007/978-3-319-62428-0\_2 \textcolor{white}{\citet{carlos2017}}

\item \textbf{Carlos, M. R.}, González, L. C., Martínez, F., \& Cornejo, R. (2016). Evaluating reorientation strategies for accelerometer data from smartphones for ITS applications. In \textit{Ubiquitous computing and ambient intelligence} (pp. 407-418). Springer International Publishing. doi: 10.1007/978-3-319-48799-1\_45 \textcolor{white}{\citet{carlos2016}}

\item Aragón, M. E., \textbf{Carlos, M. R.}, González, L. C., \& Escalante, H. J. (2016). A machine learning pipeline to automatically identify and classify roadway surface disruptions. In \textit{Proceedings of the sixteenth Mexican international conference on computer science -- ENC'16. ACM Press. doi: 10.1145/3149235.3149238} \textcolor{white}{\citet{aragon2016}}

\end{itemize}

\subsection{Data and Code}

The data collected for the experiments presented in this document, and the source code for their implementations are available at the following URLs:

\url{http://accelerometer.xyz/datasets}

\url{http://github.com/ricardo-carlos}

\section{Document Outline}

The rest of this document is organized as follows: Chapter \ref{lit_rev} presents background information about the general area of study and some terminology, after which the state of the art is summarized for the three main problems addressed in this study. Chapter \ref{reorientation} deals with the problem of aligning the axes of the sensors in a smartphone with those of the vehicle inside which it is placed, a non-trivial correction required to give spatial meaning to the signals collected from opportunistic sensing. 

The detection of road anomalies is addressed in Chapter \ref{potholes}, by means of threshold-based methods and machine learning algorithms applied to vertical acceleration. Chapter \ref{characterization} deals with the estimation of the depth of potholes and the condition of speed reducers, by applying classification, regression, and ranking techniques on different features extracted in the time and frequency from acceleration time series, their first derivative, and their first two integrals. The detection of risky driving by exploiting different representations is covered in Chapter \ref{risky}.

Finally, Chapter \ref{conclusions} presents a pipeline in which all of the previous problems are simultaneously addressed, followed by recommendations derived from the conclusions of this study.

%% file: chapters/background.tex
\chapter{Background and State of the Art}
\label{lit_rev}

\section{Intelligent Transportation Systems}

Intelligent Transportation Systems (ITS) is an area of study that integrates telecommunications, electronics, and information technology with transportation engineering to plan, design, operate, maintain, and manage transportation systems, aiming to increase their efficiency and safety, and allowing the flow of information between transportation management entities and road users while reducing the environmental impact. Governments \citep{EU2010-40} and professional associations \citep{itsa2019} see opportunities in the development of such systems to save lives, improve mobility, productivity, and quality of life. 

Even if ITS deals with all modes of transport, road transportation is the most widely discussed \citep{xu2016}, and the most visible modality. It is easy to see some ITS already in use, such as adaptive cruise control, collision avoidance and parking assistance systems, offered in vehicles manufactured in recent years. More complex systems are still being developed, aiming to fuse measurements and information from sources as varied as vehicles, pedestrians, and even infrastructure, to make inferences about the context and provide information to prevent and detect collisions, mitigate traffic, detect impaired drivers, score and improve a drivers' skills, or evaluate the condition of infrastructure \citep{engelbrecht2015, wahlstrom2017}. 

Because of the complexity of these tasks, ITS is very related to other paradigms and areas of study, like the Internet of Things, a paradigm that considers an environment full of sensors, mobile devices and radio-frequency tags that interact and cooperate with each other to reach common goals \citep{giusto2010}. Machine Learning is also of great importance for ITS, providing algorithms and models to make computers extract some meaning out of the contextual data and then be able to make predictions and take decisions, employing expertise gained in Natural Language Processing, Computer Vision, Signal Processing and Data Mining. 

 
 

\section{Road Roughness}

Many terms are used to describe defects in roads, and some times they are considered equivalent in the ITS literature. This lax terminology complicates the discussion and comparison of the different approaches in the literature for road anomaly detection. This lack of common terminology has been addressed in civil engineering, and manuals with standard terms and descriptions have been written for road professionals. However, this information is not as widely disseminated in the ITS community. A summary of this terminology is now presented as a reference.  

Roughness is one of the most intuitive ways to describe road condition, and is defined as ``the deviations of a pavement surface from a true planar surface with characteristic dimensions that affect vehicle dynamics, ride quality, dynamic loads, and pavement drainage'' \citep{bennett2007}, and the International Roughness Index (IRI) is the most widely accepted metric in civil engineering to measure it. This index quantifies unevenness by measuring total vertical displacement per unit of traveled distance (e.g., m/km) \citep{pierce2013}. This kind of metric reflects the general condition of a road segment, but does not deal with the number, location, severity, or type of individual defects found in the road.

An individual road anomaly is a small segment of road in which such deviation is found, and different types are discussed in civil engineering. Among these are distress features such as potholes, dropoffs, blowups, rutting, shoving, or joint and manhole cover deficiencies \citep{miller2014}. Traffic calming devices such as speed bumps, speed humps, speed cushions, speed slots \citep{johnson2004}, metal, plastic, or ceramic bumps \citep{cactus2019}, are also considered as road anomalies in this work. Table \ref{tbl:intro:anomalies} summarizes the definitions and characteristics for the above mentioned types of anomalies.

\begin{table}[!ht]
	\caption{Characteristics of some road anomalies.}
	\label{tbl:intro:anomalies}
	\centering
	\begin{tabular}{ L{3cm} L{7cm} L{3cm} }
	    \toprule
    	\textbf{Anomaly} & \textbf{Characteristics} & \textbf{Dimensions} \\
		\midrule
		Metal, plastic, or ceramic bumps & Hard half spheres, installed for signaling or speed reducing & Diameter: 10-20 cm, Height: 4-8 cm \\
		\midrule
        Speed bumps &  Asphalt, concrete, plastic, metal, or rubber raised areas with circular, or parabolic profiles & Length: 30-90 cm, Height: 7-10 cm \\
        \midrule 
        Speed humps &  Asphalt or concrete  raised areas with circular, parabolic, or flat-topped profiles & Length: 3-4 m, Height: 7-10 cm \\
        \midrule 
        Speed cushions & Raised areas across the road with 30+ cm separations & Width: 1-2 m, Length: 3 m \\
        \midrule 
        Speed slots & Raised areas across the road with 30+ cm separations & Width: 2 m, Length: 3-4 m \\
		\midrule 
		Potholes & Bowl-shaped holes in pavement surface & Diameter: 15+ cm, Depth: 2.5+ cm, Area: 0.02+ $m^2$ \\
		\midrule 
		Dropoffs & Difference in elevation   resulting from non-uniform settling of material layers & Height: 1+ cm \\
		\midrule
		Rutting & Longitudinal surface depression in the wheel path & Depth: 1+ cm \\
		\midrule
		Shoving & Vertical displacement of pavement & Height: 1+ cm \\
		\midrule
		Patches & Area of replaced or added material & Area: 0.1+ $m^2$, Height: 1+ cm \\
		\midrule
		Joint and manhole cover deficiencies & Cracking, breaking, chipping, fraying of slab edges & Height: 1+ cm\\
		\midrule		
		Blowups & Localized upward movement of material at transverse joints, loose fragments \\
		\bottomrule 
	\end{tabular}
\end{table}

\section{Mobile Sensing}

Mobile sensing refers to the usage of sensing devices that are not fixed to a specific location, that can move with a person, vehicle, or object of interest. It can be considered a branch of the mobile computing movement \citep[pp. 1-5]{yan2014}, and for a long time was mostly practiced by creating ad hoc sensing platforms assembled from an array of individual electronic components, some times paired with handheld PCs \citep{choudhury2008}, and frequently requiring researchers to work with purpose specific operating systems \citep{strazdins2010}. 

Materializing some of the visions of the ubiquitous computing movement \citep{weiser1999}, sensor-equipped mobile phones quickly established a new mobile sensing paradigm (mobile phone sensing, also called smartphone-based sensing) in the 2000s because of the standardized nature of their APIs and components, their processing, storage, and networking capabilities \citep{padmanabhan2008}. These devices not only offered clear technical benefits, but also came with a very quick adoption as an everyday item among the general population \citep{lane2010}. Currently, over three billion smartphones  \citep{newzoo2018} and 5.7 billion mobile broadband subscriptions \citep{ericsson2018} are  considered to be in active use. Smartphone-based sensing is now so common that it is frequently assumed when the term mobile sensing is used, even if it is not the only mobile sensing alternative. Among the sensors that can be found inside smartphones are accelerometers, gyroscopes, magnetometers, thermometers, barometers, hygrometers,  global navigation satellite system  receivers, cameras, and microphones \citep{androidsensors}.

Two modalities can be identified when mobile devices are used as sensing nodes: participatory, and opportunistic. Participatory sensing gets its name from the fact that users of the mobile phones are directly involved in the sensing process, performing direct action or making decisions about what data is collected and when. On the other hand, opportunistic sensing leaves to software in the mobile device the specific sensing actions and decisions, leaving the owner of the device free to perform normal daily activities without being concerned with data acquisition tasks \citep{khan2013}. 

Of particular interest for ITS are smartphone-based vehicle telematics, an application of mobile sensing in which data collection is performed by mobile devices from inside moving vehicles, and transmitted by means of mobile networks. This form of sensing can be considered superior to sensors fixed to vehicles, not just because it is easier and cheaper to deploy, scale, and upgrade the sensing platform, but because it allows a two-way communication channel, allowing to provide some form of instantaneous feedback to the drivers \citep{wahlstrom2017}.

The benefits of mobile sensing do not come without challenges, and among the most relevant we can find: 

\begin{itemize}
\item the low quality of sensors, since their intended purpose is to offer interaction capabilities at a low cost and not high precision measurements;

\item unpredictable orientation of the sensors' axes and location of the device itself, because users interact with the devices in many different ways during their daily activities;

\item noise in the readings, both from the sensors themselves and from the some times unpredictable interaction of the users with the device;

\item battery drain, caused by the extra usage of the sensors, CPU, and wireless networks by the sensing applications \citep{wahlstrom2017}.

\end{itemize}

Even if the accelerometers on smartphones are by no means of high precision, their performance is comparable with specialized sensors for low frequency excitations ($\leq$20 Hz) \citep{dedominicis2014}, which correspond to most of the phenomena of interest for ITS. The receptors for Global navigation satellite systems (such as Global Positioning System (GPS), Global Navigation Satellite System (GLONASS), and Galileo) allow us to get an approximate location, speed and direction of travel from smartphones, and the widely deployed Assisted-GPS allows us to get location estimates faster than with conventional receivers \citep{bierlaire2013}. However, the usual expected error in location estimates ($<$10 m) might be increased by buildings in urban settings because constructions can prevent a direct line of sight of the constellation of satellites of the navigation system, and might create reflections of a satellite's signal \citep{miura2015}. These problems are well known, and there are proposals for methods to simulate the expected error \citep{giofre2017}, cluster locations to minimize the effect of noisy data \citep{strutu2014}, or attempt to reduce the error in location with more advanced methods  \citep{bo2013, miura2015}.

\section{Time Series Fundamentals for ITS}

\subsection{Sensor Reorientation}
\label{rotations}

\begin{figure}
    \centering
    \subfigure[ ]{
	    \includegraphics[height=3cm]{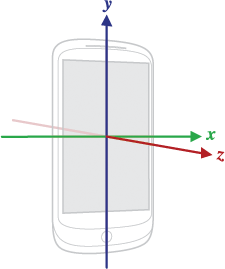}
	    \label{fig:intro:axes_a}
    } 
    \hspace{2cm}
    \subfigure[ ]{
	    \includegraphics[height=3cm]{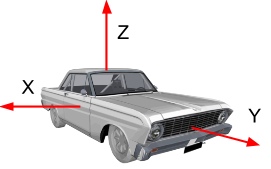}
	    \label{fig:intro:axes_b}   
    } 
    \caption[Coordinate systems used by the Android API]{Coordinate systems used by the Android API for a) for mobile devices, and b) automotive implementations. Arrows show the positive direction for each axis. Source: Android Open Source Project, accessed 2019-01-19, \url{https://source.android.com/devices/sensors/sensor-types}. }
	\label{fig:intro:axes}
\end{figure}

The accelerometers, gyroscopes, and magnetometers found in Android smartphones report their readings in a coordinate system with three orthogonal axes ($X$, $Y$, and $Z$) assigned in a convention, shown in Figure \ref{fig:intro:axes_a}, that is relative to the mobile device and is not changed by interaction with the device. When data is being captured, each of the above mentioned sensors returns a vector in three dimensions that correspond to these axes. 

A similar coordinate system can also be assigned to a vehicle in order to model its movement. The convention used in this text for the axes of a vehicle is the one defined in the Android automotive implementation, shown in Figure \ref{fig:intro:axes_b}. In this convention the meaning of the axes, from the perspective of the driver of the vehicle, is as follows: $X$ is the lateral axis (i.e. left-right), $Y$ is the longitudinal axis (i.e. backward-forward), and $Z$ is the vertical axis (i.e. down-up).

It is possible to place a smartphone inside a car and use the mobile device as an inertial measurement unit (IMU) to acquire data about the movement of the vehicle. 
However, in order to make sense of the reported readings, the axes of the sensors must be aligned with the axes of the vehicle. Axes are said to be aligned if the angle between each axis of the sensor and each of the corresponding axes of the vehicle is zero, and are  otherwise considered unaligned. If, for example, the $Z$ axis of the sensors and the vehicle is properly aligned and there is a misalignment of 135º for the $Y$ axis, the accelerations for an event of forward acceleration in a straight line for the vehicle will be interpreted as breaking and turning to the right from the data collected in the smartphone.

Given that unaligned sensors are to be expected most of the time for opportunistic sensing (because smartphones are frequently manipulated by their users and the position and orientation of the devices cannot be assumed under normal day-to-day usage), the issue of correcting sensor orientation becomes very relevant. 

Reorientation (i.e. taking values from one frame of reference and expressing them in another) is possible by means of a linear transformation, as demonstrated by Euler's rotation theorem. This theorem states that, regardless of how a coordinate system is rotated, it is always possible to find an axis in space about which a rotation of the initial values ends at the final, desired, orientation \citep{bar-itzhack1989}. This transformation can be expressed as three successive rotations \citep[pp. 83-84 ]{kuipers1999} and there are different possible conventions, because the same transformation can be accomplished with different rotation matrices applied in different orders \citep[pp. 150-151, 607-610]{goldstein2001}. One such convention is now described.

Consider two coordinate systems, namely the device frame of reference ($x,y,z$) and the vehicle frame of reference ($x',y',z'$). We can transform the first, reflecting what the smartphone sensed, into the second, that corresponds to the standard frame of reference of the car, by means of the rotation matrix $R$:

\begin{equation} \label{eq:intro:rotation}
    \begin{pmatrix}
        x'\\
        y'\\
        z'\\
    \end{pmatrix}
    = R
    \begin{pmatrix}
        x\\
        y\\
        z\\
    \end{pmatrix}    
\end{equation}

$R$ is the product of three rotation matrices: $R = R_z \times R_y \times R_x$, where 
\begin{equation} \label{eq:intro:r}
    R_x = 
    \begin{pmatrix}
        1 & 0           & 0 \\
        0 & \cos{\beta} & -\sin{\beta} \\
        0 & \sin{\beta} & \cos{\beta} \\
    \end{pmatrix},
    R_y = 
    \begin{pmatrix}
        \cos{\alpha} & 0 & -\sin{\alpha} \\
        0            & 1 & 0 \\
        \sin{\alpha} & 0 & \cos{\alpha} \\
    \end{pmatrix},    
    R_z = 
    \begin{pmatrix}
        \cos{\varphi}  & \sin{\varphi} & 0 \\
        -\sin{\varphi} & \cos{\varphi} & 0 \\
        0              & 0             & 1 \\
    \end{pmatrix}
\end{equation}

From these equations we can see that three angles are required to perform the rotation: $\alpha$, $\beta$, and $\varphi$, which correspond to roll, pitch, and yaw, respectively. The first two values can be easily extracted considering acceleration readings of a smartphone ($a$) and gravity ($g$), as follows:

\begin{equation} \label{eq:intro:a_b_linear}
    \alpha = \arcsin(a_x/g),
    \qquad
    \beta =  \arcsin(a_y/g)
\end{equation}

If we are only interested in performing vertical reorientation (that is, only extracting the vertical component of acceleration from the three axes without caring for the orientation of the two other axes), we can consider $\varphi=0$. In this case, the values of $x'$ and $y'$ will still be in the smartphone's frame of reference and might not reflect the experience of the driver of the car. If we are interested in readings for the longitudinal and lateral axes that match the experience of the persons inside the vehicle, we need to determine the value of $\varphi$. This is not a trivial calculation and different procedures reported in the literature to estimate this angle, along with their limitations, are addressed in Chapter \ref{reorientation}.

Given that most of the time the strongest acceleration experienced in a vehicle comes from gravity, the vertical component of all accelerations detected by the sensors can be approximated with the Euclidean norm of the values in the three axes \citep{jain2012}, a simpler and faster calculation if we are only interested in  acceleration in the vertical axis:

\begin{equation} \label{eq:intro:l2}
    z' \approx \sqrt{x^2 + y^2 + z^2}
\end{equation}


\subsection{Time Series Representations}


Time series are collections of observations registered chronologically. They present challenges for analysis because of their high dimensionality, the large size of the volume of data that conforms it, the speed at which data collections are updated, and the noise present in them. A fundamental problem in time series analysis is how to represent them, finding approaches to project them into different domains, frequently with lower dimensionality, that are more suitable for tasks such as clustering, classification, and regression \citep{fu2011}.

Filtering is a very common first step in time series feature extraction, although not mandatory. Mathematical transformations are applied to remove unwanted noise and isolating  useful components in data sequences. Some of the most commonly used filters come from analog signal processing: low-pass, high-pass, band-pass, Butterworth, Chebyshev, and Bessel filters \citep[pp. 19-64]{schlichthrle2011}, with the first three being more commonly used for IMU data from smartphone sensing in ITS applications. These filters remove low or high frequency components, which are assumed to not provide useful information in other bands. Moving average and moving median filters, based on replacing each data point with the mean or median of neighboring data points \citep[pp. 277-282]{smith1999}, are also commonly used. 

\begin{figure}[!ht]
    \centering
    \includegraphics[width=8cm]{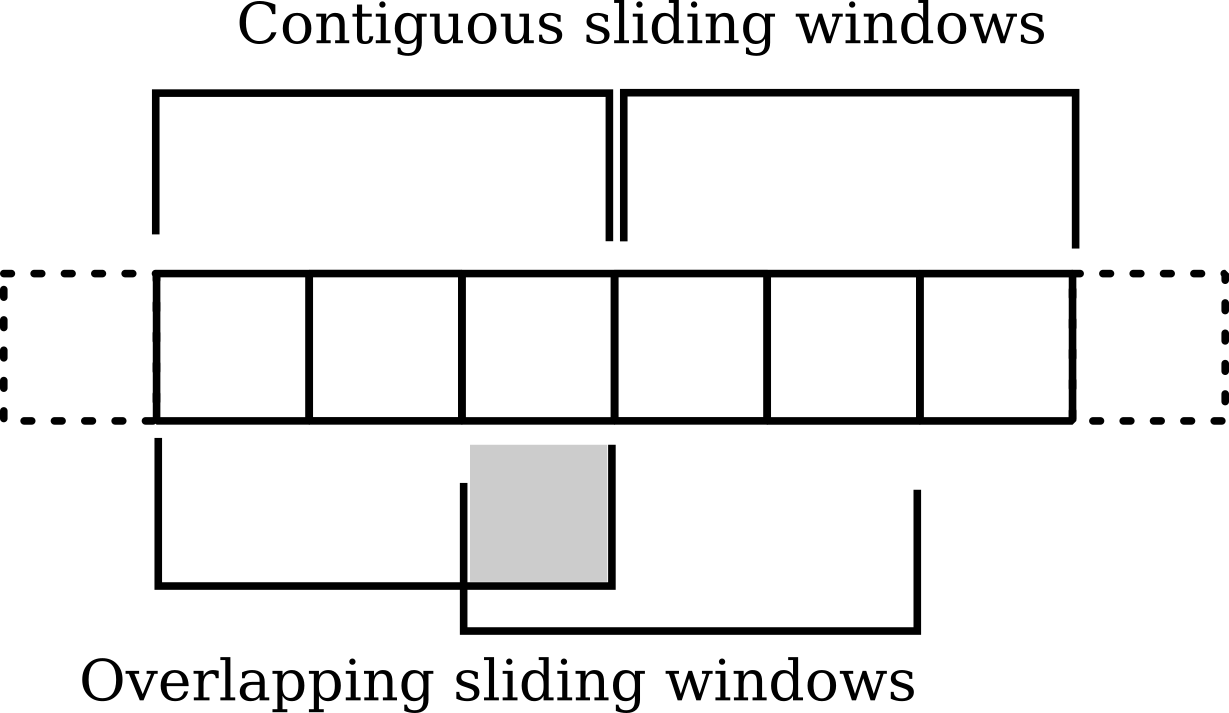}
    \caption{Examples of sliding windows with lengths of three elements, with and without overlapping.}
    \label{fig:background:windows}    
\end{figure}

After filtering comes the application of sliding windows, shown in Figure \ref{fig:background:windows}. A window is defined as a subsequence of data points in a time series, and sliding refers to starting and terminating those subsequences at a regular number of indices in the series. With sliding windows it is possible to discretize a long time series into subsequences of fixed length, producing patterns that can be considered as derived from sine curves \citep{fu2011}. Window length depends on the specific sampling frequency and task that is being performed, and in some cases windows overlap is desired. Overlap occurs when two windows share a number of data points because the interval at which sliding windows are extracted is less than the number of data points in the window. This practice increases the number of windows to process, but also increments the probabilities of not truncating the important patterns. If the time series is not too long for the desired application, it can be considered in its entirety as one window. Further processing for time series often occurs at the window level. 

Centering and scaling are preprocessing techniques  used to regularize data sets that contain time series (or windows) with different scales and ranges, with noticeable outliers or presenting heteroscedasticity. A simple way to perform a correction is by calculating the mean ($\bar x$) and standard deviation ($\sigma$) of the elements in the time series ($\{x_0 \ldots x_n$\}), and then transforming all data points as $x_i = (x_i - \bar x)/\sigma$. More complex strategies exist in which least squares analysis is performed over the data set, and each data point is then expressed in terms of the results. Among the benefits of centering and scaling are: reduced complexity of the models required to process the time series, improved fit to the data, avoidance of numerical problems \citep{bro2003}.

Different measures calculated over windows, both from the time and frequency domains, can be used as features to represent them for further analysis. Among the most frequently used we can find: mean, median, standard deviation, range, maximum value, minimum value, root mean square, trapezoidal integral, zero-crossings, signal magnitude area, signal vector magnitude, jerk profile, differential signal vector magnitude, DC component, spectral energy, information entropy, dominant frequency \citep{figo2010}. 

The calculation of these features transforms a series of data points into single numbers, summarizing some aspect of the whole window. More detailed analysis can be performed on the output of the Fourier or short-time Fourier transforms, focusing on specific bands in the frequency domain. An alternative is provided by wavelets, which project their input onto a set of basis functions that are not sinusoidals and allow analysis at different resolution levels but also preserving temporal information \citep{barford1992}. The spectral (or detail) coefficients are used individually (at all or specific levels of analysis), or combined by addition or some other operations \citep{figo2010}, and multiple coefficients from analysis at different levels can be combined to obtain features from signals \citep{diab2012}.

Another approach to define features for time series is some form of comparison in the time domain. This can be done, for example, with the Pearson correlation coefficient or the cross-correlation of a time series with respect to another \citep{figo2010}. Dynamic Time Warping (DTW) is a shape similarity metric that allows two time series of different length to be compared by using dynamic programming to find the best possible alignment  \citep{petitjean2011}. Other metrics can also be used, like the Mahalanobis distance \citep{prekopcsk2012}, to establish similarities among time series in a data set.

\begin{figure}[!ht]
    \centering
    \includegraphics[width=9cm]{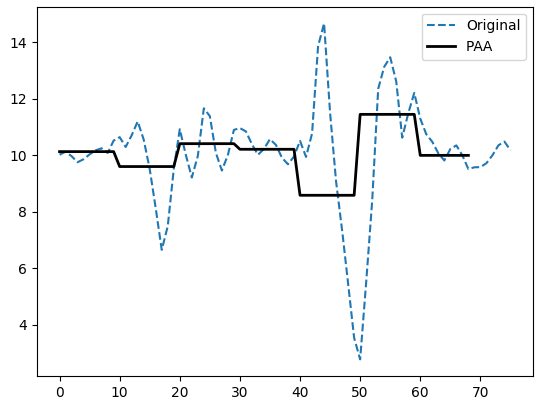}
    \caption{A plot for a time series, along its PAA representation using a word size of 10 and alphabet size of 7.}
    \label{fig:background:paa}    
\end{figure}

A simple approach for dimensionality reduction is Piecewise Aggregate Approximation (PAA), in which  $n$ neighboring data points in a window form a word and are represented by their mean value, essentially producing an approximation of the original series with a linear combination of box functions \citep{keogh2001}. If the number of possible box functions is restricted to $m$ amplitude values (the alphabet size), the output of PAA can be discretized to produce a piecewise representation of a time series that uses an alphabet of size $m$; this approach is called Symbolic Aggregate Approximation (SAX) \citep{lin2003, senin2013}. Once a time series is converted to a string of symbols, it becomes possible to apply text processing algorithms. Figure \ref{fig:background:paa} shows a time series, along its PAA representation. 

This idea of finding discrete entities is extended by the concept of a shapelet, a subsequence in a time series that is representative of the event of interest \citep{ye2009}. This concept links a visual idea, a shape that can be seen when plotting data, with a time series reflecting something that is not related to visual information. A sequence of such discrete elements can be then treated like text, as a sequence of symbols. 

\begin{figure}
	\centering
	\includegraphics[width=14cm]{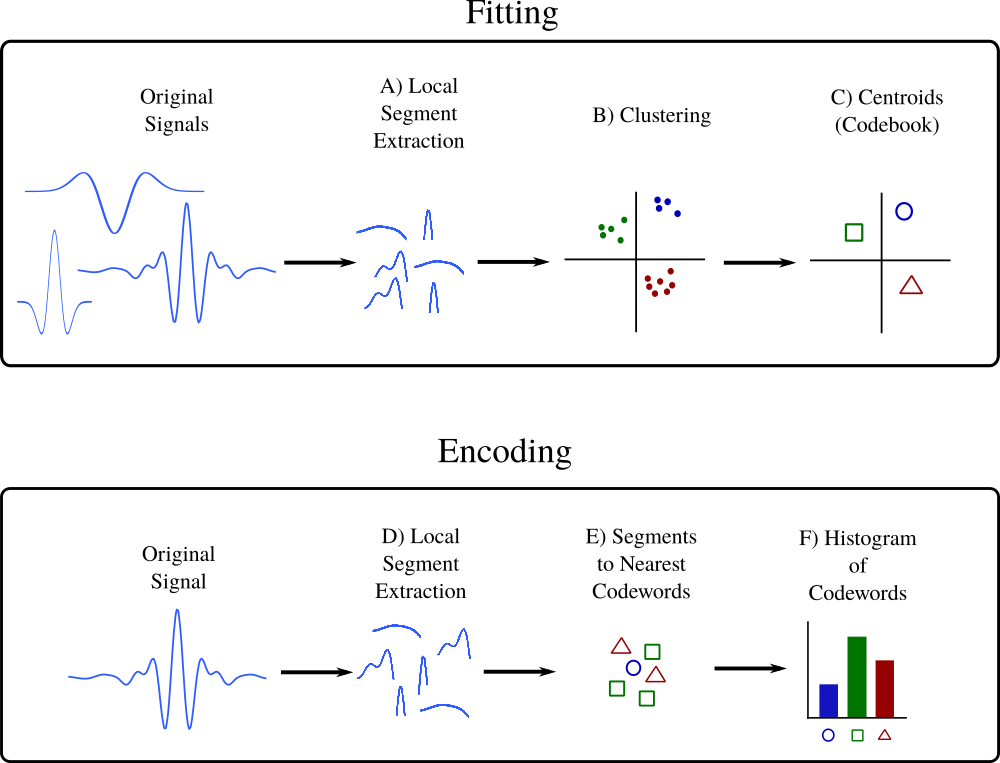}
	\caption[Bag of words model]{
		Application of the two stages of a Bag of Words model. In the first stage (fitting): A) a collection of time series is split in segments of length $L$. B) all segments are considered as points in an $L$-dimensional space, and a clustering algorithm is used to find $K$ clusters. C) The centroids of the clusters are considered as \textit{codewords}, collectively forming a \textit{codebook}. In the second stage (encoding): D) each signal is split in segments of length $L$. E) each window is represented as a codeword by finding the nearest previously determined centroid, converting the time series to a sequence of codewords. F) a histogram is calculated for this sequence of codewords, a vector of $K$ dimensions containing the number of times each codeword is required to recreate the original time series.
	}
    \label{fig:risky:bow}
\end{figure}

A time series can be split in windows of size $L$ and each window can be then considered as a vector with $L$ dimensions. By means of clustering, a set of $K$ centroids can be found, and a concatenation of those centroids (that can be considered shapelets) approximates the original time series. The centroids found are expected to be more representative of the data points they replace than their mean value, used in representations such as SAX, or some other descriptor. This centroid-based approach has been successfully combined in the bag of words methodology for time series classification \citep{wang2013}, by first learning regularized segments of time series (fitting) and then allowing to represent new signals (encoding) using only the learned elements. Figure \ref{fig:risky:bow} describes the bag of words model.

Given the high noise content frequently found in time series data \citep{lin2012}, random sampling has been applied to attempt to extract representative shapelets with the Bag of Features approach \citep{baydogan2013}, in which multiple subsequences of random length are selected from random locations, and then regularized with a Random Forest. Both the centroid-based based bag of words and the bag of features methodologies transform a time series into a histogram representation that indicates the frequency of the learned shapelets in the time series being transformed, a technique commonly used for natural language processing.

\subsection{Machine Learning Tasks}

Machine learning deals with the algorithms and statistical models that computers can use to solve problems without being programmed to specifically run the solution. Among the most common tasks performed with these techniques are classification, regression, and ranking, under a supervised scheme (analyzing examples for which the right result is known, and adjusting the model to produce the desired output). 

In the case of classification, a function $\phi$ is found from pairs of examples ${(x_i, y_i)}$ in which $x$ are feature vectors in $m$ dimensions and $y$ are discrete values that represent the class of each example. The intention is to be able to predict the unknown class to which new feature vectors are associated. The process for regression is similar, except that the function's output is not discrete, but continuous.

Ranking is a different task in which the input for the function is a list of feature vectors, and the output is a permutation of the input vectors, now with a different order. This function is found by analyzing list examples that present this order, and can be done with a point-wise approach, by using the output of a regressor or classifier for one sample as the ordinal score for the ranking. Another possibility, a pair-wise approach, is done by training a classifier that takes as input two feature vectors, and its two possible output labels indicate if the first or second element in the pair has the higher ordinal score, or not. This classifier can be used as a comparison function to be applied in a sorting algorithm, to produce a list with the desired order \citep{rigutini2011}.

\subsection{Evaluation Metrics}

An important part of the machine learning process is the evaluation of the proposed model, in which the suitability of the learned function for the specific task is quantified with metrics. Given that most of the time only a limited amount of data is available for model development, a technique used to ensure that the model is general enough to produce good results on an independent data set is the statistical analysis of the metrics obtained with cross-validation. 

Cross-validation consists on splitting data in two sets with no intersection: a training set, used to train the model, and a testing set, used to evaluate its performance. This process of splitting data is repeated several times, and  statistical analysis is performed on the metrics calculated with the testing samples. Multiple cross-validation techniques are found in the literature, such as shuffled splits and k-fold, both in their regular and stratified variants \citep{arlot2010}.

Shuffled splits produce random permutations of the elements in the data set, which are then split in sets assigned for training and testing, frequently leaving 20\% or more of the elements in the last group. K-fold  divides the data set in $K$ groups,  performing training and testing $K$ times, each time designating one of these $k$ groups for testing and using the rest of the data for training. In the stratified versions the total percentage of samples in the data set for each class is preserved in both the training and testing sets.

Different learning metrics exist for the most typical supervised learning tasks of classification, regression, and ranking, allowing to quantify different characteristics of the performance for the evaluated method. Descriptions for the metrics used in the rest of the document are now presented.

\subsubsection{For Classification}

The metrics used for classification tasks are frequently discussed in terms of binary classification, but can be extended to a multiclass scenario. Two outputs are possible for binary classification of an item: positive, if the item belongs to the target class, and negative if it doesn't (which is the same as saying that it belongs to the other class).

Once a classification function has being established from training data, it is applied to a set of testing examples (for which the correct output is known, but of which the classifier has no knowledge). A prediction is made for each element in the testing set, and the output is compared to the known correct answer (ground truth), with four possible outcomes. A binary prediction is considered a true positive (TP) if both the prediction and the real known class a sample indicate that the sample belongs to the target set; a true negative (TN) is when both the prediction and ground truth indicate that the sample does not belong to the target set; when the prediction is negative (the sample does not belong to the target set) but the correct answer is that the sample does belong to the target set, a false negative (FN) occurs; the fourth possibility, false positive (FP), is that the sample does not belong to the target set, but the prediction indicated that it does. Some common metrics for the evaluation of binary classification, based on the balance of the four possible outcomes, are now described:

\begin{itemize}

\item \textbf{Accuracy}: the proportion of correct results (both TP and TN) among the total of examined cases.

\begin{equation}
\small{Accuracy=\frac{TP+TN}{TP+FN+FP+FN}}
\end{equation}

\item \textbf{Sensitivity}: the proportion of TP that are correct; also known as true positive rate, or recall.

\begin{equation}
\small{Sensitivity=\frac{TP}{TP+FN}}
\end{equation}

\item \textbf{Specificity}: the proportion of actual negatives that are correctly identified.

\begin{equation}
\small{Specificity=\frac{TN}{TN+FP}}
\end{equation}

\item \textbf{Precision}: the proportion of TP out of the total number of detections.

\begin{equation}
\small{Precision=\frac{TP}{TP+FP}}
\end{equation}

\item \textbf{F-measure}: also known as F1-score, the harmonic mean of precision and sensitivity. 

\begin{equation}
\small{F1=\frac{2TP}{2TP+FP+FN}}
\end{equation}

\item \textbf{G-means}: the root of the product of class-wise sensitivity. It is used when there is imbalance in the number of examples for each class, and penalizes results when the classifier is better at positive over negative examples, and vice versa \citep{kubat97}.

\begin{equation}
\begin{split}
G & = \sqrt{\frac{TP}{(TP+FN)} \times {\frac{TN}{(TN+FP)}}}
\end{split}
\end{equation}

\item \textbf{AUC}: Area under the receiver operating characteristic (ROC) curve. Reflects the degree of separability between classes at all levels of the decision thresholds, in terms of probability.  

\begin{equation}
\small{ 
AUC = \frac{1}{2}\left(\frac{TP}{TP+FN}  + \frac{TN}{TN+FP}\right),
}
\end{equation}

\end{itemize}

\subsubsection{For Regression}

Since regression output is continuous, the  metrics used to evaluate it focus on quantifying the difference between the predictions ($y$) and the expected values ($\hat{y}$).  

\begin{itemize}

\item \textbf{Relative error}: The average of errors as a ratio of the expected value.

\begin{equation}
	\label{eq:error_rate}
    RE = \frac{1}{n}\Sigma_{i=1}^{n}{\Big|\dfrac{y_i - \hat{y}_i}{y_i}\Big|}
\end{equation}

\item \textbf{Root-mean-square error}: The magnitude of the average error, in the same units of the target value.

\begin{equation}
RMSE = \sqrt{\frac{1}{n}\Sigma_{i=1}^{n}{\Big(y_i - \hat{y}_i \Big)^2}}
\end{equation}

\item \textbf{Coefficient of determination}: The extent to which the variance of the variable of interest is predictable. For this metric a value between 0 and 1 indicates the extent to which unknown examples will be correctly predicted; zero and negative values indicate poor prediction.

\begin{equation}
\quad R^{2}(y_i, \hat{y}_i) = 1 - \frac{{\sum_{i=1}^{n} {(y_i - \hat{y}_i)^{2} } }}{{\sum_{i=1}^{n} {(y_i - \bar{y})^{2} } }},
\end{equation}

\end{itemize}

\subsubsection{For Ranking}

In the case of ranking, a possible way to quantify the quality of the output of a ranking proposal is to measure the discrepancies between its output and the the desired order.

\begin{itemize}

\item \textbf{Kendall rank correlation coefficient ($\tau$)}: The proportion of different pairs between two ordered sets \citep{salkind2007}. $P$ is the number of concordant pairs, $Q$ is the number of discordant pairs, $T$ is the number of ties only in the first list, and $U$ is the number of ties only in the second list (a tie is not counted if it occurs in both lists)

\begin{equation}
\small{ 
\tau = \frac{P-Q}{\sqrt{(P+Q+T)*(P+Q+U)}} 
}
\end{equation}

\end{itemize}

%% file: chapters/lit_rev.tex

\section{Road Anomaly Detection and Classification}

The usage of mobile sensing to indirectly estimate road roughness dates from at least 2001, when attempts were made to correlate low speeds of timber hauling trucks to poor road quality \citep{forslof2015}. This initial analysis was followed by the usage of embedded systems equipped with GPS, cellular network communication, and high-resolution accelerometers placed at the rear axle of a vehicle to analyze vibrations and determine the condition of roads. That system was 70\% correct when compared to the average results of subjective visual inspections performed by experts, and was considered a cheaper alternative to other procedures \citep{forslof2015}. By 2002, proposals for the use of wireless networks for vehicle telemetry start to appear, focusing on the detection of shocks (produced by potholes or the unexpected release of load) in industrial and commercial transport vehicles \citep{rouillard2002}. In 2008, Fourier analysis was used to calculate the power spectral density function of the surface of roads from fixed accelerometers installed in a fleet of vehicles. This analysis allowed to classify the overall condition of segments of roads as very good, good, or poor, with a more detailed investigation (requiring specialized equipment and personnel) being necessary for roads found to be in poor shape \citep{gonzalez2008}. 

The detection and classification of individual road anomalies was first addressed by \citet{eriksson2008}. Their proposed system uses an embedded computer to record GPS coordinates and the readings of vertical and lateral acceleration provided by accelerometers fixed inside the cabin. Data was analyzed by applying a series of threshold filters to classify time series reflecting smooth road, potholes, manhole covers, railroad crossings, and expansion joints. 

Smartphone-based mobile sensing was first applied for individual road anomaly detection by \citet{mohan2008}. Among a general evaluation of the capabilities of smartphones for ITS applications, their work presents detection of potholes and bumps by means of threshold discriminants. They only address the detection of anomalies, without detailed classification. 

After these seminal works, the literature on road anomaly detection can be navigated by categorizing works in those that use threshold-based algorithms, and those that employ Machine Learning techniques. In both cases, the objective of the authors was to adjust the parameters of their algorithms and find a transformation for their data in order to perform detection or classification over their own data sets. Threshold-based algorithms can be further subdivided in those that work with raw data, and those that work with transformations (such as calculating the first or second derivative of the data, or some statistical descriptor calculated from a sliding window).

Among the threshold-based works we find \citet{desilva2009}, who investigated a statistically defined threshold of $m$ times the standard deviation for vertical acceleration processed with a low-pass filter for pothole detection. \citet{sebestyen2015} applied a simple moving average filter to tridimensional acceleration to detect bumps and potholes. \citet{orhan2013} and \citet{fazeen2012} work with raw acceleration, but attempt to classify pothole and bumps from raw vertical acceleration data. \citet{yagi2010} combined vertical and lateral information from both accelerometers and gyroscopes to detect bumps with a threshold applied to the standard deviation of windowed signals. \citet{yi2015} also evaluated thresholds over the standard deviation, but only considering vertical acceleration. \citet{mednis2011} investigated different threshold heuristics to detect potholes of different sizes, pothole clusters, gaps, and drain pits, without being concerned of the specific type of anomaly in the road, working on raw vertical acceleration, its first derivative, and on the standard deviation of sliding windows. \citet{wang2013} evaluated similar approaches, but only for pothole detection. 

\citet{astarita2012} worked with a series of threshold filters over raw vertical acceleration to detect multiple types of road anomaly, and later used a high-pass filter and a custom data-transformation based on the rage of windowed data \citep{astarita2014}. Another transformation was tested by \citet{sinharay2013} for road anomaly detection, followed by classification as potholes or bumps, using thresholds over the second derivative and standard deviation for low frequency acceleration readings on three axes. \citet{kalra2014} evaluated the first derivative of low frequency triaxial acceleration for anomaly detection. \citet{harikrishnan2017} worked on detection and classification of bumps and potholes by considering the maximum of absolute values in lateral and vertical acceleration. A summary of the characteristics of the previous threshold-based methods is presented in Table \ref{tbl:lit_rev:thresh_methods}.

\begin{table}[!ht]
	\caption[Summary of data characteristics for works employing threshold-based algorithms.]{Summary of data characteristics for works employing threshold-based algorithms.}
	\label{tbl:lit_rev:thresh_methods}
	\centering
\resizebox{\columnwidth}{!}{%
	\begin{tabular}{ llllll }
	    \toprule
    	\textbf{Work} & \textbf{Sensors} & \textbf{Axes} & \textbf{Freq. (Hz)} & \textbf{Filter} & \textbf{Features} \\
		\midrule		
		
		\citet{eriksson2008} & Acc & Z, X & 380 & High-pass & Acc. \\
		\citet{mohan2008} & Acc & Z & 310 & &  Acc.\\
		\citet{desilva2009} & Acc & Z & 100 & Butterworth & Acc.\\
		\citet{yagi2010} & Acc, Gyr & Z, X & 100 & &  Stdev.\\
		\citet{mednis2011} & Acc & Z, X, Y & 26, 74, 52, 98 & & Acc., stdev, jerk\\
		\citet{astarita2012} & Acc & Z & 5, 100 & & Acc.\\
		\citet{fazeen2012} & Acc & Z, X & 25 & High-pass & Acc.\\
		\citet{orhan2013} & Acc & Z & 40 & & Acc.\\
		\citet{sinharay2013} & Acc & Z, X, Y & 4-6 & & Mean, stdev.\\
		\citet{astarita2014} & Acc & Z & 5, 16, 50, 100 & High-pass & Range\\
		\citet{kalra2014} & Acc & Z, X, Y & 5 & &  Jerk\\
		\citet{sebestyen2015} & Acc & Z, X, Y & 90 & SMA & Acc.\\
		\citet{wang2015} & Acc & Z, X, Y & 124 & &  Acc., jerk, stdev.\\
		\citet{yi2015} & Acc & Z & 40, 80, 100 & &  Stdev.\\
		\citet{harikrishnan2017} & Acc & Z, X & 50 & Max-abs & Acc.\\

		\bottomrule 
	\end{tabular}
}%
\end{table}

Among the first applications of machine learning algorithms we find \citet{tai2010}, applying a SVM over statistical features (mean, range, standard deviation, maximum, minimum) of triaxial acceleration and speed split with sliding windows in order to detect potholes, sunk-in manhole covers, and missing pavement. One of the most robust feature vectors in the literature was presented in \citet{perttunen2011}, where more than 95 features in the time and frequency domains were classified by a SVM to detect bumps, speed bumps, and large potholes from normal road. Other works take similar approaches, producing feature vectors that summarize sliding windows with statistical measures and feeding them into support vector machines \citep{bhoraskar2012, jain2012, mohamed2015}.

Statistical summarization and frequency analysis are the source of most features, but wavelet-based transformations (common in digital signal processing applications) have been some times used \citep{cong2013, seraj2015, perttunen2011}. Mel-frequency cepstral coefficients (common in speech analysis) also have been explored to enrich the calculated representations \citep{tecimer2015,perttunen2011}. In terms of machine learning algorithms, SVM has been the classifier of choice, but works using other alternatives exist \citep{martinez2014, tecimer2015, aragon2016, silva2017, brisimi2016}.

Different techniques have started to be explored in the last years. The bag of words representation was first applied to road anomaly classification in \citet{gonzalez2017}, presenting an alternative to the feature vectors described above that is capable of surpassing their performance with less involvement of the researcher in the feature engineering process. The fusion and direct comparison of threshold-based heuristics and machine learning technique were addressed in \citet{carlos2018}, where an ensemble of detectors is proposed to take advantage of the different strengths of the most relevant strategies presented in the literature.

A summary of the algorithms and the most relevant components of the feature vectors of the works that use machine learning to address the detection and classification of road anomalies is presented in Table \ref{tbl:lit_rev:ml_methods}. 

\begin{table}[!ht]
	\caption{Summary of characteristics for works employing machine learning algorithms. }
	\label{tbl:lit_rev:ml_methods}
	\centering
\resizebox{\columnwidth}{!}{%
	\begin{tabular}{ llllll }
	    \toprule
    	\textbf{Work} & \textbf{Sensors} & \textbf{Axes} & \textbf{Freq. (Hz)} & \textbf{Algorithm} & \textbf{Features} \\
		\midrule		
        \citet{tai2010}       & Acc & Z, X, Y & 25 & SVM & Mean, range, stdev, speed. \\
        \citet{perttunen2011} & Acc & Z, X, Y & 38 & SVM & 95+, time/freq. domains. \\
        \citet{bhoraskar2012} & Acc & Z, X, Y & 50 & SVM & Mean, stdev. \\
        \citet{jain2012}      & Acc & Z, X, Y &    & SVM & Stdev. mean cross, slope. \\
        \citet{cong2013}      & Acc & Z       & 38 & SVM & Wavelet packet decomposition. \\
        \citet{seraj2015}     & Acc, Gyr & Z, X, Y & 47, 93 & SVM& Time/freq. domain, wavelets. \\
        \citet{mohamed2015}   & Acc, Gyr & Z, X & & SVM & Mean, stdev., range. \\
        \citet{tecimer2015}   & Acc, Gyr, Mag & Z, X, Y & 50 & KNN, RBFN, NB, LMT, MLP, SVM & Freq. domain, MFCC.\\
        \citet{brisimi2016}  & Acc & Z, X, Y & 50 & SVM, RF, LR, Adaboost & Stats., freq. domain. \\
        \citet{silva2017}    & Acc & Z, X, Y & 50 & RF, GB, DT, NN, SVM & Stats., integral. \\
        \citet{singh2017}    & Acc & Z & 10 & DTW & Acc. \\
        \citet{gonzalez2017} & Acc & Z & 50 & ANN, SVM, RF, KNN, DT, NB, KR & Bag of words \\
        \citet{carlos2018} & Acc & Z & 50 & SVM, Ensemble & Mean, stdev, range, thresholds. \\        
		\bottomrule 
	\end{tabular}
}%
\end{table}

\subsection{Influence of Speed}

The effects of speed in the detection of road anomalies are frequently mentioned in the literature. The simple cases are using the speed reported by GPS sensors to decide weather or not the vehicle is in movement, allowing to avoid false positives caused by passengers getting inside the vehicle or slamming doors \citep{eriksson2008}, or assuming that vehicles slow down on roads in bad condition \citep{forslof2015}. However, more serious implications of vehicle speed have been reported: anomalies can be missed at low speeds \citep{mednis2011} and small anomalies can appear to be more significant than they really are at high speeds because of high readings of vertical acceleration \citep{eriksson2008, astarita2012, yi2015}, leading to proposals that use different detection strategies \citep{mohan2008} or take speed into account in the calculation or processing of features \citep{fazeen2012, yagi2010, sinharay2013, harikrishnan2017}. 

Attempts have been made to indirectly address speed dependency, by applying measures that don't directly modify the detection strategy. \citet{seraj2015} assumed the effect of speed is present in acceleration signals as amplitude modulation, and presented a filter to remove it and improve anomaly detection. \citet{perttunen2011} modeled the relationship between speed and inertial data as a trend in the value of calculated features and attempted to approximate that trend with linear regression, and then subtracting that trend from the features employed for classification. \citet{orhan2013} used time windows of varying lengths to analyze acceleration, with the length being reduced as the vehicles go faster. 

\section{Road anomaly profiling}

Another possible path for the classification of road anomalies is classifying them by their dimensions. For example, when describing a pothole, we can focus on its depth, length, and width, and these more precise characteristics could allow us to consider it of more or less relevance for drivers and authorities: drivers could receive advanced warnings while driving, while transportation agencies would get more detailed information about road sections in need of repair. The same logic can be applied for speed bumps, to analyze their height and determine if they are still working as effective speed reducers, or if they need to be replaced or repaired. Few examples of this task have been reported in the literature.

\citet{fazeen2012} mention the estimation of bump height by using physics equations dealing with displacement, acceleration, and time, reporting that different estimates are made at different speeds. They applied a dynamic weight to improve their results, reporting good estimates at 32 km/h. However, they also report that their method becomes unreliable at low speeds, and that the way in which the vehicle approaches a bump also affects their results. They hypothesized this method can be used to estimate pothole depth. However, their characterization of anomalies was only explored with one example.

A method for pothole depth estimation based on the double integration of vertical acceleration is described in \citet{harikrishnan2017}. They consider the five vertical acceleration samples neighboring the peak that is associated with an anomaly. This method was evaluated on one pothole and one speed bump, and it was found that speed had a great impact in their predictions: the best estimations were made between 15 and 20 km/h, and lower or higher speeds quickly increased the error. 

In \citet{xue2017}, a one degree of freedom vibration model is used to predict the vertical displacement of the wheel. They assume that the profile of a pothole can be recovered by modeling the real vibrations of the wheel that falls into it from acceleration data collected with a smartphone. To solve the equations, information about the vehicle and the placement of the smartphone are required, along the captured data. The issue of GPS error and latency is addressed by calculating the time between when the front and rear wheels pass over the anomaly. They evaluate their proposal over a data set collected from 23 different potholes (with 2,760 time series in total), considering different vehicles, speeds, and sensor placements, achieving a relative error of 15\% for pothole depth, and by means of aggregation manage to reduce it to 13\%.

\section{Detection of Aggressive Driving Maneuvers}

Driver safety has been an important topic for ITS, in particular for insurance telematics, because of the link between aggressive driving maneuvers and traffic accidents \citep{klauer2009, osafune2016}. While speeding and harsh braking are commonly considered as dangerous driving, frequent or abrupt lane changing have been associated with aggressive driving, accounting for up to 10\% of all vehicle crashes \citep{lee2004}.

Speeding, the most obvious form of aggressive driving, is easily detected from the speed estimates reported by a GPS and will not be treated in this text. Other aggressive driving actions have been mostly analyzed through acceleration readings in the lateral and longitudinal axes of the vehicles, but magnetometers \citep{castignani2015} and gyroscopes have also been considered \citep{johnson2011, seraj2015}. 

As with road anomaly detection, threshold detectors have been common for the automatic detection of aggressive driving events \citep{dai2010, eren2012, fazeen2012, eboli2016, eboli2017}. The procedures, some times called end-point detection, are similar to those used to detect road roughness: a metric, representative of high energy events, is calculated from acceleration and then a comparison is made with respect to a threshold value to decide if the event can be considered as aggressive. An advanced form of these methods is developed in \cite{vlahogianni2017}, by employing the MODLEM algorithm to find the minimal set of decision rules that maximizes the discriminative power to detect harsh acceleration, braking, and cornering.

More refined analysis has been performed to identify the type of aggressive event, and not just the general occurrence of aggressive driving, by comparing windows of acceleration with previously known examples of maneuvers such as cornering, swerving, braking, accelerating, etc. \citep{saiprasert2015, engelbrecht2015}.

Machine learning algorithms have been applied for both the detection and classification of aggressive driving \citep{meseguer2013, vanly2013, hong2014, tecimer2015}, assembling their feature vectors with statistical data extracted over sliding windows, as well as spectral information. Examples of these are  \citet{zylius2017}, where a handcrafted feature vector containing histogram features, correlation coefficients, data threshold validation, jerk profile, and spectral information, to discriminate between safe and aggressive driving; \citet{predic2015}, where  feature vectors containing statistical and signal processing metrics were created to detect lane changes, obstacle avoidance, and harsh braking, with decision trees. \citet{bejani2018} made an uncommon proposal, using an ensemble integrated by a decision tree, a multilayer perceptron, a support vector machine and a k-nearest neighbor classifier to detect dangerous and safe maneuver. 

Once aggressive driving maneuvers are identified, it is possible to assign a score to a driver to quantify the level of aggressiveness displayed when driving. Such a task was addressed in \citet{lopez2018} by applying genetic programming to produce a scoring function, based on the frequency of different types of aggressive maneuvers and the relevance of each type of event.

\section{Sensor Reorientation}

Since smartphones are used as sensing devices for many ITS studies, the orientation of sensor readings is a very important issue. The simplest approach to solve the disorientation of sensor readings consists on fixing the smartphones to the vehicles in some way, guaranteeing that the sensors' axes will align to those of the vehicles and that data will be referenced in a known convention, but also ensuring that alignment will not change while data is being acquired. Examples of this are using a holster \citep{fazeen2012}, applying adhesive tape to the dashboard \citep{douangphachanh2013}, and fixing the device to the floor of the vehicle \citep{martinez2014}. 

Given the intended purpose and normal usage patterns of mobile phones, it is not possible to assume that smartphone sensors will be aligned to the vehicle, or that their orientation will remain constant. It is also not reasonable to expect that end-users of smartphone-based ITS applications will invariably take action to ensure their devices are placed in a fixed and known orientation. The alternative is assuming the orientation of the devices will have to be corrected, employing one of the procedures described in Section \ref{rotations}. 

The usage of the Euclidean norm of triaxial acceleration has been reported in \citet{jain2012}, \citet{tecimer2015}, and \citet{sebestyen2015}. Partial reorientation, extracting acceleration in the vertical axis, was used in \citet{astarita2012}, \citet{orhan2013}, \citet{astarita2014}, and \citet{singh2017}. This approach is useful when detecting road anomalies, given that most of the information comes from the vertical axis, but it is not useful for the detection of driving events that are mostly reflected in the other axes. In such cases, triaxial reorientation is required. The estimation of two angles for vertical reorientation can be made with trigonometric equations employing acceleration values. However, the third angle is more complicated, and different approaches have been applied to estimate it.

\citet{mohan2008} tried solving an equation to find the direction in which force is detected when the GPS reports the vehicle is drastically reducing its speed. \citet{promwongsa2014} and \citet{vlahogianni2017} estimated the missing rotation angle as the difference between the direction of travel reported by the GPS and the angle towards true north obtained from the magnetometer. 

\citet{bhoraskar2012} employed Android's API to reorient smartphone data first into geographic frame of reference (that is, as if the device's Y axis pointed to the geographic north and the X axis was aligned west to east, with the Z axis pointing corresponding to the vertical), and then use the bearing provided by the GPS to align the readings with the vehicle's axes. A similar approach is proposed in \citet{alasaadi2016}, but using Principal Component Analysis (PCA) to find the angle that best separates longitudinal and lateral acceleration instead of finding the third rotation angle by magnetic means. 

In some works the reorientation procedure is described in terms of Euler angles, but no mention is made about the estimation of the required angles \citep{wang2015}.

%% file: chapters/reorientation.tex
\chapter{Virtual Reorientation of Smartphone Sensors}
\label{reorientation}

\section{Motivation}

Most ITS works assume knowledge of the orientation of the sensors used to acquire data, and this becomes problematic when smartphones are considered as the main sensing platform, given that their location inside the vehicle is highly variable under real-life conditions. 

Since the axes of a smartphone used to perform opportunistic sensing usually do not align with those of the vehicle in which it travels, it is necessary to correct their orientation by means of a linear transformation. Most works just assume data to be in a known orientation, and only a few  deal with the specifics of how this transformation is performed. Virtual reorientation refers to a method capable of producing readings that reflect how a device would have experimented events if its axes were properly aligned with those of the vehicle, independently of their real alignment.

Although different approaches have been proposed for the reorientation of sensor data obtained from smartphones, it is not clear from the literature if the problem can be considered as solved, if all the proposals are equivalent, or what kind of error is to be expected when performing these procedures. As mentioned before, noise, sensor response, and an unpredictable environment are the norm for opportunistic sensing.

This section presents experimental work performed to investigate different reorientation strategies used in smartphone-based sensing for ITS applications, focusing on accelerometer data but with results that apply to all on-board inertial sensors, given that axes are shared among them. Four different proposals for the extraction of vertical acceleration were evaluated, and the estimation of yaw (the angle of rotation around the axis perpendicular to the floor) required for triaxial reorientation was attempted with two different methodologies. 

\section{Aims and Objectives}

The main objective of this chapter is the evaluation of the sensor reorientation strategies described in the literature, in order to determine if they are a reliable alternative when data is acquired by means of opportunistic sensing. Reorientation methods are judged by the similarity between signals captured with sensors known to be aligned with the vehicle, and reoriented signals that were obtained from unaligned ones.

The secondary objectives are the comparison of the different reorientation strategies, to determine if all are equally functional or if one should be chosen over the rest, and an estimation of the error to be expected when these transformations are applied. 


\section{Methodology}

\subsection{Data}

Up to six smartphones were placed inside a car to record accelerometer, magnetometer, and gyroscope readings with a sampling frequency of 50 Hz. Two devices were oriented with respect to the car's axes (reflecting ground truth), and fixed to a dashboard holder and the console between the frontal seats. The other devices were placed in the vehicle's door, a dashboard compartment, and in pockets found in the trousers and shirt of the driver, with each of these smartphones having at least one axis with improper alignment with respect to the vehicle. Except for the devices considered as ground truth, all smartphones were placed freely inside the vehicle, without being fixed to any surface. The locations and freedom of movement of the devices were chosen to try to replicate normal usage conditions for everyday activities.

Multiple locations inside the vehicle were used because it was found in preliminary observations that the frequency and influence of noise was dependant on the placement of the smartphones. For example, high frequency noise is not found in acceleration signals captured when the devices are placed in the driver's pockets, but is very noticeable when data was captured from the door compartments. Magnetometers are highly affected by the electrical systems of the cars, and the different device placements might result in different levels of noise induction. 

GPS data was captured at 1 Hz in all devices, including location, direction of travel (bearing), and speed. The location of a device inside the vehicle could have some effect on GPS reception, in addition to the known problems caused by the topography and the buildings in the area.

Two groups of signals captured in the manner described above were used in our experiments. The first consists of sixty short time series reflecting the acceleration data captured when the vehicle passed over asphalt speed bumps. The waveform for these events is very distinctive, and simplifies evaluating the extraction of vertical acceleration. Twelve time series were considered ground truth, with data properly oriented, and forty-eight were known to be disoriented. The average length for the time series was 270 data points (about 5.4 s). 

The second group of signals, used to evaluate azimuth estimation, consists of twenty-five time series and reflects thirteen minutes of driving five times over a 4.8 km predefined route. One device was fixed to a holder in the dashboard, the other smartphones were placed in the left door compartment, a dashboard compartment, the driver's trousers, and on the copilot's seat. All devices were placed so that their longitudinal axes (Y) were aligned with the vehicle's.

\begin{figure}[!ht]
    \centering
    \includegraphics[width=14cm]{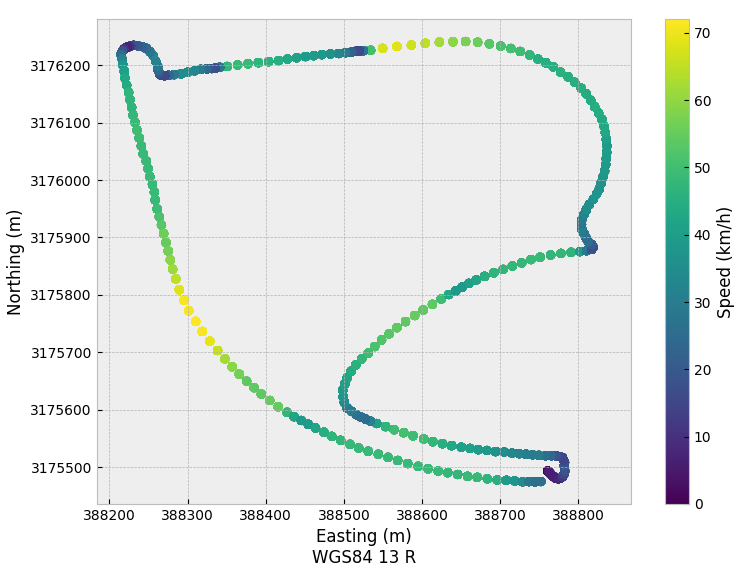}
    \caption{Trajectory and speed of the vehicle for the third data set, in UTM coordinates.}
    \label{fig:reo:uach_circuit}    
\end{figure}

A third data set was assembled to specifically address the problem of rotation around the vertical axis when triaxial reorientation is performed. Four smartphones were firmly fixed inside a car, two were attached to the dashboard and had their longitudinal axes oriented with respect to the vehicle's, and two more were attached to surfaces in the console and the right door with a disorientation of -90º (cw) and approximately 135º (cw) with respect to the vehicle's direction of travel. Data was collected simultaneously in all four devices (with the same sampling frequencies as in the previous data sets) over an approximate distance of 2.79 km, at speeds between zero and eighty km/h (38.7 km/h avg.). There was little to no GPS signal obstruction or reflection during data collection. The circuit had a difference of about 33 m between the lowest and highest elevation points, and contained four tight turns. The trajectory of the vehicle is shown in Figure \ref{fig:reo:uach_circuit}.

\subsection{Algorithms}

The evaluation of the results of vertical and triaxial reorientation strategies was done by calculating the Normalized Cross-Correlation (NCC) and the Dynamic Time Warping (DTW) distance between ground truth and the reoriented versions of the acceleration signals that were captured in unknown orientations. The NCC score goes from -1 to 1, with 1 being the best outcome for this test, 0 meaning signals are not correlated, and a negative value indicating inverse correlation. Zero or low DTW coefficients indicate a high similarity for the time series when compared with ground truth. 

All acceleration time series were filtered before performing the tests, in order to remove obvious noise and extract the frequency components that are relevant for road anomaly and aggressive driving detection. A simple moving average filter (with a sliding window of ten elements) was used, and then a low-pass filter ($\alpha$ = 0.4) was applied. This filter removes individual random extreme values and most of the high frequency noise produced by the vibrations of the engine.

The acceleration signals were also centered to zero mean and scaled by standard deviation, trying to standardize them and reduce variations caused by the uncalibrated nature of the sensors. This process also mitigates the influence of the noise experienced by the devices that were not firmly fixed.

Simultaneous acceleration signals were manually aligned to reduce the effect of temporal shifting. Because of latencies from the local area network established with the smartphones, the operating system, and the data capture application, simultaneous signals can be significantly shifted (from 20 milliseconds to about a second).

\section{Experiments and Results}

\subsection{Vertical Reorientation}

Four proposals for the extraction of vertical reorientation are considered: those by \citet{jain2012}, \citet{tundo2013}, \citet{astarita2012}, and \citet{promwongsa2014}. In the first, the Euclidean norm of triaxial readings is considered as an alternative to reorientation, given that it is very close to vertical acceleration. The other three rely on finding two angles (roll and pitch) to perform rotations around the X and Y axes of the smartphones sensors. Quaternions are used to obtain a rotation matrix in  \citep{tundo2013}, while Euler angles are used in \citet{astarita2012} and \citet{promwongsa2014}. 

\begin{figure}[!ht]
    \centering
    \includegraphics[width=14cm]{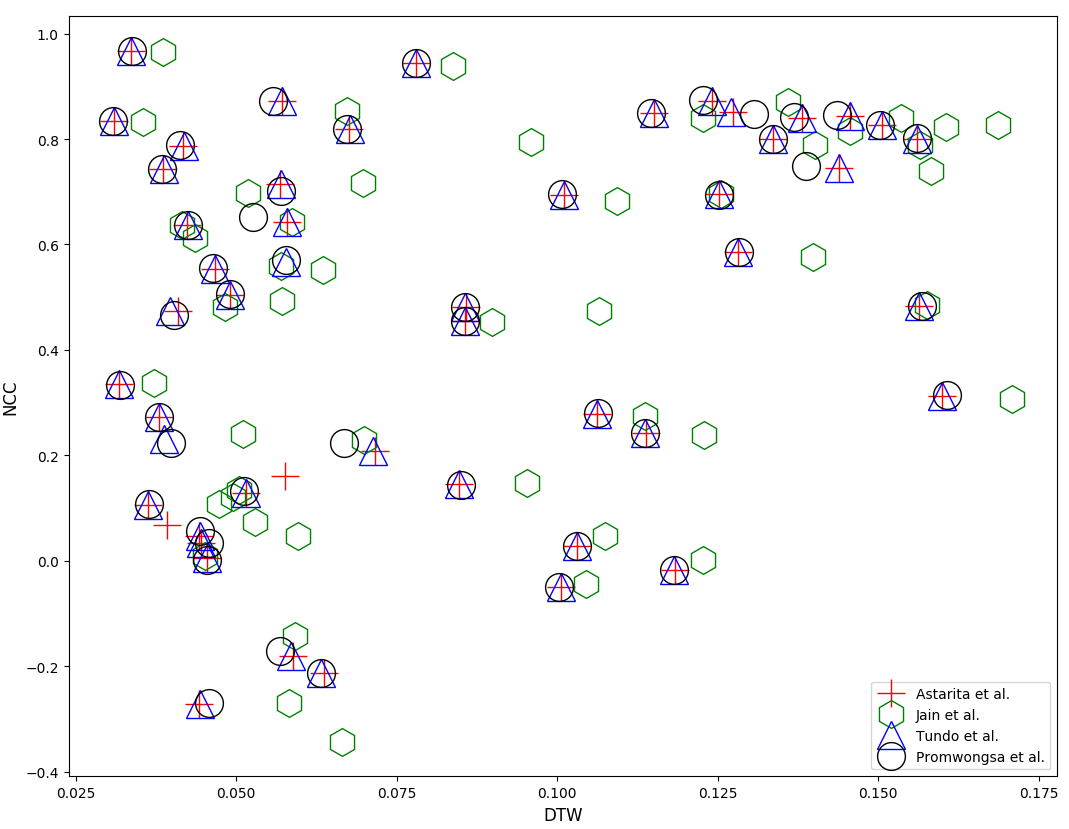}
    \caption{NCC and DTW by method for vertical reorientation.}    
    \label{fig:reo:ncc_dtw_vertical}    
\end{figure}

The effectiveness of vertical reorientation (i.e. finding the roll and pitch angles to rotate readings around the X and Y axes of the smartphone as to match the vehicle's) was tested by performing the four procedures previously described over data from the first group of acceleration time series. After that, NCC and DTW were calculated between pairs of simultaneous signals: a time series reflecting ground truth and the output of the reorientation procedure. The results are plotted in Figure \ref{fig:reo:ncc_dtw_vertical}.

In general, the low DTW distances (less than 0.18) between reoriented signals and ground truth suggests the procedures are effective and that vertical acceleration signals captured with smartphones can be transformed to match those experimented by the vehicle. It can also be seen that the different procedures produce almost identical results over the same time series.

\begin{figure}[!ht]
    \subfigure[]{
        \includegraphics[width=7cm]{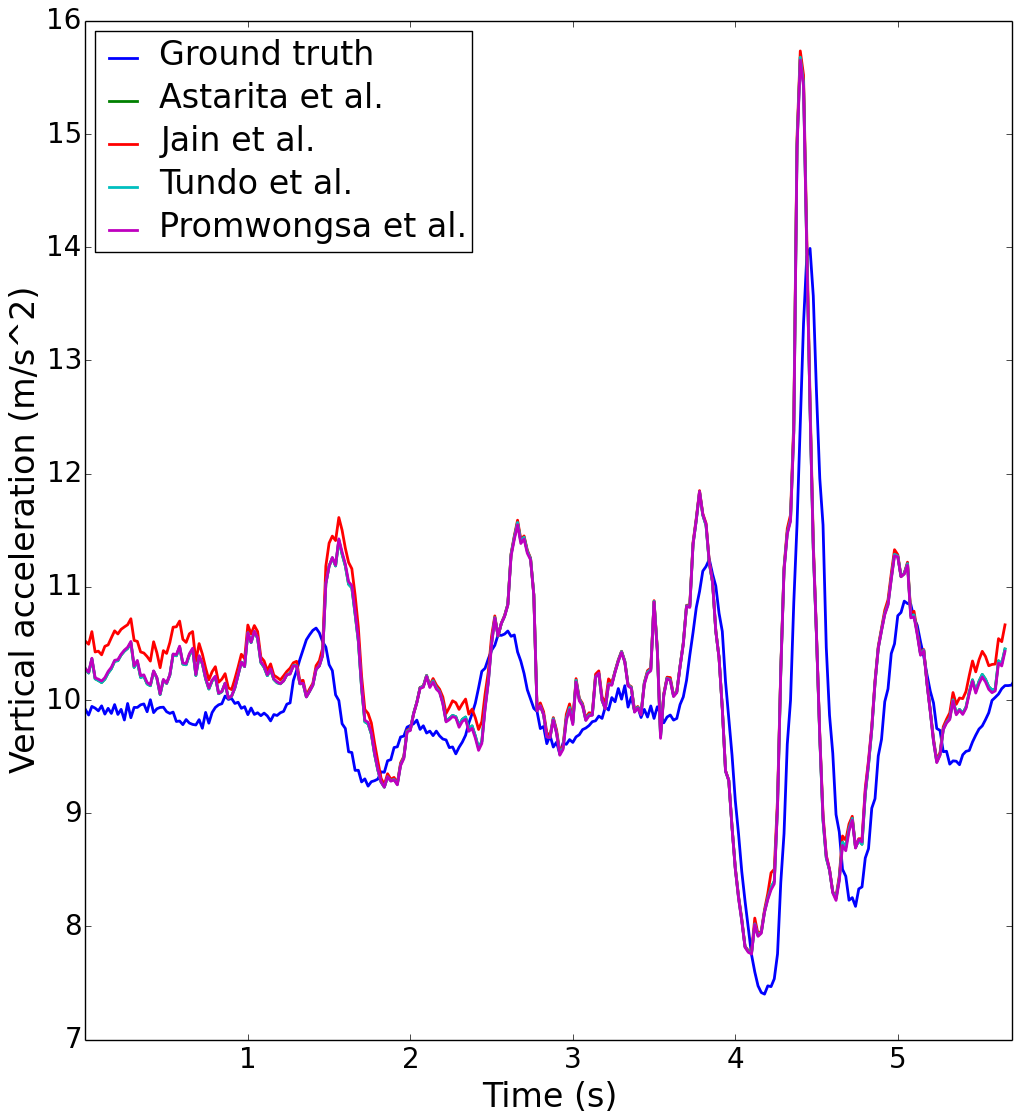}
    }
    \hfill
    \subfigure[]{
        \includegraphics[width=7cm]{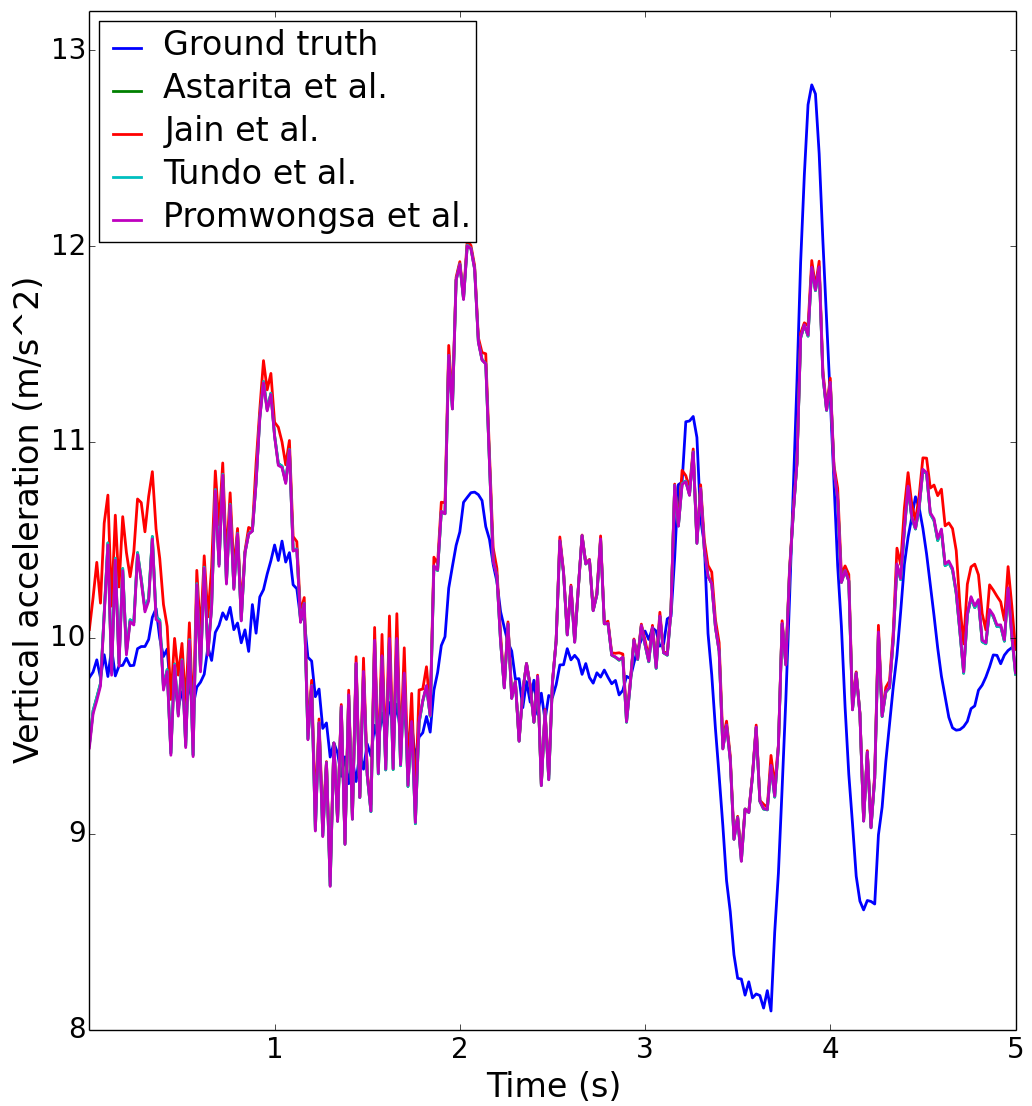}
    }
    \caption[Examples of the worst cases for vertical reorientation]{A) Reorientation results with NCC $\approx$ -0.3 and DTW $\approx$ 0.042. B) Reorientation results with NCC $\approx$ 0.25 and DTW $\approx$ 0.16. Raw reoriented data for both examples. }
    \label{fig:reo:worst_vertical}  
\end{figure}

A one-way ANOVA test ($\alpha$ = 0.01) was applied to the NCC and DTW scores for the four methods, resulting in no statistically significant differences among them. Visual inspection of the results confirms all three reorientation alternatives and the Euclidean norm produce results very similar to ground truth, even with all the noise and uncertainty associated with the sensors and data employed. Figure \ref{fig:reo:worst_vertical} shows the reoriented signals and ground truth for the two  groups of time series with the biggest differences in terms of NCC and DTW. Even for these extreme cases it can be seen that the reorientation is effective, and that all four alternatives produce very similar results.

It can be concluded from this analysis that the evaluated vertical reorientation proposals, and the usage of the Euclidean norm as a substitute for vertical reorientation, are suitable for ITS applications, and that their results can be considered as equivalent when analyzing vertical acceleration. 

\subsection{Azimuth Estimation}

Azimuth estimation consists on fusing data from magnetometer and accelerometer to have a compass \citep{ozyagcilar2015} and obtain the angle between the smartphone and the magnetic north (azimuth). This angle can then be compared with the direction of travel (bearing), obtained from GPS, to get information about the orientation of the smartphone with respect to the vehicle in which it travels \citep{promwongsa2014}. 

The stability of azimuth estimation was evaluated before testing the reorientation approach that relies on it, because of the electromagnetic noise created by the electrical systems of a car and the known sensibility of smartphone magnetometers to environmental magnetic disturbances \citep{kang2012}. 

It has to be noted that the official Android API documentation states the following about the estimation of rotation matrices with the getRotationMatrix method: "If the device is accelerating, or placed into a strong magnetic field, the returned matrices may be inaccurate" \citep{androidrotation}. It is advisable to take into account the magnetic declination for the zone, as it can be significant in some locations. 



The second group of signals described above was used for azimuth estimation. The direction of travel reported by the GPS was considered as ground truth, given that it is the most reliable estimate available for the angle of the device with respect to true north with the available equipment. Because of the length of the time series, DTW was not calculated and only NCC was used as a metric of similarity. Table \ref{tbl:reo:ncc_azimuth_bearing} shows the NCC between bearing and azimuth for each device in each lap around the circuit. Bearing and azimuth are expected to be very highly correlated, since the longitudinal axis of each device is aligned with the vehicle's.

\begin{table}[!ht]
  \caption{Normalized Cross-Correlation between  azimuth and bearing at each smartphone location.}
  \label{tbl:reo:ncc_azimuth_bearing}
  \centering
  \begin{tabular}{c lllll}
    \toprule
    	Lap & Loc. 1 & Loc. 2 & Loc. 3 & Loc. 4 & Loc. 5 \\
    \midrule        
          1 & 0.980  & 0.051  & 0.966  & 0.876  & 0.737  \\
          2 & 0.965  & 0.206  & 0.766  & 0.627  & 0.721  \\
          3 & 0.969  & 0.504  & 0.948  & 0.896  & 0.453  \\
          4 & 0.965  & 0.206  & 0.972  & 0.889  & 0.864  \\
          5 & 0.946  & -0.619 & 0.460  & 0.795  & 0.846  \\
    \bottomrule
  \end{tabular}
\end{table}

NCC shows that both bearing and the estimated azimuth are very similar, except for the smartphone placed in the left door. Hardware malfunction was assumed as the cause for the discrepancies; no further experimentation was done to explain that anomaly. A one-way ANOVA test with $\alpha$ = 0.01 showed that there are no statistically significant differences for NCC between locations 1,3,4, and 5. 

\begin{figure}[!ht]
  \centering
  \includegraphics[width=14cm]{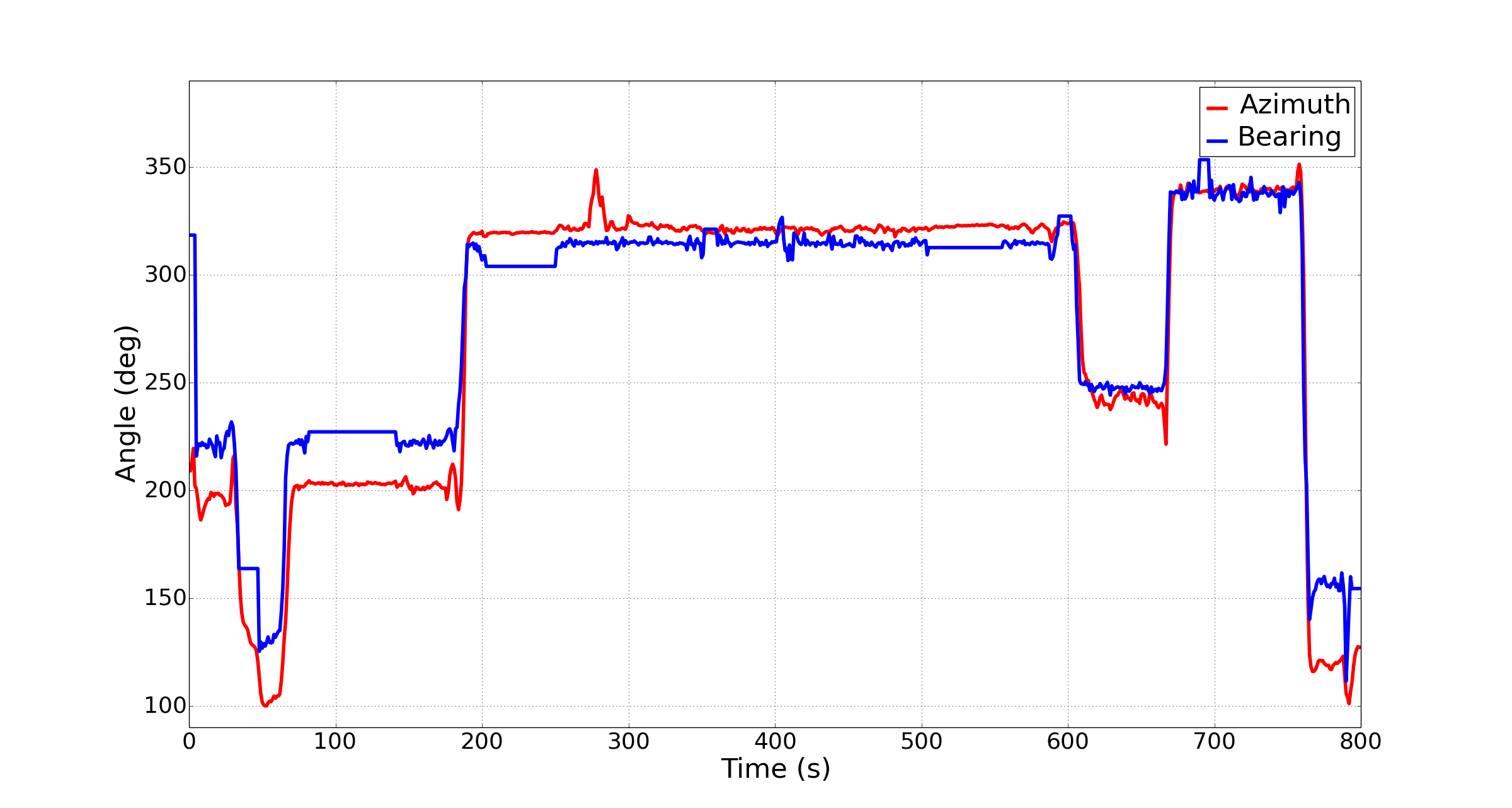}
  \caption{Calculated azimuth and GPS reported bearing} 
    \label{fig:reo:azimuth_bearing}  
\end{figure}

Figure \ref{fig:reo:azimuth_bearing} shows an example of estimated azimuth and GPS reported bearing. Both time series are heavily correlated, but there are sections with noticeable discrepancies.

From these results it can be concluded that azimuth estimation produces acceptable results, but there are frequent momentary discrepancies between azimuth and bearing. This discrepancies can be problematic if reorientation is performed with estimates performed over a short period of time.

\subsection{Triaxial Reorientation}

Two procedures for triaxial reorientation are now compared. These procedures achieve vertical orientation, correcting both pitch and roll as in the previous section, but also deal with yaw estimation to rotate signals around the Z axis. The first method, by \citet{promwongsa2014}, calculates yaw as bearing minus azimuth, while the second solves a differential equation to find the correction that aligns the three-dimensional acceleration vector with the rear of the car when the vehicle experiences a sharp reduction in speed (i.e. braking).

\begin{figure}[!ht]
    \centering
    \includegraphics[width=14cm]{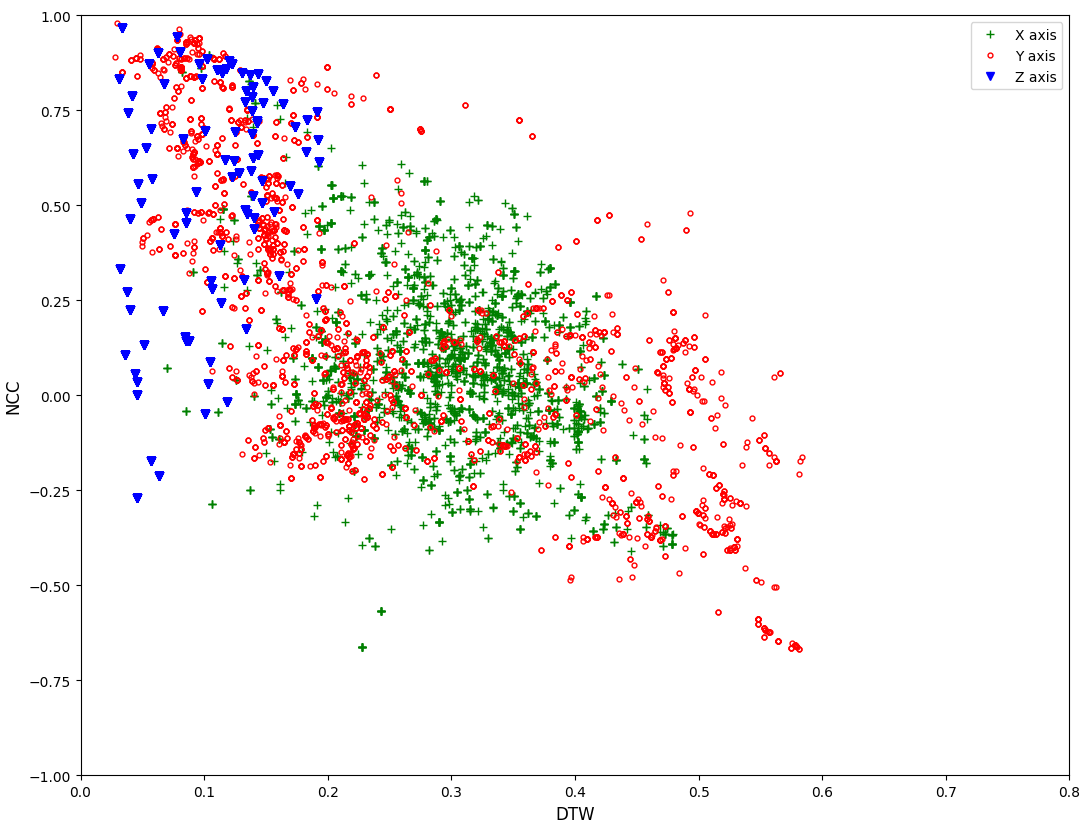}
    \caption{NCC and DTW by axis for compass-GPS based reorientation.}    
    \label{fig:reo:ncc_dtw_mag}  
\end{figure}

\begin{figure}[!ht]
    \centering
    \includegraphics[width=14cm]{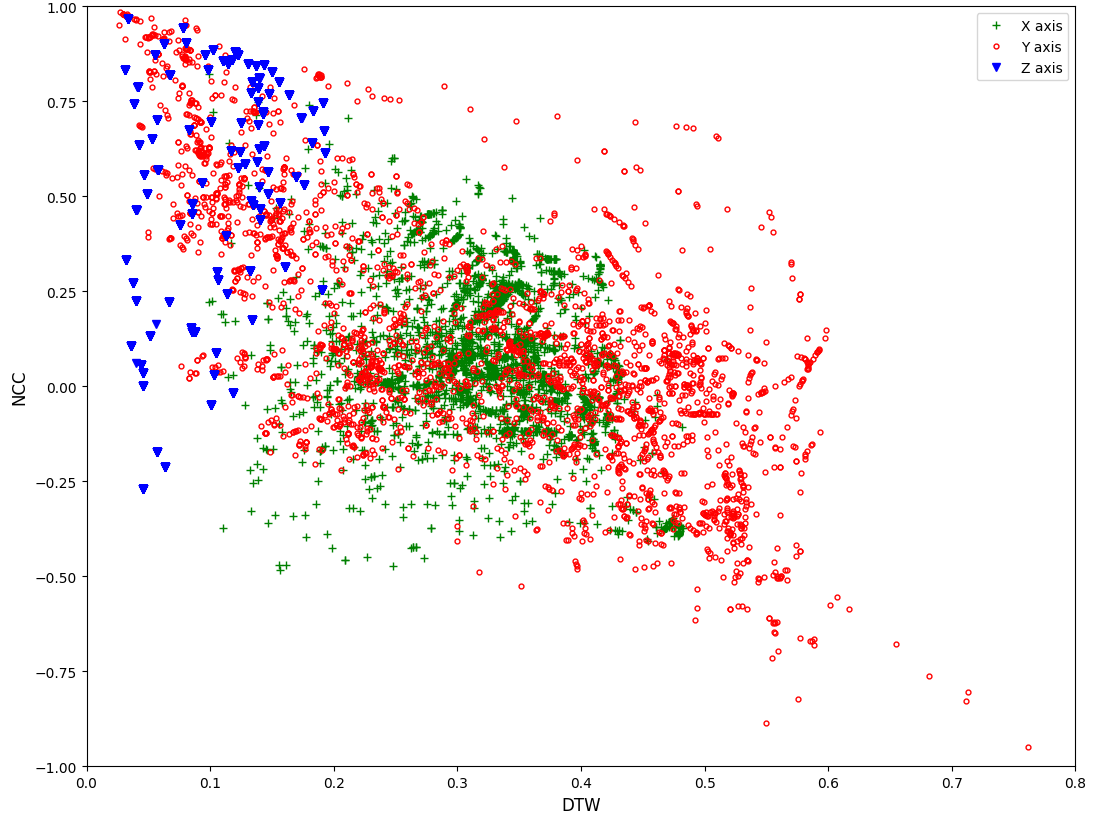}
    \caption{NCC and DTW by axis for accelerometer-GPS based reorientation.}    
    \label{fig:reo:ncc_dtw_acc}    
\end{figure}

For the evaluation of both approaches, a rotation angle was estimated two times per second over a period of one minute, and the resulting angles were used to perform virtual reorientation of acceleration. Given the effects of noise previously observed (see Figure \ref{fig:reo:azimuth_bearing}), a histogram with 15º bins was calculated over the angle determined, and the center of the bin with highest count was chosen as the estimate. For the method described in \citet{mohan2008}, the rotation angles were estimated from the acceleration values in periods of sharp deceleration, found by visual inspection of plots for speed for the data collection session. 
The time series reoriented using both methods were compared to ground truth using the metrics of NCC and DTW. The results of the comparisons using all the yaw estimates are shown in Figures \ref{fig:reo:ncc_dtw_mag} and \ref{fig:reo:ncc_dtw_acc}.

In both plots the best possible score is a point in the upper left corner, and it can be seen that the time series for acceleration in the vertical axis tend to be much closer to that corner than those for the other two axes. This indicates low similarity with respect to ground truth for longitudinal and lateral acceleration, due to incorrect yaw estimates. Dispersion indicates big variations in the estimated rotations, and it can be seen that while vertical reorientation was consistently performed, the reorientation results for the lateral and longitudinal were much less stable.

Knowing that there were errors in yaw estimation, an analysis was performed to quantify the discrepancies. This was done by manually estimating the correct rotation angle, by visual inspection of the signals and taking into account the placement of the devices during data capture, managing to get signals to match ground truth. The previous automatically calculated estimates were compared with the manually obtained yaw angles to quantify the expected error for the two evaluated methods of reorientation. 

It was found that many estimations have errors of more than 20º for both methods. This level of error is significant, because it was empirically observed that it complicates the classification of driving maneuvers. The procedure based on azimuth estimation outperformed the second alternative, having between 20\% and 40\% of the predictions below the $\pm$ 20º threshold for all smartphones.

After these findings, further experimentation was performed to address the error in yaw estimation with bearing and azimuth. Yaw was calculated over the third data set, considering longer series of readings. It was found that for short time series of bearing and azimuth, yaw estimation was again highly inconsistent and in many cases with more than 20º of error; the same for the median values of those predictions. However, the median values of the estimations over the complete length of the time series (about four minutes) were correct in all four devices. Table \ref{tbl:reo:uach_estimation} presents a comparison of the known disorientation angles and the median of the estimations.

\begin{table}[!ht]
  \caption{Median yaw estimations over the 4.3 minute circuit.}
  \label{tbl:reo:uach_estimation}
  \centering
  \begin{tabular}{cccc}
    \toprule
    	\textbf{Device} & Real angle & Median of  all estimations. \\
    \midrule        
          Device 1 & 0º & 6.11º \\
          Device 2 & 0º & 5.08º \\
          Device 3 & $\approx$ 135º (cw)  & -145.71º (cw) \\
          Device 4 & -90º (cw)  & -84.16º (cw) \\
    \bottomrule
  \end{tabular}
\end{table}

The conclusion from this study is that triaxial reorientation of smartphone data should not be considered as trivial, and that precautions must be taken if these alternatives are employed. There is a high probability of significant error for estimations based on data that only reflects brief periods of time. An aggregation of estimates made over longer periods of time will provide more reliable results because of the high levels of noise, low quality and lack of calibration of the sensors, and the uncontrolled nature of opportunistic sensing.

\section{Contributions}

The viability of different proposals in the literature for vertical and triaxial reorientation of data captured from smartphones with opportunistic sensing was tested. While mathematical principles that allow for this transformation have been firmly established for centuries, the viability of their application for ITS applications within the constraints imposed by the available hardware and the typical usage of mobile devices had not been fully studied. These procedures have been treated as trivial in the literature, focusing on the mathematical tools but disregarding the problems of estimating the required angles to perform a correct rotation \citep{vlahogianni2017, wang2015}.

It was found that vertical reorientation can be considered as a solved problem, and that all of the approaches considered in this study are equally effective. It was observed that the usage of the Euclidean norm produces results almost identical to the  reorientation procedures, giving an simpler and computationally less expensive  alternative if only acceleration in the vertical axis is needed.

However, we conclude that that reorientation of longitudinal and lateral acceleration is not a trivial problem. The estimation of the angle required to extract that information is prone to significant errors when it's performed by using the accelerometers, magnetometers, and GPS receiver of mobile devices. Further work is required to refine the procedures to estimate this angle in order to increase the precision of the readings obtained from opportunistic mobile sensing with smartphones over short time periods, and it should be assumed that even if the estimation of this angle can some times produce accurate results, it can frequently have errors larger than $\pm$ 20º. The usage of long periods of observed data is recommended to reduce the error in the reorientation of longitudinal and lateral axes.

The results presented in this chapter were published in: 

\item \textbf{Carlos, M. R.}, González, L. C., Martínez, F., \& Cornejo, R. (2016). Evaluating reorientation strategies for accelerometer data from smartphones for ITS applications. In \textit{Ubiquitous computing and ambient intelligence} (pp. 407-418). Springer International Publishing. doi: 10.1007/978-3-319-48799-1\_45

%% file: chapters/potholes.tex
\chapter{Road Anomaly Detection and Classification}

\label{potholes}

\section{Motivation}

As previously mentioned, a very significant part of the work on the detection and classification of road anomalies during the last ten years has mainly depended threshold-based methods. Those detection methods are simple, of low computational cost, and their decision logic is easily explainable. However, they have the inherent problems of struggling to balance the number of false-positives and false-negatives, and being used in vehicles with different characteristics without a big impact in their performance.

The three most frequently cited articles on roughness detection, \citet{mohan2008,eriksson2008,mednis2011}, have over 2,500 citations combined as of July 2019, according to Google Scholar, and are still not just being cited but influencing new works on ITS. They can be considered very representative of threshold-based methodologies for road anomaly detection, and the general problems of the literature on that topic.

Some road anomaly detection proposals have been based on Machine Learning techniques: using statistical measures, signal processing techniques, and analysis in both the time and frequency domains to form a feature vector that is classified, most of the time, with a support vector machine. Representative works that use these techniques are \citet{perttunen2011} and \citet{seraj2015}, where many different features are calculated from time windows.

Analyzing the literature on road anomaly detection with either approach is problematic because of the heterogeneity in the objectives, methodologies, and non-standard data sets. It is difficult to compare if one way to set a threshold level is more effective than others, or if some features used in one work have more discriminant power than those proposed elsewhere. It is difficult to replicate the results because of the lack of public data sets, and attempting to collect data sets that match those described in the literature is extremely difficult.

This chapter presents a comprehensive comparison of the most representative approaches for threshold-based road anomaly detection and a methodology based on machine learning techniques that uses features commonly found in the literature. This comparison aims to put in perspective the previously reported results, contextualizing the performance of the methodologies that are still influencing new research. 

The fusion of detectors has not been reported in the literature. That is, detectors are either based on a threshold or on machine learning techniques, but both approaches have not been combined. Even if authors have extended detectors previously discussed, as in \citet{mohan2008}, their final pipeline consists of one detector. Both of these ideas are addressed with two proposals: a binary classifier that takes as input a feature vector of statistical descriptors and information from threshold application, and an ensemble of detectors.

\section{Aims and Objectives}

The work presented in this chapter has two objectives. The first is to address the lack of clarity with respect to the effectiveness of the most common alternatives in road anomaly detection based on vertical acceleration, establishing which of the most representative and influential proposals in the literature produce the best results. This is done by applying the different methodologies to one robust data set, addressing the same tasks, and evaluating the results under the same metrics. 

The second objective is to study the techniques discussed above to create detectors more effective than those that rely on just one principle. This is done in two ways: with a binary classifier that combines statistical descriptors with information obtained from thresholds to determine if a road anomaly is present in a vertical acceleration time signal, and with an ensemble that decides with a majority vote if the acceleration signal reflects road in normal condition of if an anomaly has been passed.

\section{Methodology}

\subsection{Data}

\begin{figure}[!ht]
    \subfigure[]{\includegraphics[width=6cm]
    {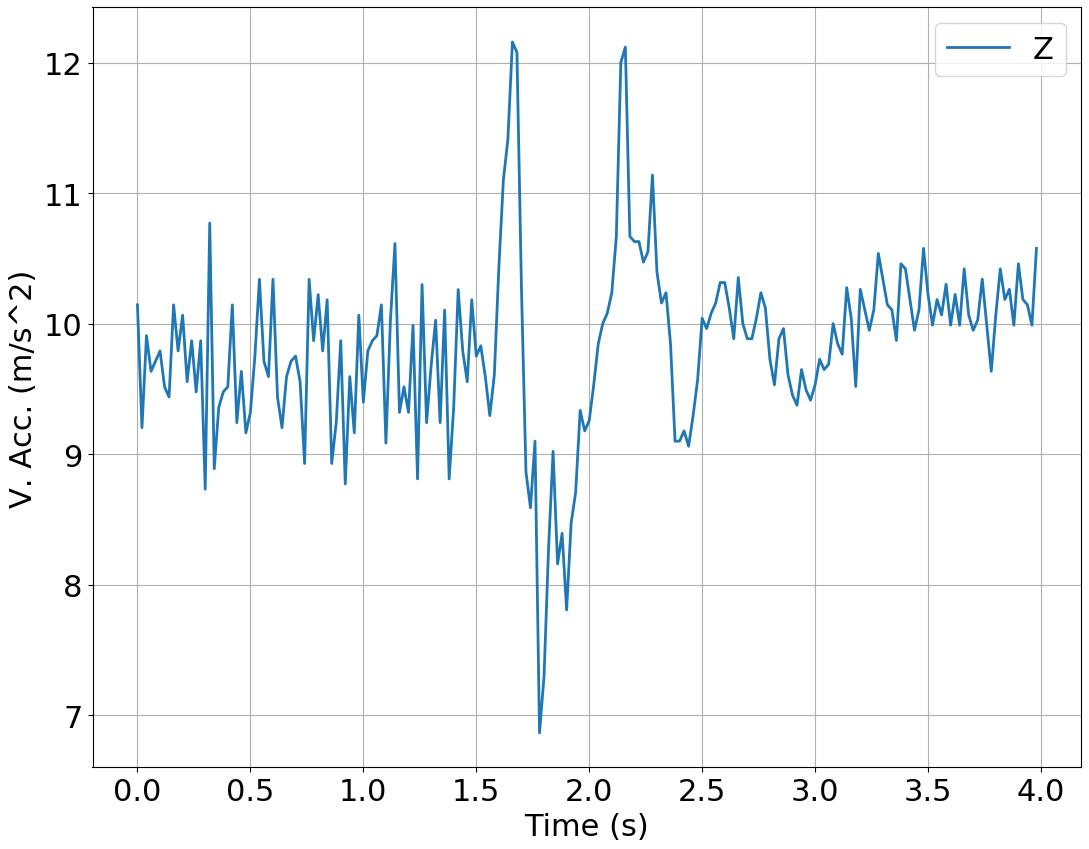}}
    \hfill
    \subfigure[]{\includegraphics[width=6cm]
    {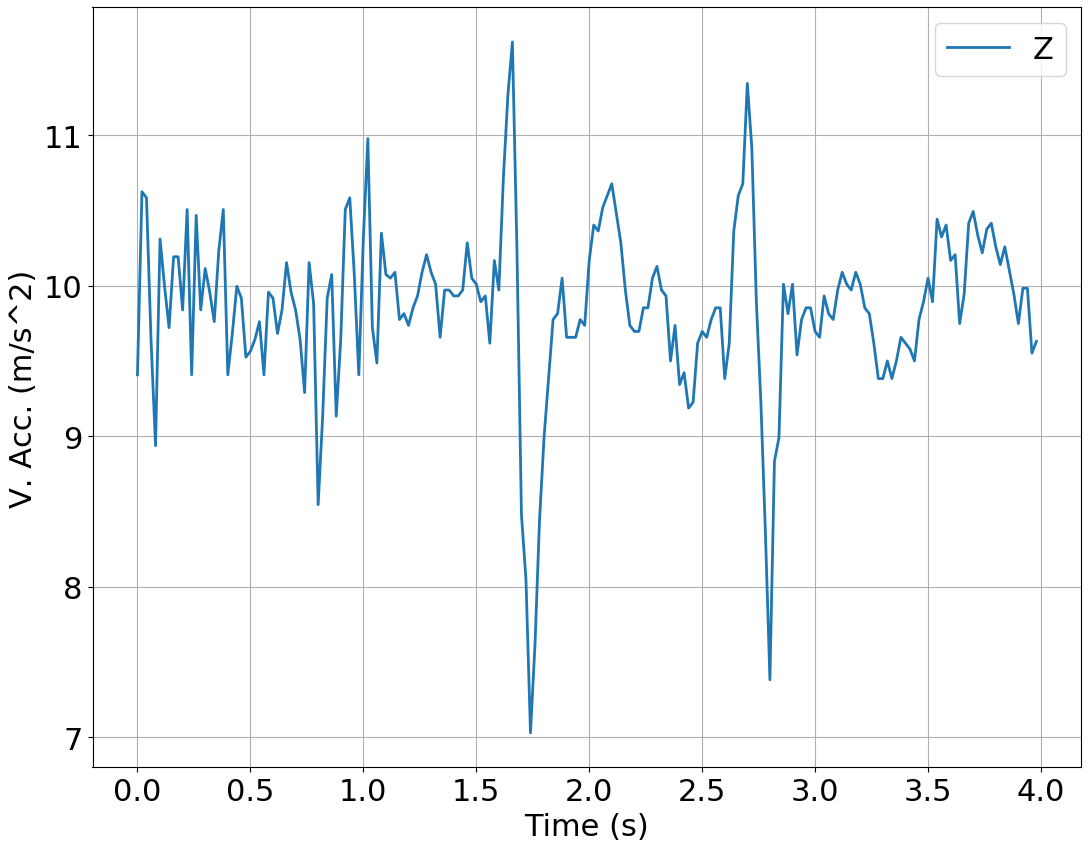}}
    \hfill
    \subfigure[]{\includegraphics[width=6cm]
    {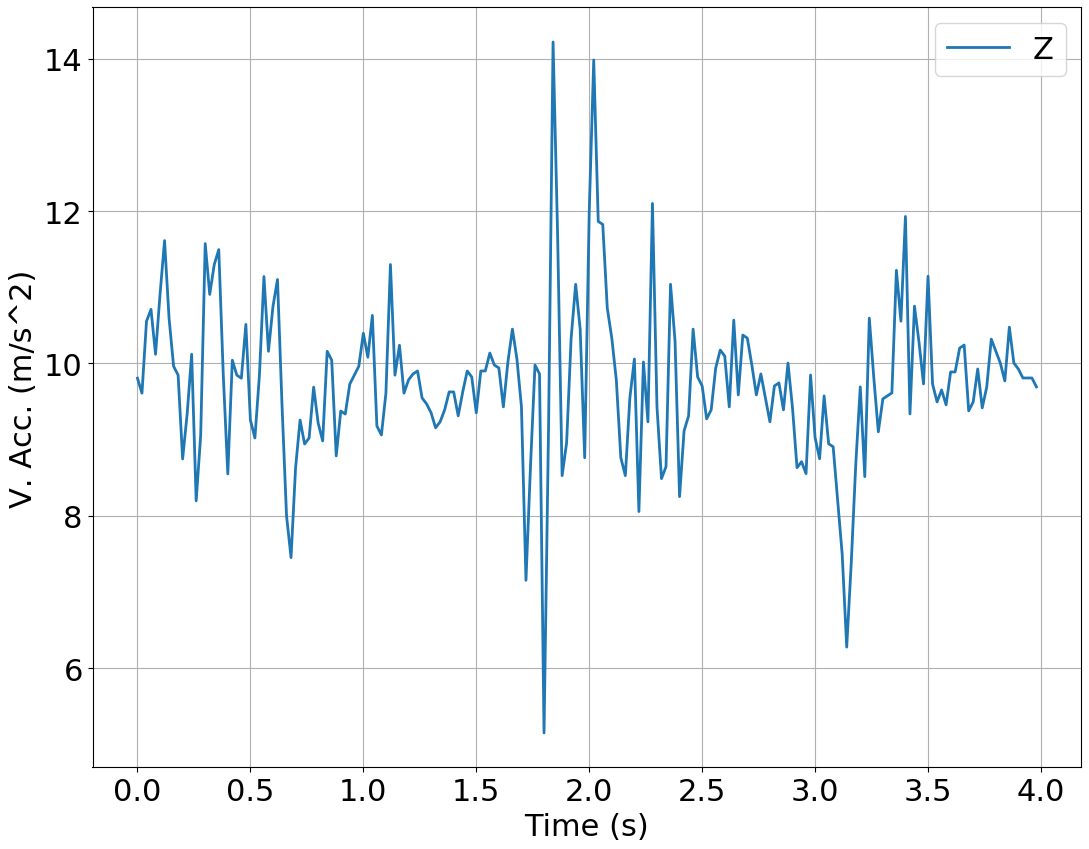}}    
    \hfill    
    \subfigure[]{\includegraphics[width=6cm]
    {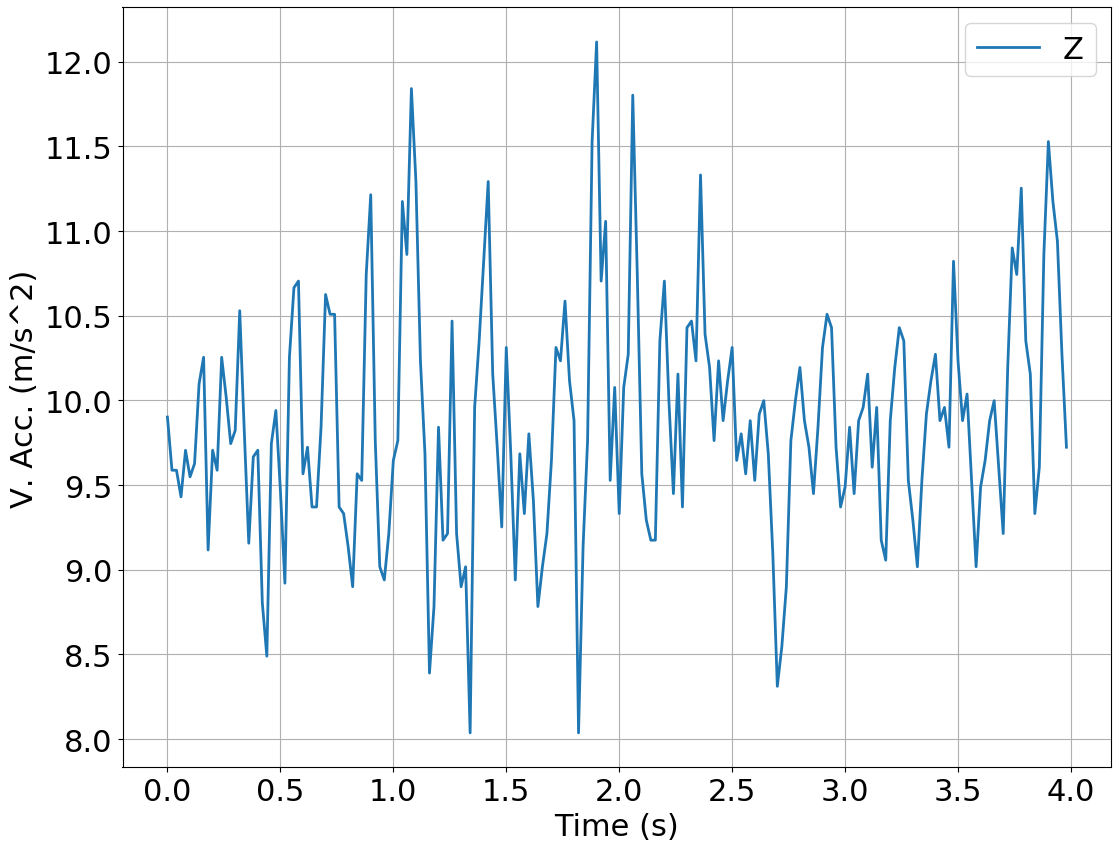}}        
    \caption[Vertical acceleration examples]{Examples of vertical acceleration signals: A) Speed bump, B) Metal speed bumps, C) Pothole, D) Normal road.
    }\label{fig:whoswho:samples}
\end{figure}

Six different vehicles (including small, medium sized and large models of cars, and pick ups spanning twenty years in their models) were used to collect data by placing smartphones in different locations inside them. Acceleration readings were captured at 50 Hz and minor orientation corrections were applied manually; GPS data was reported at 1 Hz. In total, 237 samples in four categories (detailed in Table \ref{tbl:whoswho:dataset}) were recorded, including normal road segments in good condition and three commonly found types of anomalies, generally discussed as events of interest in the literature: asphalt bumps, metal bumps, and potholes. Examples of acceleration time series for these four types of events are shown in Figure \ref{fig:whoswho:samples}.

\begin{table}[!ht]
	\caption{Number of samples per category for the data set.}
	\label{tbl:whoswho:dataset}
	\centering
	\begin{tabular}{ll}
		\toprule
		Category & Samples\\
		\midrule
        Asphalt Bump & 81\\
        Pothole & 56\\
        Metal bump & 50\\
        Normal road & 50 \\\hline\\
        Total & 237\\
        \bottomrule
	\end{tabular}
\end{table}

Given that the threshold algorithms were proposed to work on long continuous time series that reflected many minutes of driving, and not on short individual ones presenting events lasting a few seconds, an application was created to randomly concatenate a number of these signals, so that long time series with a number of anomalies could be created. Knowing with certainty the time, speed at impact, and type of each anomaly in these synthetic long time series (virtual roads), the aforementioned threshold-based detection strategies can be evaluated under controlled conditions. Sixty time series reflecting "virtual roads" were generated with this application,  thirty used for training and thirty more for testing (listed in Table \ref{tbl:whoswho:roads}). In all cases the virtual roads used for  training contained fewer anomalies than the ones used for testing. 

\begin{table*}[!ht]
  \caption{Virtual roads with the type and number of anomalies present.}
  \label{tbl:whoswho:roads}
  \centering
  \resizebox{\columnwidth}{!}{%
  \begin{tabular}{c c c}
      \toprule
	  Road & Training & Testing\\
      \midrule
	  1  & 16 potholes & 40 potholes\\
	  2  & 20 potholes & 50 potholes\\
	  3  & 30 bumps & 50 bumps\\
	  4  & 20 bumps & 50 bumps\\
	  5  & 20 metal bumps & 30 metal bumps\\
	  6  & 10 metal bumps & 40 metal bumps\\
	  7  & 50 potholes and bumps & 50 potholes and bumps\\
	  8  & 20 potholes and bumps & 80 potholes and bumps\\
	  9  & 20 bumps and 20 metal bumps & 50 bumps and metal bumps\\
	  10 & 20 potholes, 20 bumps and 20 metal bumps & 100 potholes, bumps and metal bumps\\
      11 & 5 bumps & 10 bumps\\
      12 & 5 potholes & 10 potholes\\
      13 & 5 metal bumps & 10 metal bumps\\
      14 & 3 bumps, 3 potholes & 5 bumps, 5 potholes\\
      15 & 3 bumps, 3 metal bumps & 5 bumps, 5 metal bumps\\
      16 & 3 potholes, 3 metal bumps & 5 potholes, 5 metal bumps\\
      17 & 3 bumps, 3 potholes, 3 metal bumps & 5 bumps, 5 potholes, 5 metal bumps\\
      18 & 15 bumps & 25 bumps\\
      19 & 15 potholes & 25 potholes\\
      20 & 15 metal bumps & 25 metal bumps\\
      21 & 8 bumps, 8 potholes & 15 bumps, 15 potholes\\
      22 & 8 bumps, 8 metal bumps & 15 bumps, 15 metal bumps\\
      23 & 8 potholes, 8 metal bumps & 15 potholes, 15 metal bumps\\
      24 & 8 bumps, 8 potholes, 8 metal bumps & 15 bumps, 15 potholes, 15 metal bumps\\
      25 & 30 bumps & 40 bumps\\
      26 & 20 potholes & 25 potholes\\
      27 & 25 metal bumps & 25 metal bumps\\
      28 & 15 bumps, 15 potholes & 20 bumps, 20 potholes\\
      29 & 15 bumps, 12 metal bumps & 20 bumps, 20 metal bumps\\
      30 & 15 potholes, 15 metal bumps & 20 potholes, 20 metal bumps\\
      \bottomrule
  \end{tabular}
  }%
\end{table*}

The acceleration signature for a typical road anomaly has a length of one to four seconds. At 50 Hz, that is from 50 to 200 individual acceleration readings. Since detections can be triggered by one individual reading or a short sequence of them, it is possible to have more than one detection within the duration of an event. In order to avoid multiple detections per event, an anomaly is considered as detected if one or more detections are reported within its duration.

\subsection{Algorithms}

The three threshold-based detectors studied tackle the same problem in similar fashion, but there are notorious differences in their approaches. The one presented in \citet{eriksson2008} (PP) works with vertical and lateral acceleration, feeding high-pass filtered readings to their threshold heuristic. Two different thresholds are applied in \citet{mohan2008} (NER), depending on the speed of the vehicle, to unfiltered vertical acceleration.  \citet{mednis2011} presents four proposals, applying thresholds to the raw signal (ZTH), its first derivative (ZDIF), and the standard deviation of sliding windows (STDEV), and looking for events for which acceleration in all three axes approach zero (GZE). The three works focus on road anomalies in general terms, having pothole detection as the specific common task, but also addressing other different types of road roughness: \citet{mednis2011} mentions pothole clusters and drain pits, \citet{eriksson2008} lists railroad crossings and joints, and \citet{mohan2008} adds speed bumps. 

The previous detectors were implemented, and the thresholds reported by the authors of the evaluated proposals were applied at first, but the algorithms' parameters were fine tuned using the virtual roads assigned for training, looking to maximize the F1 scores. The differences between the original parameters and the ones found in this process are attributable to significant differences between their data and ours, caused by variations in road quality and the response of the suspension systems in the vehicles; threshold heuristics are prone to this significant variability. 

A SVM was proposed to treat anomaly detection as a binary classification problem feeding it with twelve features calculated over sliding windows , taking as reference the work of \citet{perttunen2011} (PERT). The feature vector included five statistical measures (mean, standard deviation, variance, coefficient of variation, and range). Four more values are included, consisting of the difference between the value of the previously mentioned features and a reference threshold, inspired by \citet{mednis2011} both in intention and reference values, followed by the sum of the differences with respect to these thresholds, and the difference with respect to a threshold applied to the sum of confidence values. The number of threshold values that were surpassed for the statistical features (akin to an aggregation of the threshold detectors). The different values for thresholds, and the other parameters evaluated for this detection approach are presented in Tables \ref{tbl:whoswho:svm_thresh} and \ref{tbl:whoswho:svm_params}.

 \begin{table}[!ht]
	\caption{Threshold values for features ($g$ stands for gravity).}
	\label{tbl:whoswho:svm_thresh}
	\centering
	\begin{tabular}{c c }
		\toprule
		Values  & Threshold\\
		\midrule
         Mean & g*0.3\\
         Standard Deviation & g*0.15\\
         Coefficient of Variation & g*0.015\\
         Difference (Max-Min) & g*0.2\\
         Sum of confidence score & 3\\
        \bottomrule
	\end{tabular}
\end{table}

\begin{table}[!ht]
	\caption{Parameters used in the Grid Search process for feature and model selection.}
	\label{tbl:whoswho:svm_params}
	\centering
	\begin{tabular}{c c }
		\toprule
        Parameters & Possibilities \\
		\midrule
		Confidence scores & range (0.1, ..., {\bf 0.8}, 0.9) \\
		C & 1,{\bf 10},15,20 \\
        Kernels & Poly, lineal, sigmoid, and {\bf RBF} \\
        Window size & 10, 20, {\bf 30}, 50, 100\\
       \bottomrule
	\end{tabular}
\end{table}

 The combination of classifiers has been successfully used both in the pattern recognition and machine learning communities, attempting to improve the results obtained by individual classifiers \citep{zhihuazhou2012}. It has been found that a group of weak learners, those that manage to produce results only slightly better than chance, can be combined to collectively perform better than the best individual member \citep{hansen1990}, transforming weak learners into strong ones \citep{schapire1990}. Because of this, a third approach is evaluated: an ensemble of the threshold heuristics ("swarm"). The outcomes of the six threshold detectors were aggregated into one result with a simple majority vote.

\section{Experiments and Results}

\begin{figure}[!ht]
    \subfigure[]{\includegraphics[angle=270,width=7cm]
    {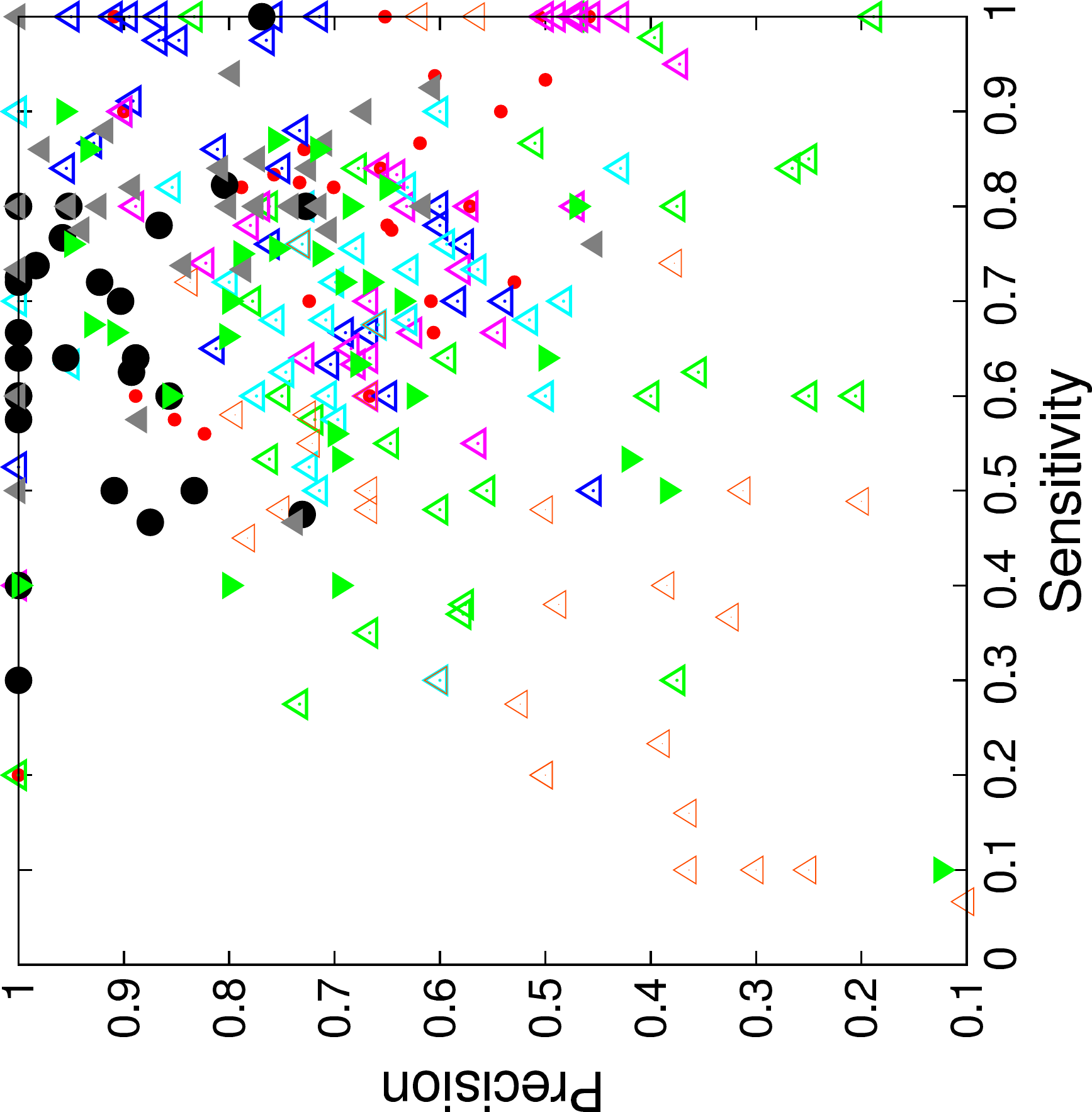}}
    \hfill
    \subfigure[]{\includegraphics[angle=270,width=7cm]
    {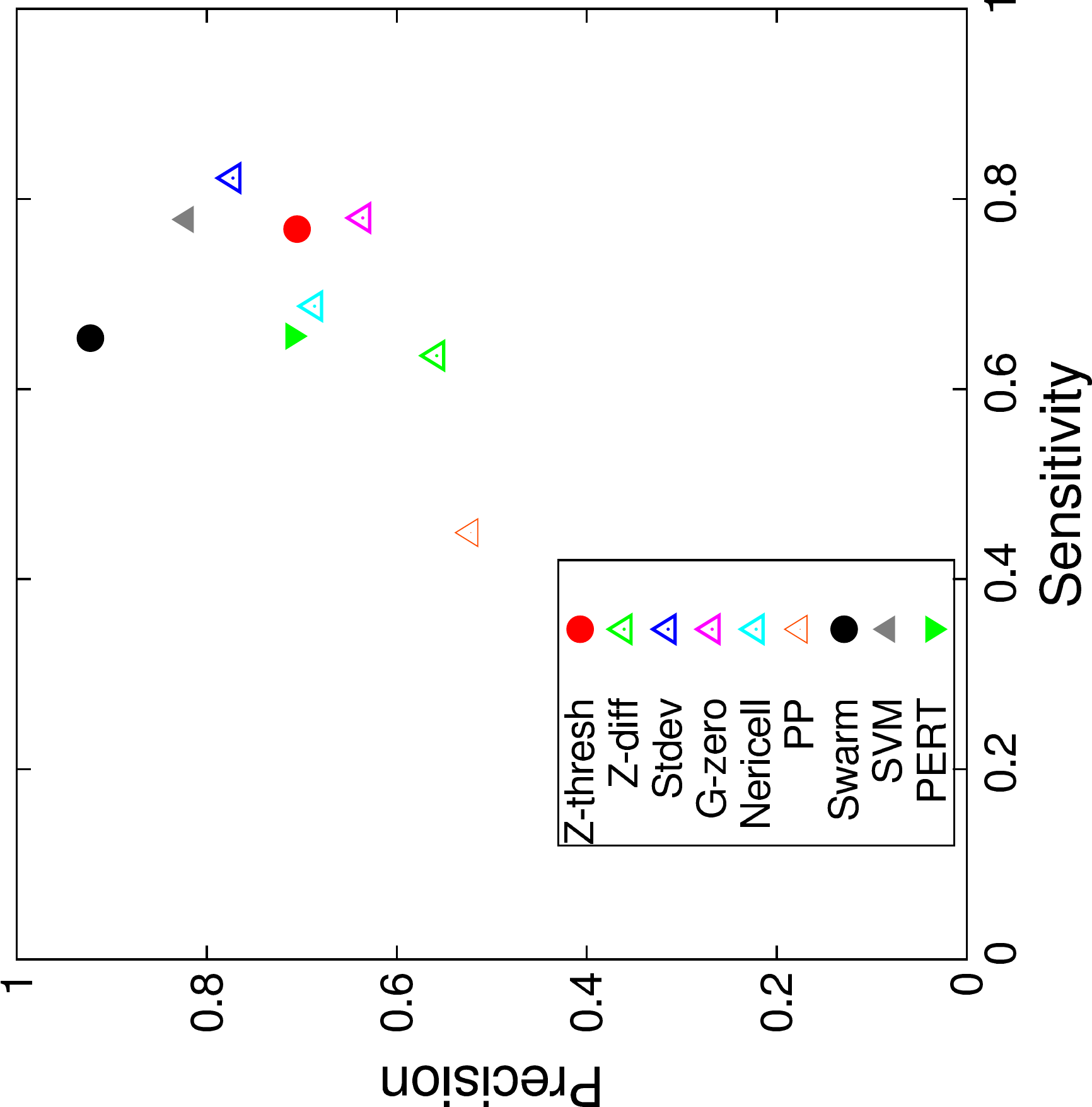}}
    \hfill
    \caption[Sensitivity and precision for all detectors]{A) Sensitivity and precision of every approach for the 30 roads. B) Average performance for all the approaches.
    }\label{fig:whoswho:scatter}
\end{figure}

The different threshold heuristics, binary classifier, and ensemble of threshold heuristics defined above were trained and evaluated on virtual roads that mix sections of road in good condition with samples of anomalies of different types and severity, spanning different temporal lengths, different speeds, and vehicles. Their performance was measured in terms of sensitivity and precision, attempting to present the ability to detect all the anomalies and being correct in detecting the real ones, without triggering false alarms. Figure \ref{fig:whoswho:scatter} shows the results of the evaluated detectors in terms of precision and sensitivity, both individually for each road and their average. From the average values it can be seen that the strongest detectors in terms of both metrics are the SVM detector (SVM), the ensemble (SWARM), and the standard deviation threshold (STDEV).

The results obtained by the best performer in each of the virtual roads are presented in Table \ref{tbl:whoswho:results}. In 14 cases STDEV managed to get the best results, but in terms of average F1 performance (shown in Table \ref{tbl:whoswho:avgs}) SVM produces better results. 

\begin{table}
  \caption{Best results for every road in the data set.}
  \label{tbl:whoswho:results}
  \begin{center}
  \begin{tabular}{c c c c c}
  \toprule
  Road & Best & Sensitivity & Precision & F1 Score\\
  \midrule
  1 & SVM & 0.775	&  0.939  & 0.8493  \\
  2 & SVM &  0.82   &  0.891  & 0.8541  \\
  3 & PER & 0.9		& 0.95  & 0.9278  \\
  4 & STDEV & 0.9	& 0.9 & 0.9743  \\
  5 & Z-DIFF & 0.7 & 0.7777 & 0.7368  \\
  6 & Z-DIFF & 0.8	& 0.7619 & 0.7804  \\
  7 & STDEV & 1 & 0.8947 & 0.9444  \\
  8 & STDEV & 0.975 & 0.8478 & 0.9069  \\
  9 & PER & 0.86	& 0.716  & 0.7818  \\  
  10 & STDEV & 0.86 & 0.8113 & 0.8349  \\
  11 & STDEV & 0.9 & 1 & 0.9473  \\
  12 & SVM &1  & 1 & 1  \\
  13 & SVM & 0.5 & 1 & 0.6666  \\
  14 & STDEV & 1 & 0.7142 & 0.8333  \\
  15 & SVM & 0.6 & 1 & 0.75  \\
  16 & STDEV & 1 & 0.9090 & 0.9523  \\
  17 & SVM & 0.8 & 0.923 & 0.85  \\
  18 & STDEV & 1 & 0.8666 & 0.9285  \\
  19 & SVM & 0.8 & 0.952 & 0.8695  \\
  20 & STDEV & 0.76 & 0.76 & 0.76  \\
  21 & Swarm & 0.6666 & 1 & 0.8  \\
  22 & STDEV & 0.8666 & 0.9285 & 0.8965  \\
  23 & Swarm & 0.7666 & 0.9583 & 0.8518  \\
  24 & STDEV & 0.9111 & 0.8913 & 0.9010  \\
  25 & STDEV & 0.975 & 0.8666 & 0.9176  \\
  26 & STDEV & 0.84 & 0.75 & 0.7924  \\
  27 & SVM & 0.84 & 0.807 & 0.8235  \\
  28 & STDEV & 0.975 & 0.7647 & 0.8571  \\
  29 & PER & 0.75	& 0.78  & 0.7692  \\  
  30 & SVM & 0.575 & 0.884 & 0.6969  \\
  \bottomrule
  \end{tabular}
  \end{center}
\end{table}

 \begin{table}
	\caption{Average F1 scores for all detectors.}
	\label{tbl:whoswho:avgs}
	\centering
	\begin{tabular}{c c c c c c c c c}
		\toprule
		ZTH & ZDIF & STV & GZE & NER & PP & SWA & SVM & PERT\\
		\midrule
         .706 & .528 & .769 & .676 & .675 & .45 & .753 & {\bf .785} & .668\\
        \bottomrule
	\end{tabular}
\end{table}

\begin{figure*}
    \subfigure[]{\includegraphics[width=4.5cm]{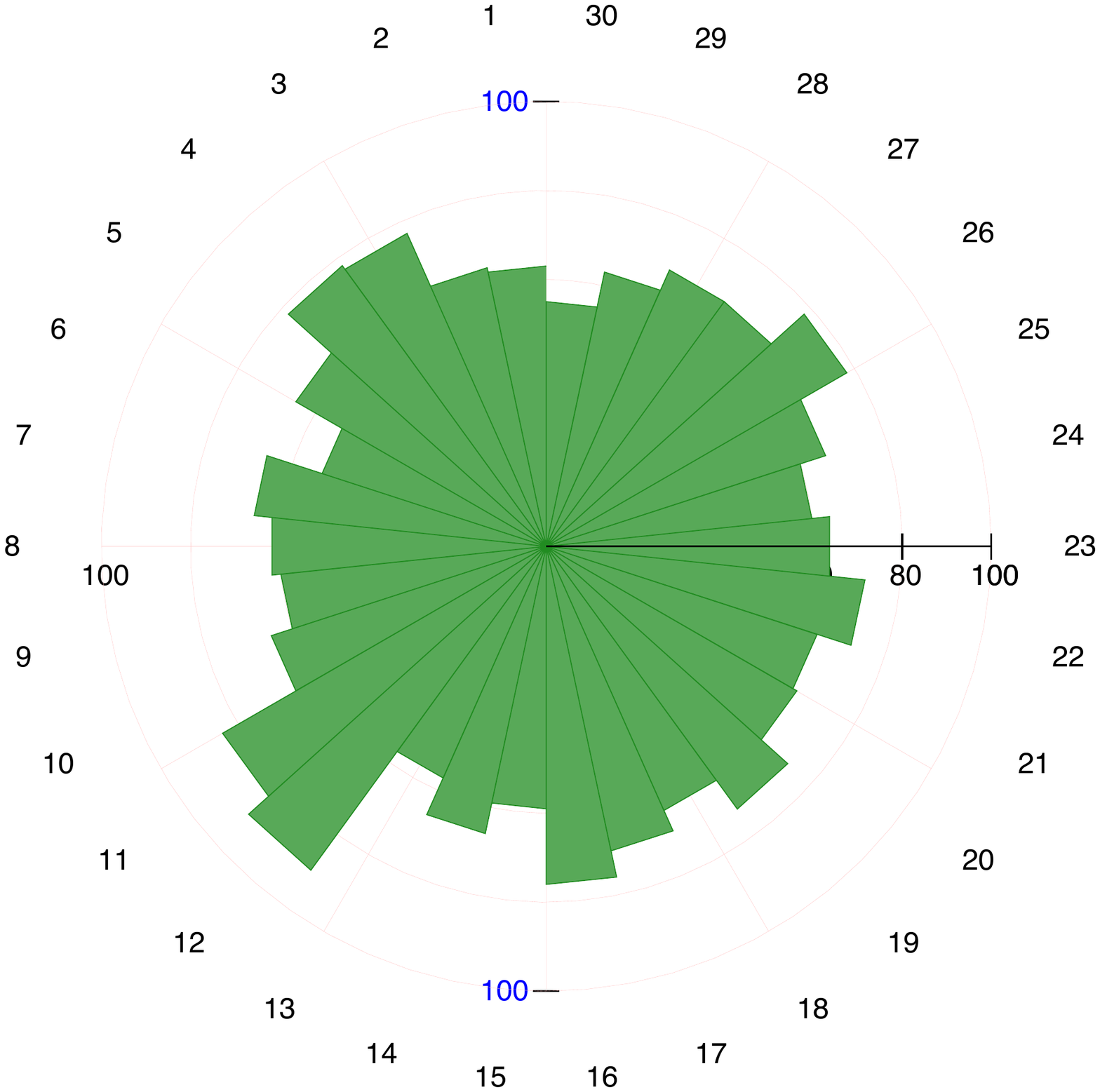}}
    \hfill
    \subfigure[]{\includegraphics[width=4.5cm]{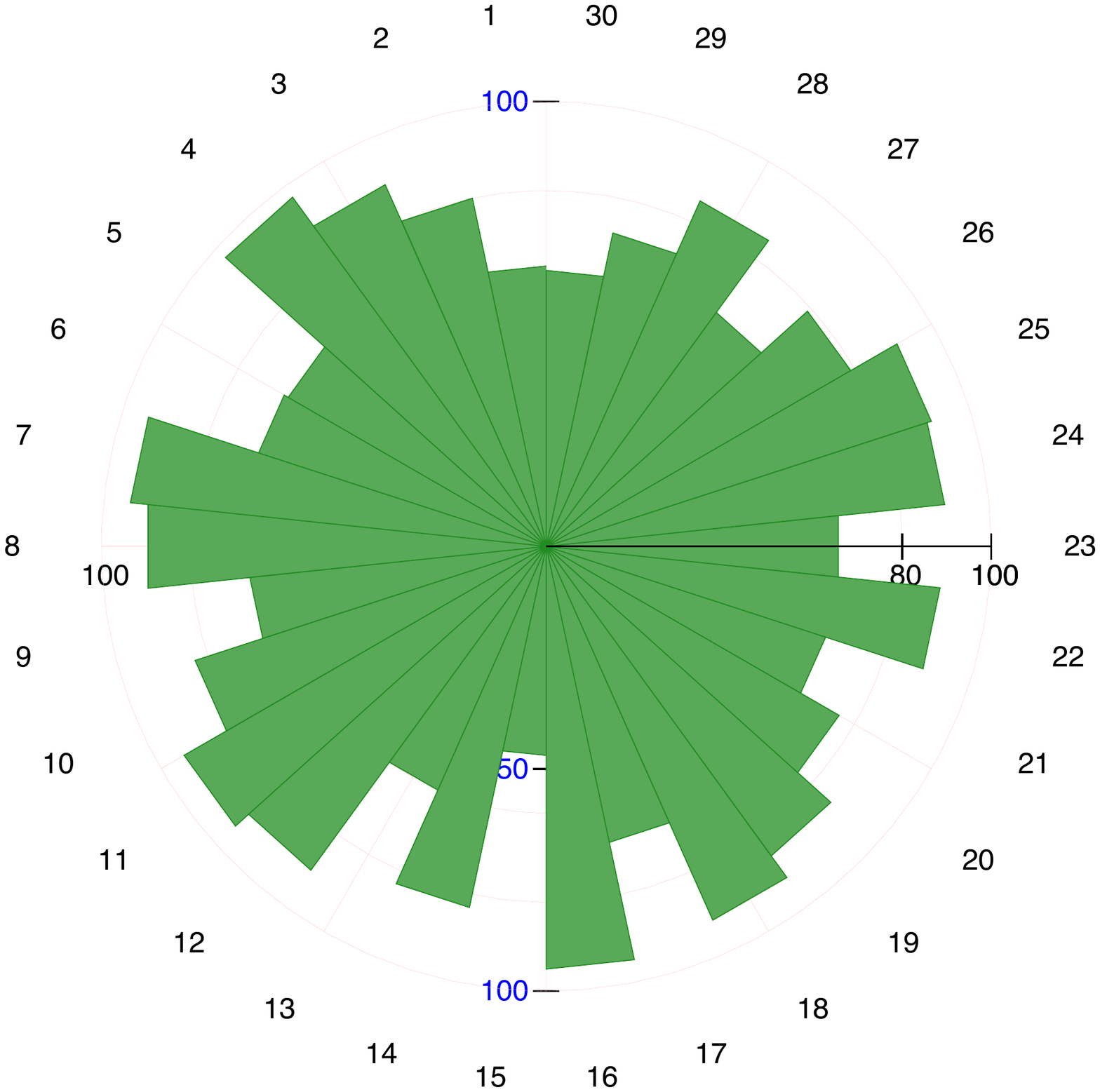}}
    \hfill
    \subfigure[]{\includegraphics[width=4.5cm]{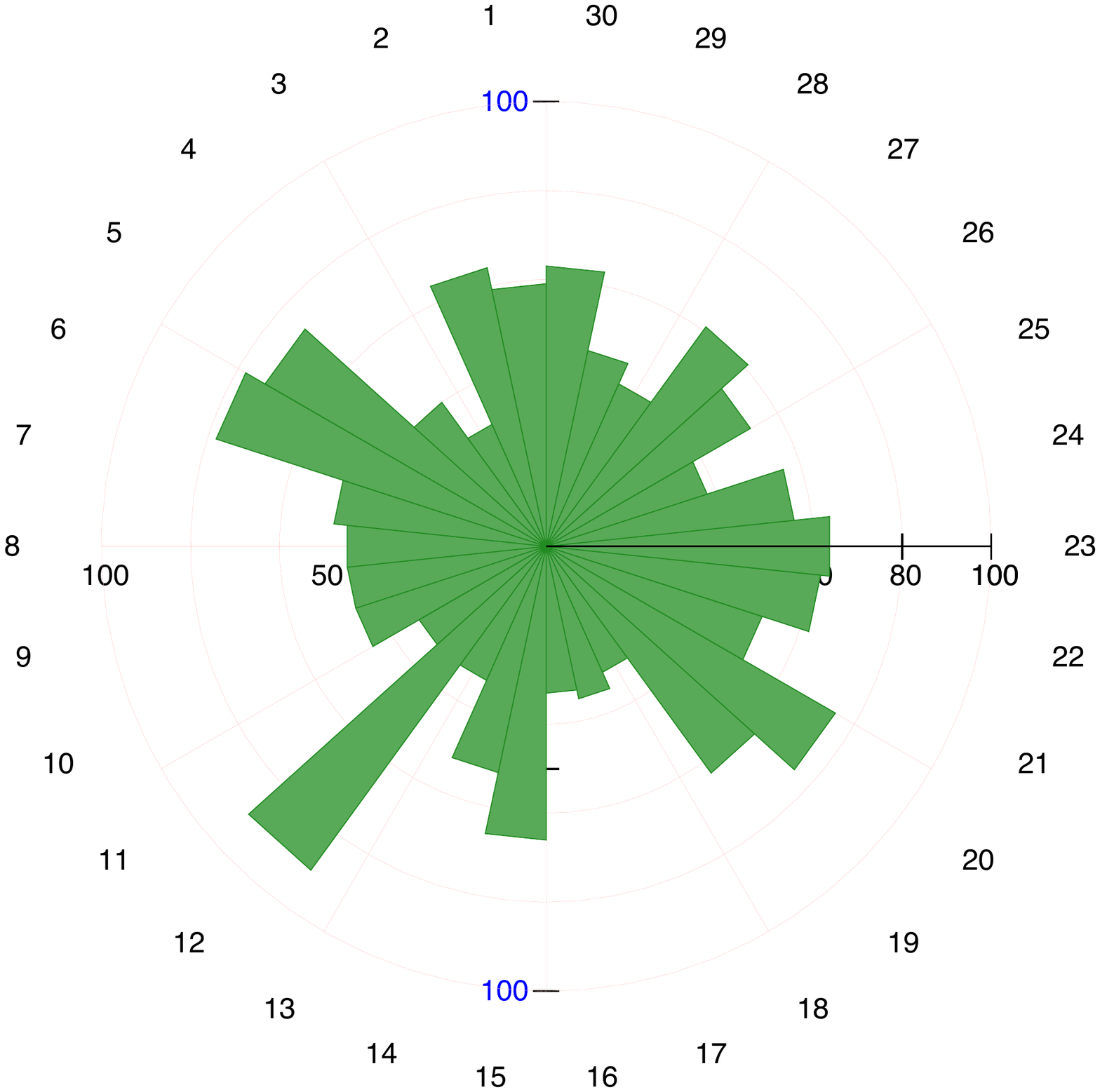}}
    \hfill
    \subfigure[]{\includegraphics[width=4.5cm]{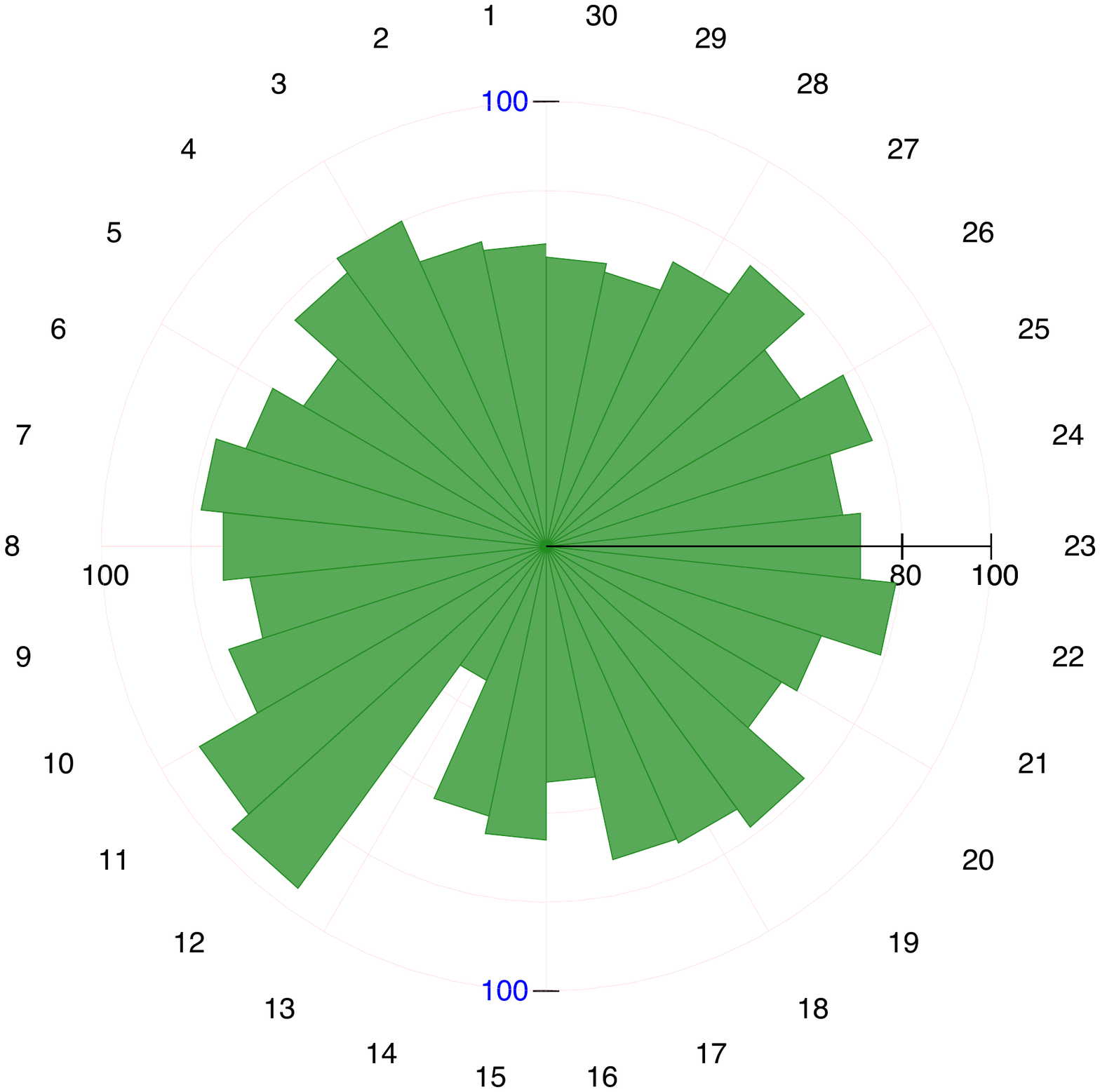}}
    \hfill
    \subfigure[]{\includegraphics[width=4.5cm]{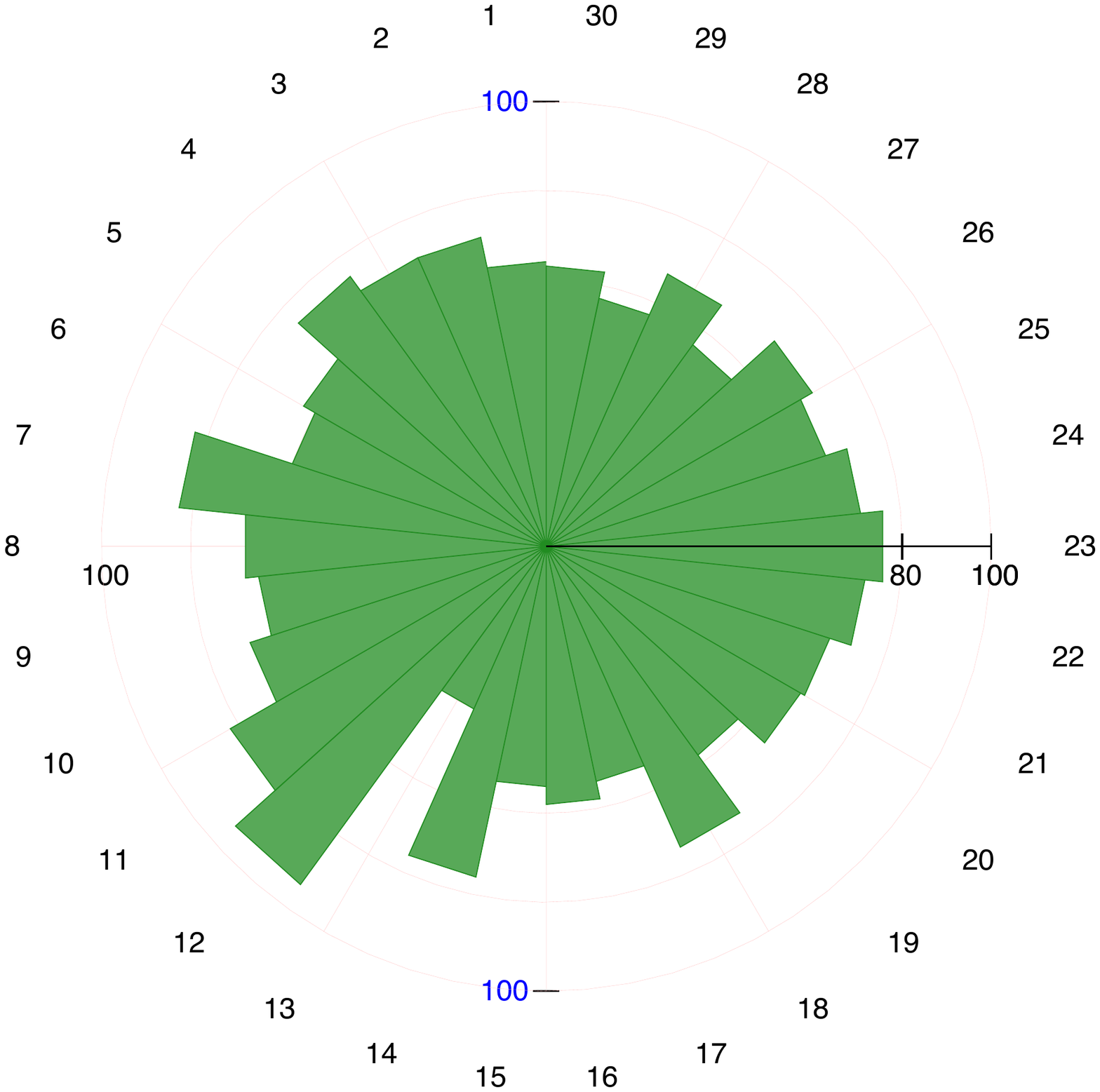}}
    \hfill
    \subfigure[]{\includegraphics[width=4.5cm]{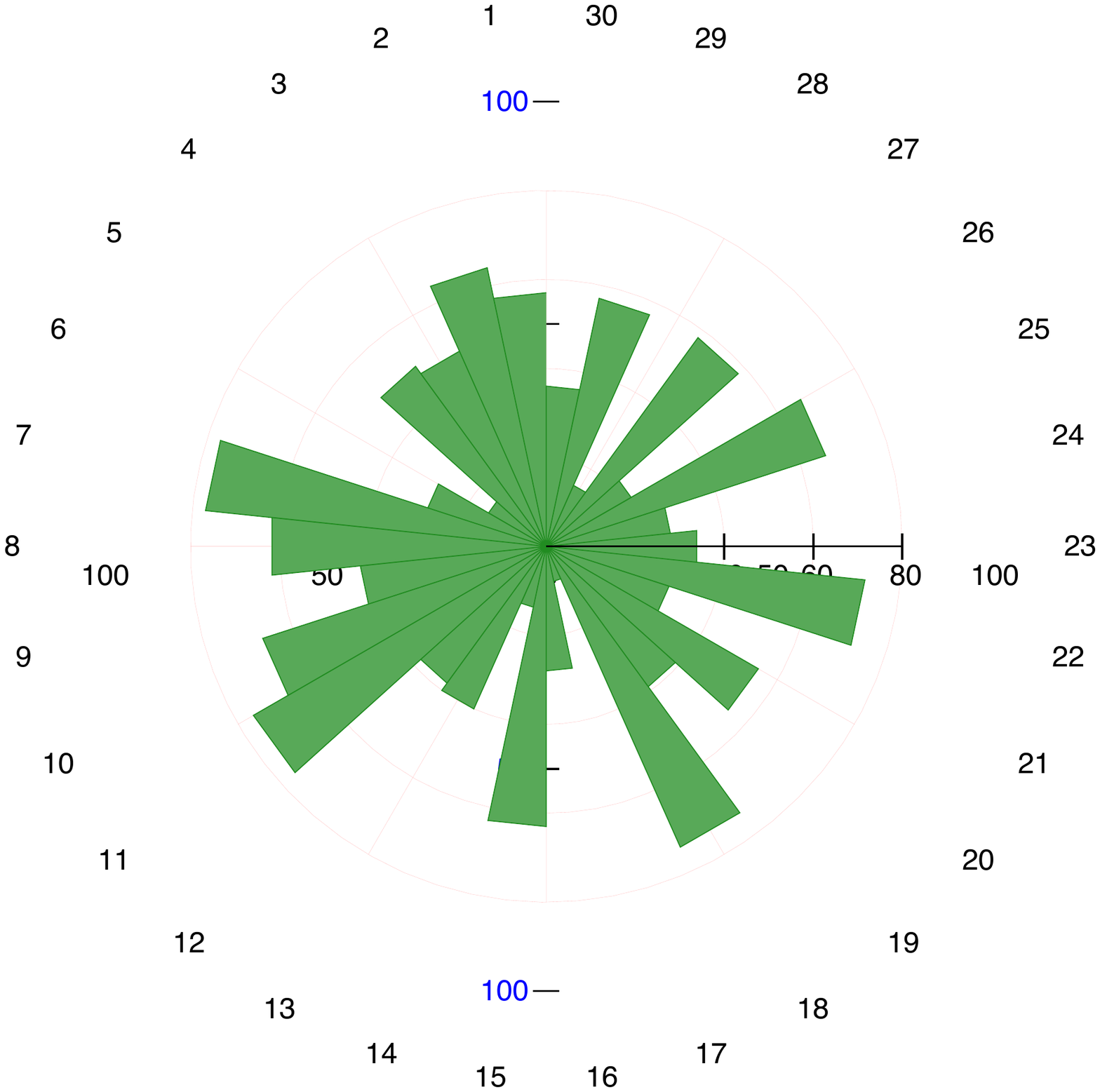}}
    \hfill
    \subfigure[]{\includegraphics[width=4.5cm]{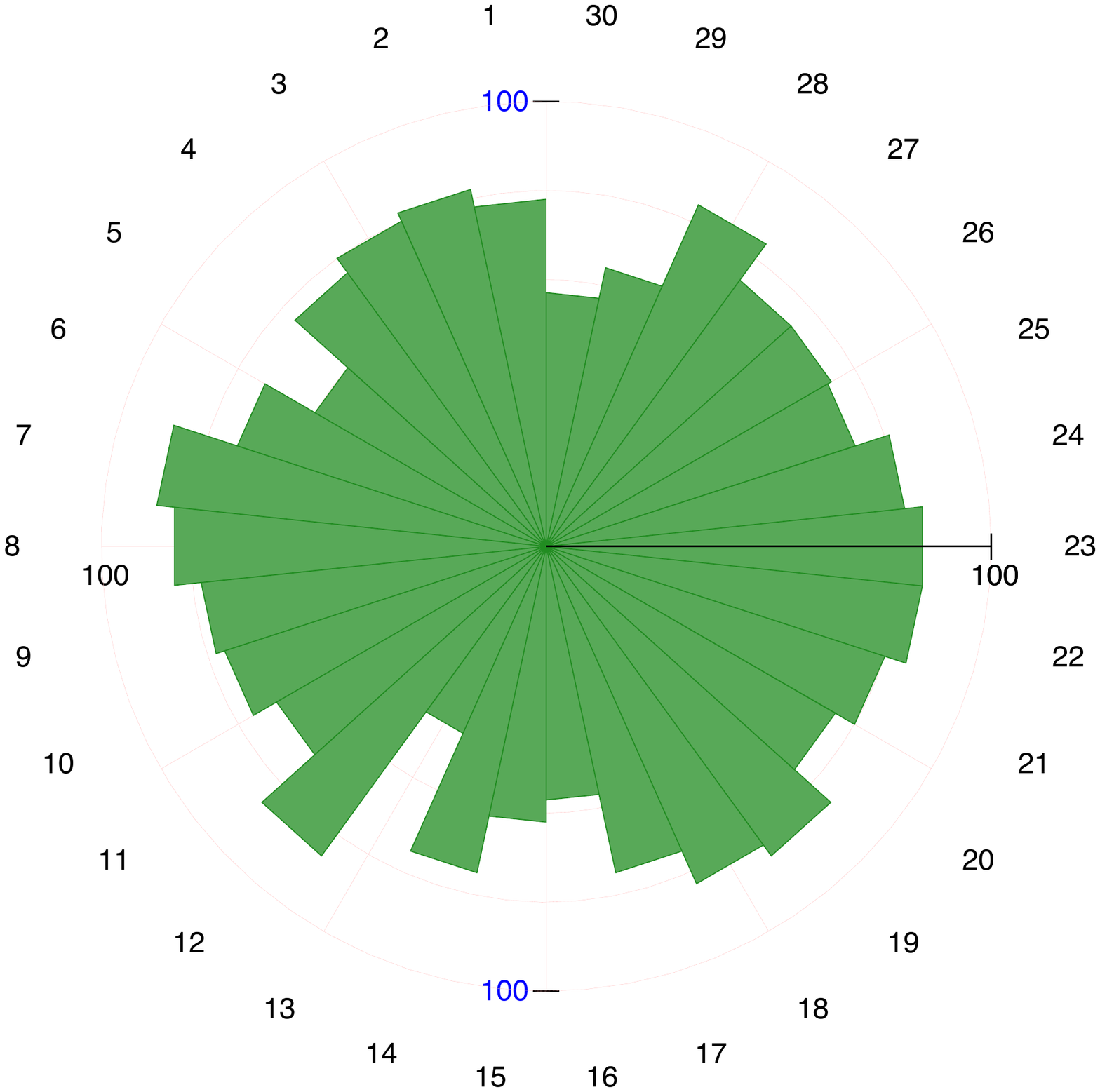}}
    \hfill
    \subfigure[]{\includegraphics[width=4.5cm]{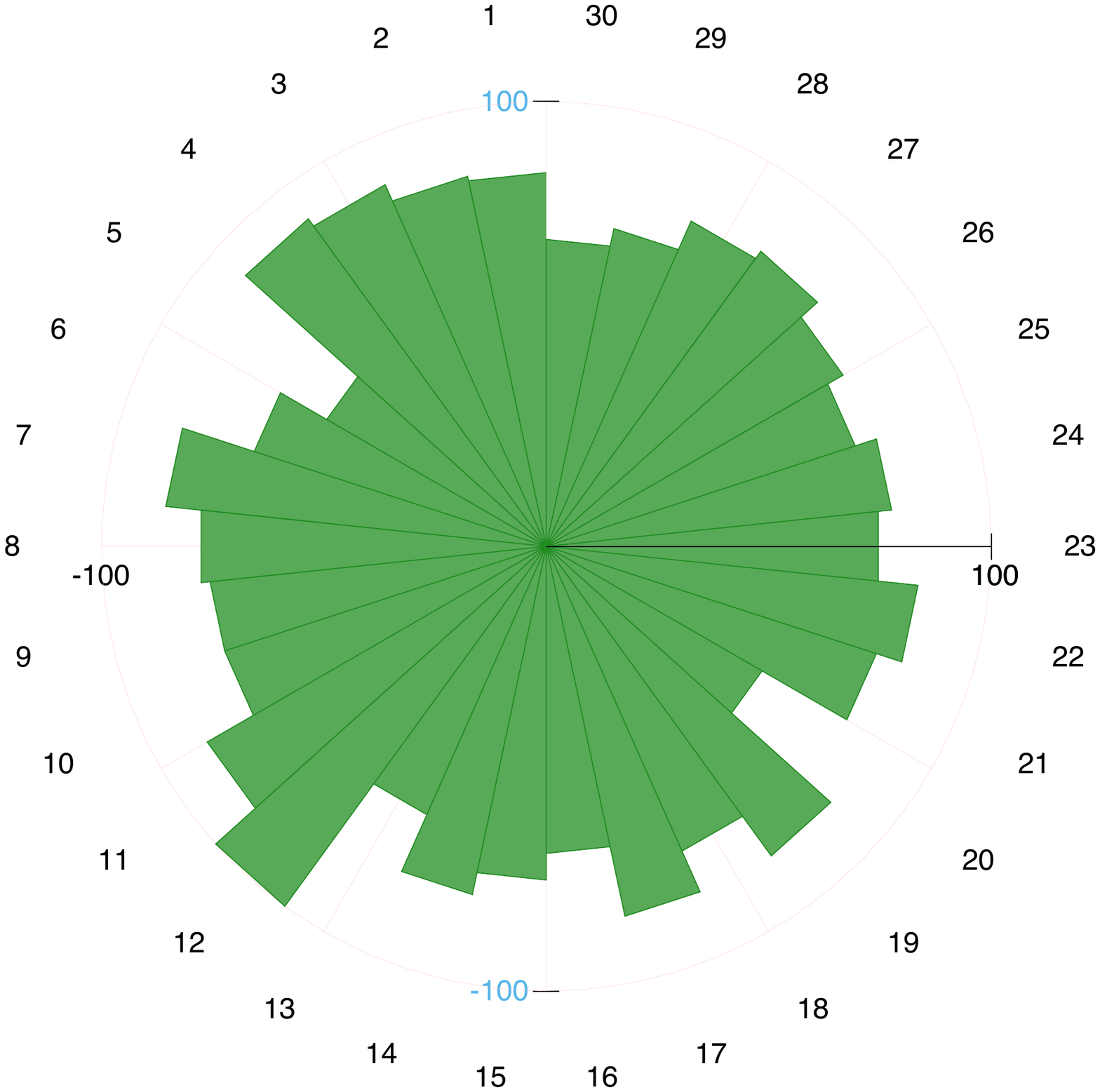}}
    \hfill
    \subfigure[]{\includegraphics[width=4.5cm]{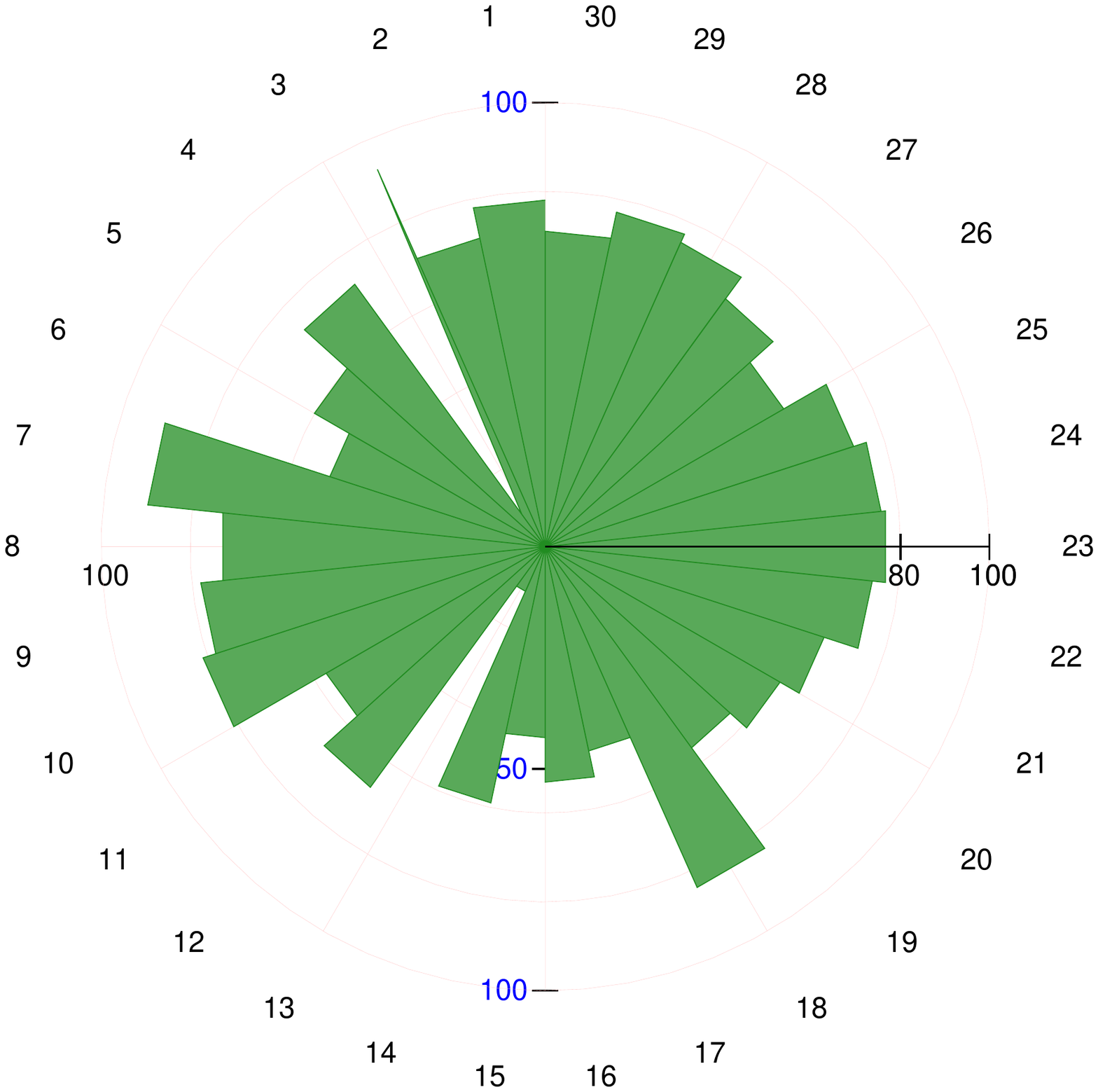}}
    \caption[Rose plots of F1 for all detectors]{Rose plots of F1 for the 30 roads for a) G-ZERO, b) STDEV(Z), c) Z-DIFF, d) Z-THRESH, e) Nericell, f)Pothole Patrol, g) Swarm, h) SVM(Z) and i) PERT.
    }\label{fig:whoswho:roseplot}
\end{figure*}

Rose plots are a variation of the common pie chart, that divide a circumference in n equal sectors and present a magnitude as a percentage of the radius of the circumference. These plots are useful when comparing an attribute in many cases at the same time. In Figure \ref{fig:whoswho:roseplot}, rose plots show the performance of all evaluated detectors for each virtual road in terms of F1. From these plots it can be seen that the detectors behave very differently, with the best performers being consistently good and the others showing greater variability. Larger colored areas indicate better overall performance.
 
\begin{figure}
  \centering
  \includegraphics[width=8.5cm]{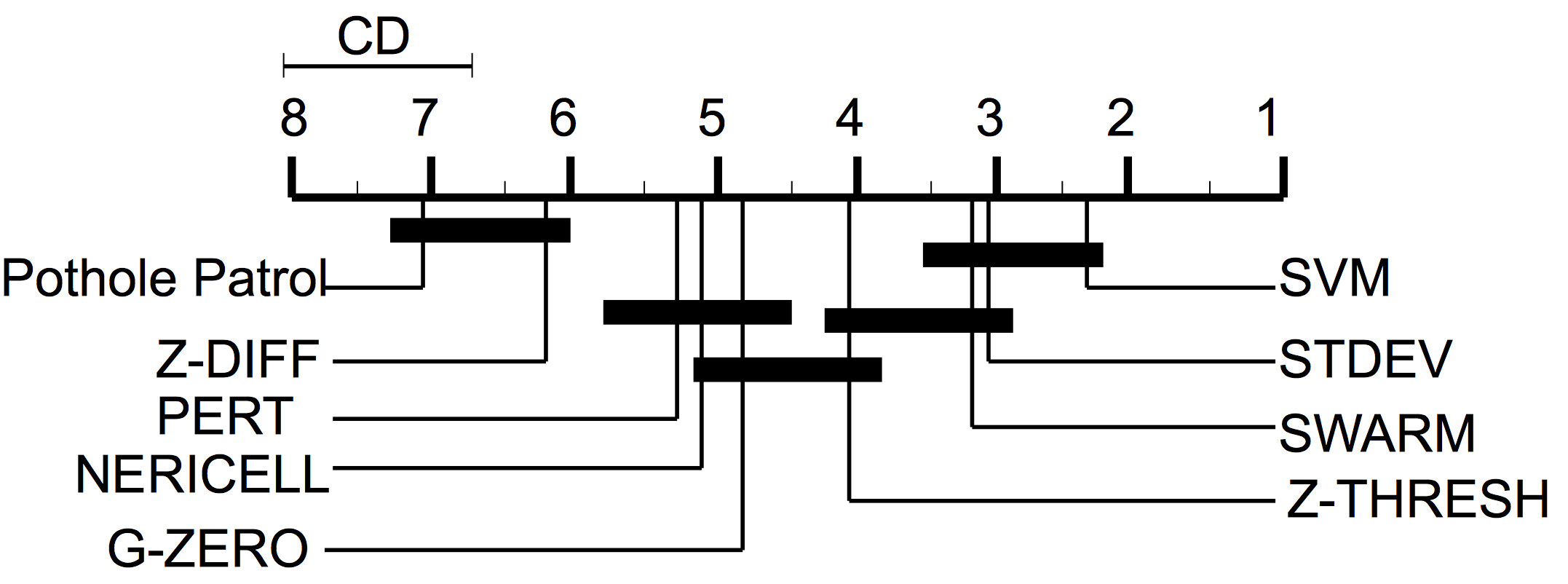}
  \caption{CD diagram with statistical comparison of F1 score for all the detectors.}
  \label{fig:whoswho:cdplot}
\end{figure}

Statistical analysis was performed on these results with the methodology proposed by \citet{demsar2006}, by means of Friedman and  Nemenyi non-parametric tests, with a level of significance $\alpha$ = 0.01. The results are presented by using Critical
Difference diagrams (CD), where the best algorithm appears rightmost. If the
difference between the rank average of the compared detectors is equal or less than the critical difference found with Demsar's methodology, then there is not enough evidence that the methods are statistically different and
the diagram joins the methods through a thick line. Figure \ref{fig:whoswho:cdplot} presents this statistical analysis.

\subsection{Threshold-based Algorithms}

In terms of sensitivity and precision Pothole Patrol (PP), being the only project applied in real-life, surprisingly obtained very poor results. Nericell, which extends the proposal found in Pothole Patrol, had slightly better results. Z-THRESH and G-ZERO come in the middle of the ranking, and STDEV showed the best performance. 

STDEV was the winner in fourteen out of thirty roads, emerging as the clear option among the threshold methods. It offers balanced precision and sensitivity, at a lower computational cost than the proposed ensemble. 

\subsection{A Binary Classification Approach}

SVM is close to STDEV in terms of sensitivity and precision. It clearly outperforms PERT, the machine-learning-based detector based on \citet{perttunen2011}.

This binary classifier was the winner in nine out of the thirty virtual roads, while PERT only wan in three. In terms of F1, this detector obtained the highest score of all the evaluated proposals.

\subsection{An Ensemble Approach}

The ensemble proposal outperforms all individual threshold-based detectors in terms of precision, with above-average sensitivity. In terms of F1 it was only below the performance of STDEV and SVM.

This detector managed to win in only two virtual roads, being misled by the frequent incorrect results of some of its members.

\section{Contributions}

A comparison of the road anomaly detection strategies proposed in the most influential works on the subject was performed, using  the same data and evaluation criteria. This is the first direct evaluation in the literature.

It was found that the most effective individual detector is the proposal by \citet{mednis2011} named STDEV, which consists on applying sliding windows to vertical acceleration and calculating their standard deviation. This dispersion metric is then compared with a threshold value, and a value higher than the threshold indicates a road anomaly is present in that segment of the acceleration time series. 

Two new proposals were made and evaluated: a majority vote ensemble of the six detectors found in the most influential works, and a support vector machine for binary classification of a feature vector obtained by applying some of the underlying ideas of the threshold-based referred detectors. These two new proposals were found to be very effective, with the SVM-based achieving the highest F1 scores in several scenarios. It was observed that the ensemble didn't manage to outperform the standard deviation detector alone, suggesting the frequent incorrect results of the other detection heuristics affected its performance. 

The data employed for this comparison, made public for the research community, is another contribution resulting from this study. Acceleration samples for individual events were stitched together to create virtual roads with a specific number and type of anomalies. Pothole Lab, a web app created for this purpose, was also made public.

The results presented in this chapter were published in: 

\item \textbf{Carlos, M. R.}, Aragón, M. E., González, L. C. Escalante, H. J., \& Martínez, F. (2018, oct). Evaluation of detection approaches for road anomalies based on accelerometer readings -- addressing who's who. \textit{IEEE Transactions on Intelligent Transportation Systems,19}(10), 3334-3343. doi: 10.1109/tits.2017.2773084

%% file: chapters/profiling.tex
\chapter{Road Anomaly Characterization -- A Fine-grained Approach}
\chaptermark{Road Anomaly Characterization -- \\ A Fine-grained Approach}

\label{characterization}

\section{Motivation}

Very few works have addressed the estimation of the dimensions of road anomalies by processing acceleration readings. In the studies where this task is attempted, models based on physics have been proposed to estimate the depth of potholes or the height of speed bumps. The simplest models are based on the relationships between time, vertical displacement, vertical acceleration, and longitudinal speed, while the more complex try to model the response of the suspension systems of the vehicle when an anomaly is encountered. 

Different factors have been considered as relevant. In some cases, knowledge of characteristics of the vehicle are required, in others careful identification of specific moments in the signal have been necessary to perform the estimation of the variable of interest. The authors of these proposals agree that the speed of the vehicle has a significant impact in the results of their estimation, but disagree on how it influences their results. 

Given the complications of the environment in which this kind of estimation is done, it is surprising that no attempt had been reported to treat the road anomaly characteristics of interest as hidden variables, approximated with statistical or machine learning techniques, instead of with direct physical models. 

In this chapter such an approach is adopted, using Machine Learning techniques to estimate physical characteristics of road anomalies by means of regression and classification, taking as input a vector of features obtained from vertical acceleration samples. The learning-to-rank paradigm of learning is also used, to produce a model capable of taking vertical acceleration time series that reflect a vehicle passing a pothole, and producing a list in which the events appear in order, according to the depth or the pothole.

\section{Aims and Objectives}

The objective of this study is the estimation of characteristics for road anomalies by means of machine learning techniques. Specifically, the depth of potholes and the condition of speed bumps (that is, if it is deemed as capable to force drivers to reduce their speed) are addressed.

Secondary objectives are the comparison of results from classification and regression methods for depth estimation, to determine if either or both approaches are equally suitable, and the  feasibility of a ranking approach to order a set of potholes by depth.

In total, proposals for four tasks are described and evaluated: 

\begin{enumerate}

    \item Pothole depth regression, to estimate the distance between road surface and the bottom of the pothole.
    
    \item Pothole depth classification, to determine the depth of a pothole in terms of a discrete scale such as \textit{shallow}, \textit{medium}, or \textit{deep}. 
    
    \item Speed bump condition classification, to determine if the condition of a speed bump still allows it to work as a speed reducing obstacle.
    
    \item Pothole depth ranking, to take a set of signals reflecting vehicles passing over a pothole and establish an order over them, such that the more serious anomalies appear first and the least detectable appear last.

\end{enumerate}

\section{Methodology}

\subsection{Data Set}

A data set was constructed by driving on eight different cars and passing over potholes and speed bumps at speeds between 20 km/h and 40 km/h. In some cases the driver passed over the same anomaly several times, changing the speed and angle the vehicle. Up to five smartphones were placed inside the vehicle in different locations, sampling acceleration at 50 Hz. After reorientation, vertical acceleration was extracted. Given that the duration of a typical event is between one and five seconds, time series were of different length; no padding or cropping was applied.

\begin{figure}
    \centering
    \includegraphics[width=8cm]{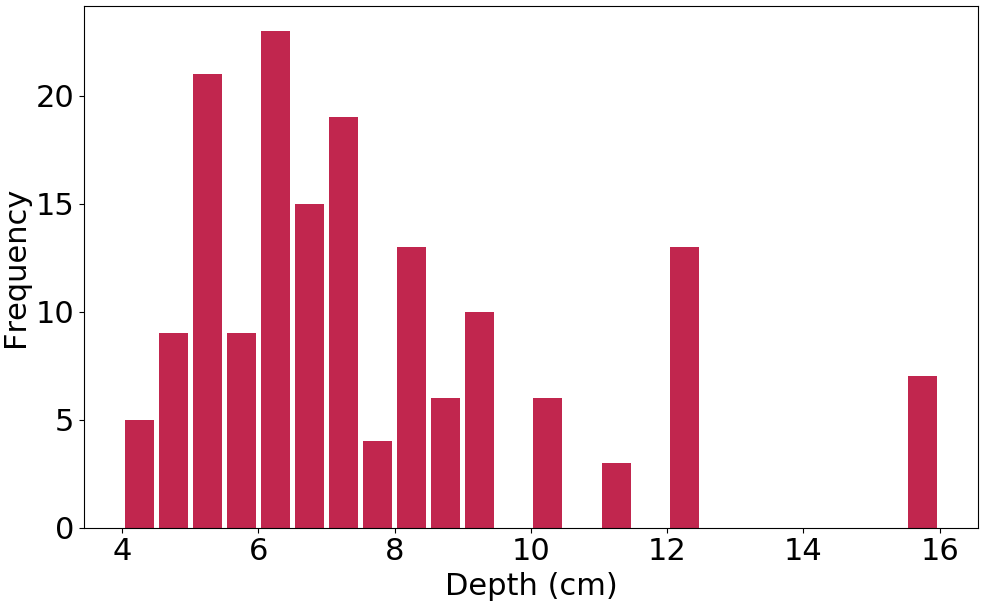}
    \caption[Dimensions for all the potholes in the data set]{Dimensions for all the potholes in the data set. The small insert shows the distribution of depths for all the individual events.}  
 \label{fig:prof:hist_depth}   
\end{figure}

The depth of potholes was considered as the distance between the road surface and its bottom, and was measured in each case (in cm). The distribution of depths for the potholes considered in this study is shown in Figure \ref{fig:prof:hist_depth}. 

For speed bumps (both asphalt and metal), a qualitative label was registered for each anomaly, describing if it actually acts as a speed reducing measure. This label was determined by consensus for each example between the driver and the copilot about their judgment of its condition and effectiveness, based on visual inspection of the width, height, and level of deterioration, and on their perception of the accelerations experienced when the vehicle passed over the anomaly (bumps in bad condition have very little effect on the vertical acceleration of the vehicle, while effective speed reducers greatly affect motion in this axis). Table \ref{tbl:profiling:dataset} shows the number of samples for each type of anomaly in the data set. 

\begin{table}[!ht]
	\caption{Overview of the data set for pothole and speed bump characterization.}
	\label{tbl:profiling:dataset}
	\centering
	\begin{tabular}{ccc}
    	\textbf{Anomaly} & \textbf{Events} & \textbf{Total samples}\\
		\toprule
        Potholes   & 163 & 857  \\        
        Asphalt Speed Bumps   & 59 (Ok: 43, Not Ok: 16) & 308  \\
		Metal Speed Bumps   & 42 (Ok: 26, Not Ok: 16)  & 211  \\     
        \bottomrule
        Total  & 264  & 1,376  \\
        \bottomrule 
	\end{tabular}
\end{table}

\subsection{Feature Extraction}

Fifty-two features in total are used for the models employed in this study in two stages. First, three additional time series were calculated, based on the original vertical acceleration signal: its first derivative (vertical jerk), its first integral (vertical speed) and its second  integral (vertical displacement). The integrals were calculated with omega arithmetic \citep{kupfer2018}. After the four time signals are available, thirteen transformations are applied over each of these time series to obtained a total of 52 features, described in Table \ref{tbl:prof:features}. All of these features are commonly used for signal analysis and summarize different aspects of the signal in one value. These descriptors were chosen under the assumption that they would reflect the patterns both in the time and frequency domains associated with high energy events. 

\begin{table}[!ht]
	\caption[Features considered to feed the algorithmic machinery]{Features considered to feed the algorithmic machinery. The ones marked with an asterisk are the 41 most relevant.}
	\label{tbl:prof:features}
	\centering
	\begin{tabular}{lllll}
    	\textbf{Descriptor} & \textbf{1st Deriv.} & 
		\textbf{Original} &         
        \textbf{1st Int.} &         
        \textbf{2nd Int.}         
        \\
		\toprule
        Mean               & 0 (*)& 13    & 26 (*)& 39(*)\\     
        Std. Dev.          & 1    & 14 (*)& 27 (*)& 40(*)\\
        Max.               & 2 (*)& 15 (*)& 28 (*)& 41(*)\\
        Min.               & 3 (*)& 16 (*)& 29 (*)& 42(*)\\
        Definite integral  & 4    & 17 (*)& 30 (*)& 43(*)\\        
        25th percentile    & 5 (*)& 18    & 31 (*)& 44(*)\\
        50th percentile    & 6    & 19 (*)& 32 (*)& 45(*)\\
        75th percentile    & 7 (*)& 20 (*)& 33 (*)& 46(*)\\
        Zero crossing rate & 8 (*)& 21 (*)& 34 (*)& 47(*)\\
        Spectral centroid  & 9 (*)& 22 (*)& 35 (*)& 48(*)\\
        Spectral flatness  & 10   & 23    & 36 (*)& 49(*)\\
        Spectral roll-off  & 11   & 24    & 37 (*)& 50(*)\\
        Spectral bandwidth  & 12   & 25 (*)& 38    & 51(*)\\
        \bottomrule 
	\end{tabular}
\end{table}

\subsection{Algorithms}

Three different types of machine learning models are used for the tasks stated in the objectives for this study: classification (Class), regression (Regress), and learning-to-rank (L2R). Classification consists of training a model capable of taking as input a vector of features and produce as output a discrete value that identifies the class to which the input belongs (e.g. "big", or "small"). Regression is very similar, except that its output is a real number.

L2R is slightly different and more challenging than the previous two paradigms, since its input is a list of objects and its output an ordered permutation of such objects. A model is expected to either produce an output that can be compared, or be in itself a comparison function, so that previously unseen examples of objects can be ordered (ranked) by some attribute (a hidden variable). 

A ranking problem can be addressed by working with regression models, by sorting a list based on the outputs of the regression, but a pairwise approach can also be used. In that case, the model takes two feature vectors (separate, concatenated, or combined by means of some transformation) and produces an output that indicates which of the two inputs ranks higher in the ordering scheme used. 

In total, eight algorithms were used. Table \ref{tbl:prof:algorithms} presents them, along with their parameters, indicating for which task they were applied. The parameters were selected from the best performing configurations in preliminary experiments.

\begin{table}[!ht]
	\caption{Algorithm-task pairings that were evaluated.}
	\label{tbl:prof:algorithms}
	\centering
	\begin{tabular}{cc}
    	\textbf{Algorithm} & \textbf{Learning task}\\
		\toprule
        Random Forest (RF), 150 trees, Gini crit  & Regress, Class, L2R \\  
        Gradient Boosting (GB), 150 trees, 3 nod/tree & Regress, Class, L2R \\  
        k Nearest Neighbor (kNN), K=5, Mink dist   & Regress, Class, L2R \\  
      	Bayesian Ridge (BR), $\alpha_1,\alpha_2,\lambda_1,\lambda_2= 1e-6$
 & Regress, L2R \\
        Decision Tree (DT), Gini crit & Regression \\   
        Multi-Layer Perceptron (MLP), 26 units, Relu & Class, L2R \\
        Support Vector Machine (SVM), Linear, C=1  & Classification \\
        Naïve Bayes Classifier  (NB) & Classification \\
		\bottomrule 
	\end{tabular}
\end{table}

Three baselines were used for the regression of pothole depth: a Linear Regression model (LR), and two dummy functions MEAN (that always return the mean of the data set) and RAND (which always returns a random number between the minimum and maximum depth in the data set).

Another baseline algorithm is used for the evaluation of pair-wise ranking: FIRST. As it name suggests, it always decides that the first signal in a pair reflects the deepest of the two inputs.

\subsection{Cross-validation}

A 10-fold cross validation scheme was used for result evaluation, with computations performed in thirty runs. For each run, data was split in ten folds and each fold was considered as the testing set once, and nine times as part of the training set. In total, the algorithms were run three-hundred times and each time the performance was evaluated; the results reported are the average of those 300 executions.


\section{Experiments and Results}

\subsection{Pothole Depth Characterization by Regression}

After analyzing the performance of the regression algorithms over the data set it was found that the best result was a relative error of 22\%, and that ensemble methods (RF and GB) manage to produce a better prediction. The Kruskal–Wallis H test (with a significance level of $0.05$) confirms the differences in the performance of the algorithms, and  post-hoc Mann-Whitney U tests validates that RF's error is lower than what the other regressors achieved. Table \ref{tbl:prof:depth_regression} presents the average results obtained with all the algorithms.

\begin{table}[!ht]
	\caption{Average performance metrics for pothole depth regression.}
	\label{tbl:prof:depth_regression}
	\centering
	\begin{tabular}{cccc}
    	\textbf{Algorithm} & \textbf{$\epsilon$} & \textbf{R$^2$} & \textbf{RMSE}\\
		\toprule
        {\bf RF}   & {\bf 0.22} & {\bf 0.45} & {\bf 1.95} \\      
        RF (41 Feat - F test) & 0.22 & 0.45 & 1.95 \\        
        GB   & 0.23 & 0.41 & 2.02 \\
		RF* (5 Feat - Gini)   & 0.23 & 0.40 & 2.03 \\
        BASELINE 1 (LR)   & 0.24 & 0.34 & 2.13 \\        
        KNN  & 0.24 & 0.33 & 2.15 \\
        BR   & 0.24 & 0.33 & 2.15 \\
        DT   & 0.29 & -0.1 & 2.78 \\ 
        BASELINE 2 (MEAN) & 0.29 & -0.02 & 2.8  \\
        BASELINE 3 (RAND) & 0.65 & -2.76 & 5.14 \\
        \bottomrule        
	\end{tabular}
\end{table}

As previously mentioned, \citet{fazeen2012}, \citet{harikrishnan2017}, and \citet{xue2017} found a correlation between vehicle speed and error in their pothole depth estimation. This is not the case for the regression approach presented here: even when the data set over which it was evaluated includes samples collected at up to 45 km/h, with an average speed of 12.5 km/h, the Pearson correlation coefficient between speed and error was found to be -0.0132. Speed is not the most relevant factor in error distribution for this methodology.

\begin{figure}[!ht]
  \centering
  \includegraphics[width=8.5cm]{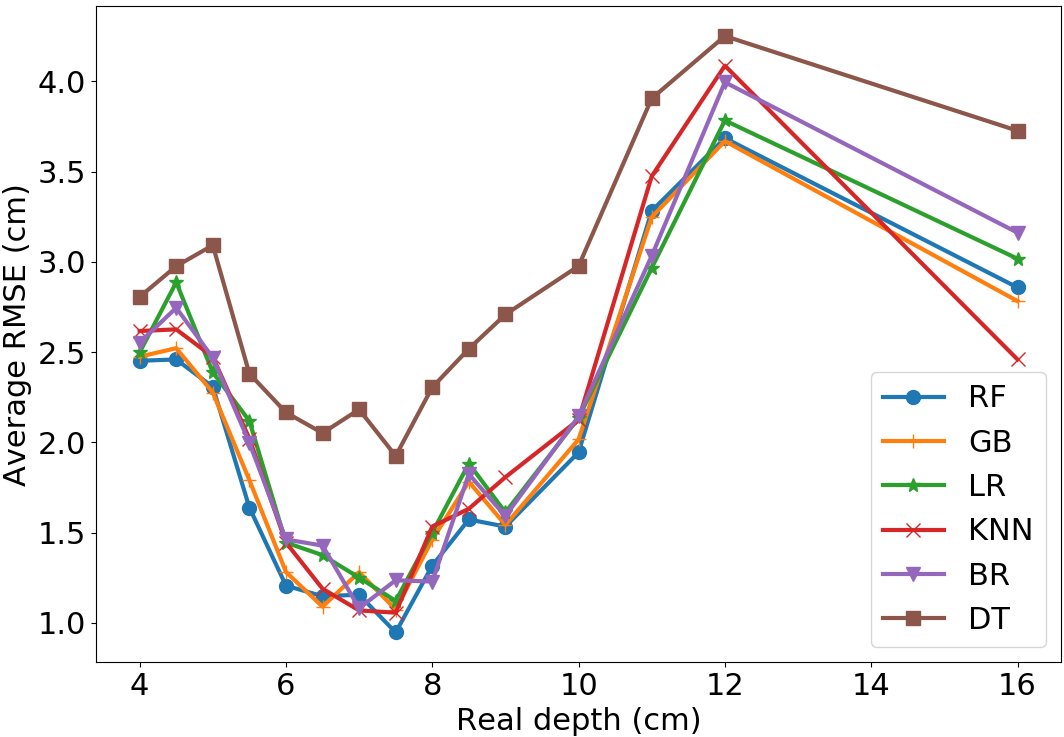}
  \caption{Average RMSE by real depth for the evaluated algorithms.}  
  \label{fig:prof:rmse_depth}   
\end{figure}

The distribution of the regression error with respect to the actual depth of the potholes was also investigated. Figure \ref{fig:prof:rmse_depth} shows how the error varies for different depths. We observe that there exists a depth range for which most of the regressors report their minimum error, these samples correspond to medium-sized potholes (5-8 cm deep) while large potholes present the largest error. A plausible explanation of these differences comes from the distribution of pothole sizes across the data set, since for some depths data includes few or no examples, in particular for very small or very large potholes.

Feature importance analysis was performed to understand the influence of the different features in the results. Two strategies were applied: first, the function for the split criterion (mean squared error) was calculated in the RF model to rank the features from the most to the least informative; second, dimensionality reduction was attempted by using the results of F-tests, analyzing the correlation between the usage of each feature and the expected target.

\begin{figure}[!ht]
 \centering
 \includegraphics[width=8.5cm]{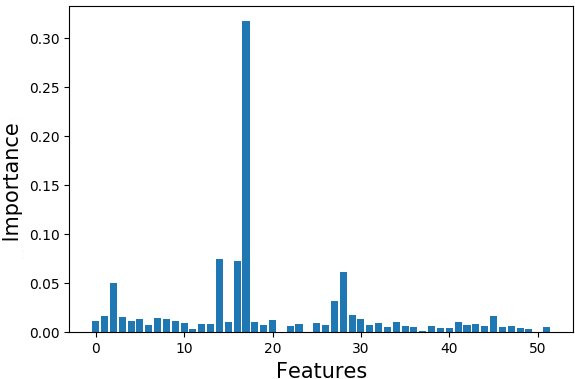}
 \caption[Importance for each feature in the vector.]{Importance for each feature. The features are represented by their ordinal position, shown in Table \ref{tbl:prof:features}. }
 \label{fig:prof:importance}   
\end{figure}

Figure \ref{fig:prof:importance} shows the relative importance for the 52 features, highlighting the five most relevant:  the definite integral of vertical acceleration, the maximum value of vertical speed (the first integral of our original signal), the standard deviation of vertical acceleration, the minimum value of vertical acceleration, and the maximum value of the first derivative of the signal (jerk). If only these five features are used, RF predictions drop by 4.5\% in terms of relative error. After performing feature selection by means of F-tests, it was found that eleven features can be removed without affecting the results for depth regression. The 41 features that can be kept are marked with an asterisk in Table \ref{tbl:prof:features}.

\subsection{Pothole Depth Characterization by Classification}

The characterization of potholes was treated as a classification problem, by defining $n$ classes that equally divide the range of depths in our data set, from 4 to 16 cm. This scheme attempts to separate potholes in $n$ classes such as "small" (4-10 cm) and "big" (10-16 cm) for $n=2$; in "small" (4-8 cm) "medium" (8-12 cm) and "large" (12-16) when $n=3$; and "small" (4-7 cm), "medium-small" (7-10 cm), "medium-large" (10-13 cm), and "large" (13-16 cm) for $n=4$.

The average performance metrics of AUC and G-means for this classification problem are shown in Table \ref{tbl:prof:depth_classification}. As it could be expected, the problem becomes more complicated when the number of labels increases. As with the regression approach, the ensemble methods produce the best results.

\begin{table}[!ht]
	\caption{Average performance metrics for pothole categorical depth characterization.}
	\label{tbl:prof:depth_classification}
	\centering
	\begin{tabular}{ccccccc}
    	& 
        \multicolumn{2}{c}{\textbf{$n=2$}} & 
        \multicolumn{2}{c}{\textbf{$n=3$}} & 
        \multicolumn{2}{c}{\textbf{$n=4$}}\\
        \textbf{Algorithm} & AUC & G & AUC & G & AUC & G \\
		\toprule
        GB    & \textbf{0.89}  &  0.6  & \textbf{0.84} & 0.61 & 0.74 & 0.47 \\ 
        RF    & 0.87 & 0.47 & 0.83  & 0.49 & \textbf{0.75}  & 0.31 \\
        MLP   & 0.84 & {\bf 0.64 } & 0.80 & {\bf 0.62} & 0.71 &  0.48 \\ 
        KNN   & 0.78 & 0.52 & 0.76 & 0.54 & 0.69 & 0.46 \\
		SVM   & 0.83 & 0.37 & 0.82 & 0.36 & 0.74 & 0.28 \\
        NB    & 0.76 & 0.57 & 0.69 & 0.55 & 0.64 & {\bf 0.51} \\        
        \bottomrule        
	\end{tabular}
\end{table}

From this analysis it is concluded that pothole depth characterization by means of classification should be performed with two or three categories at most, and used as a rough guide about the severity of the anomalies. Regression is a better approach if some higher level of granularity is required. 

\subsection{Pothole Depth Ranking}

Two approaches were explored for the problem of ranking vertical acceleration signals by the depth of the potholes they represent: first, a point-wise approach, in which the depth estimated by regression was used as the sorting criteria with the Timsort sorting algorithm, derived from merge sort and insertion sort \citep{auger2015}; and second, a pair-wise approach, based on training a binary classifier with the permutations of pairs of potholes with different depths, attempting to learn a comparison function capable of deciding which of the two inputs reflects the deepest pothole, so that function can be used by a sorting algorithm to produce a ranked list of potholes. 

Tables \ref{tbl:prof:rank_regression} and \ref{tbl:prof:rank_classification} present the average results for point-wise and pair-wise ranking, respectively. The results for the metrics calculated at the training stage are also included. Point-wise ranking managed to outperform pair-wise ranking, but for both strategies the compound error in the predictions results in low correlation between ground truth and the rankings from the proposals. 

\begin{table}[!ht]
	\caption{Average Kendall's $\tau$ over the 10-fold cross-validation for the point-wise approach.}
	\label{tbl:prof:rank_regression}
	\centering
	\begin{tabular}{ccc}
    	\textbf{Algorithm} & \textbf{$\tau$ training} & \textbf{$\tau$ testing}\\
		\toprule
  		RF   &  {\bf 0.9}  & {\bf  0.34 }\\
        GB   &  0.79 &  0.31 \\
        KNN  &  0.47 &  0.24 \\
        LR   &  0.36 &  0.28 \\
        BR   &  0.34 &  0.27 \\
		MEAN & -0.08 & -0.08 \\      
        RAND & 0.0004 & -0.02\\
        \bottomrule        
	\end{tabular}
\end{table}

\begin{table}[!ht]
	\caption{Average Kendall's $\tau$ over the 10-fold cross-validation for the pair-wise approach.}
	\label{tbl:prof:rank_classification}
	\centering
	\begin{tabular}{ccc}
    	\textbf{Algorithm} & \textbf{$\tau$ training} & \textbf{$\tau$ testing}\\
		\toprule
  		RF  & 0.{\bf 95} &{\bf 0.29} \\ 
        GB  & 0.52 & 0.29 \\
        MLP & 0.71 & 0.22 \\ 
        RAND & -0.003 & -0.002 \\
        FIRST & -0.08 & -0.09 \\
        \bottomrule        
	\end{tabular}
\end{table}

For a perfect score with the metric used in this evaluation, a one in the Kendall rank correlation coefficient, the proposed ranking must be perfect, and the position in the order of elements wrongly ranked affects the impact of the errors; this is a hard metric for which multiple errors result in heavy penalization.

\begin{figure}[!ht]
 \centering
 \includegraphics[width=8.5cm]{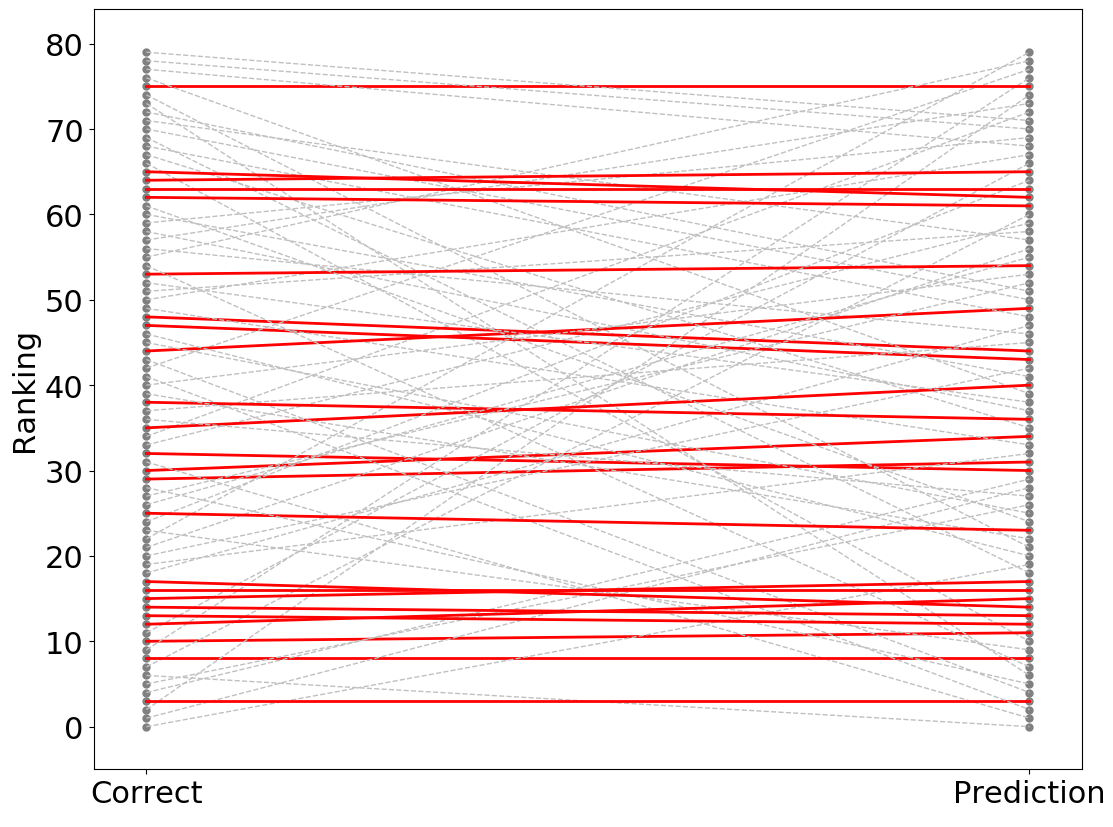}
 \caption{Contrast between the ground truth ordering and the one produced by the point-wise approach.}  
 \label{fig:prof:rankingdepth}   
\end{figure}

It was observed that the best results tended to have the most errors for potholes with depths in the middle of the 4-16 cm range, producing acceptable order for the very shallow and very deep examples. Figure \ref{fig:prof:rankingdepth} shows a comparison of the best order produced with RF in a point-wise approach with a  correct order.

\subsection{Speed Bump Functional Characterization}

The condition of speed bumps, made of asphalt or metal, was also characterized. This problem was addressed with binary classification, attempting to separate those speed reducing measures that were found to work ("OK" condition) and those that did not ("Not OK" condition). 

This problem is different from the detection of a bump because the objective is not to separate signals reflecting normal road from those that have an anomaly, but to make a sub-classification of the events, judging them by the level of the vibration induced on the vehicle. This is a difficult problem because the shapes of the signals for bumps will have some similarities, and the differences that allow for the desired classification will be harder to exploit.

The examples in the data set were grouped in two different ways. First, the conditions of all bumps were classified as OK or Not OK, regardless of the material of which they are made. In the second grouping, asphalt and metal bumps were separated, and the condition was determined taking into account the difference in material. Table \ref{tbl:prof:bump_classification} summarizes the results for both groupings.

\begin{table}[!ht]
	\caption{Average performance metrics for binary speed bump condition classification.}
	\label{tbl:prof:bump_classification}
	\centering
	\begin{tabular}{ccccccc}
    	& \multicolumn{2}{c}{\textbf{Asphalt}} &
        \multicolumn{2}{c}{\textbf{Metal}} &
        \multicolumn{2}{c}{\textbf{Combined}} \\
    	\textbf{Algorithm} & AUC & G & AUC & G & AUC & G\\
		\toprule
		GB    & 0.75 & 0.57 & \textbf{0.72} & {\bf 0.61} & 0.69 & {\bf 0.52 }\\ 
		RF    & 0.\textbf{76} & 0.54 & 0.71 & 0.55 &  \textbf{0.70} & 0.48 \\
        MLP   & 0.75 & 0.57 & 0.66 & 0.53 & 0.62 & 0.49 \\
		SVM   & 0.75 &  0.60 & 0.59 & 0.47 & 0.63 & 0.48 \\
		KNN   & 0.67 & 0.48 & 0.60 & 0.38 & 0.59 & 0.45 \\
        NB    & 0.69 & {\bf 0.61} & 0.62 & 0.48 & 0.59 & 0.41 \\        
        \bottomrule        
	\end{tabular}
\end{table}

As can be seen, it is possible to differentiate between the two classes with the proposed feature vector, in general, although with some uncertainty. It is observed that the two types of speed bumps are better characterized when analyzed separately. 

It can be concluded that this method can be used as a way to get a general idea of the condition of speed reducers, but that it requires further refinement to get more actionable information. 

\section{Contributions}

Machine learning based regression, classification, and ranking were explored for the characterization of road anomalies from acceleration time series, a new approach for a task that has been barely explored in the literature.

Pothole depth estimation was achieved with an average relative error of 22\% by performing regression with a random forest with 150 estimators over a feature vector containing time and frequency domain descriptors. Even if the proposal did not manage to outperform the previously most extensive study in the literature in terms of relative error \citep{xue2017}, it does not seem to be as strongly affected by vehicle speed (with a Pearson correlation coefficient of -0.0132 between error and speed), as all other proposals have reported.

Pothole depth characterization was also performed by classification, attempting to use the same feature vector to determine the depth of road anomalies in discrete levels. The best result was 0.89 AUC for a "shallow" and "deep" classification, and it was observed that increasing the number of categories considered noticeably increases the errors in classification.

A new approach for road anomaly characterization was proposed, based on ranking potholes by their dimensions in order to prioritize action for the most severe. Two approaches were presented: point-wise ranking, by sorting the anomalies with the value predicted by means of regression, and pair-wise ranking, by training a classifier to determine which of two examples is more severe, effectively providing a comparison function for a sorting algorithm. The point-wise approach was found to be more successful when using a random forest, with a Kendall tau coefficient of 0.34. It was observed that shallow and deep potholes were better ranked that anomalies that lie in the middle.   



The results presented in this chapter were submitted to IEEE Transactions on Mobile Computing, and are currently under review: 

\item \textbf{Carlos, M. R.} González, L. C., Wahlström, J. J., Cornejo, R. \& Martínez, F. (2018). \textit{Becoming smarter at characterizing potholes and speed bumps from smartphone data -- introducing a second-generation problem}. (Manuscript under review in \textit{IEEE Transactions on Mobile Computing})

%% file: chapters/risky.tex
\chapter{Aggressive Driving Detection}

\label{risky}

\section{Motivation}

The automatic detection of aggressive driving and the detection of road anomalies can be treated in similar ways, given that both can be observed by means of inertial sensors. The main differences in data processing for these two problems are: 1) the axes that contain most of the relevant information, vertical for road anomalies, and both longitudinal and lateral for aggressive driving; 2) the characteristics of the patterns of interest, with road anomalies presenting many high-frequency features \citep{carlos2015} and aggressive driving events having mostly low-frequency components; and 3) the symmetry present in the direction in which aggressive driving events occur, e.g. left and right sudden lane changes, sudden acceleration and braking.

As mentioned in Chapter \ref{lit_rev}, aggressive driving detection has been approached with threshold heuristics similar to those addressed for the detection of road anomalies in Chapter \ref{potholes}, and pattern matching, by choosing acceleration signals as ideal examples of aggressive maneuvers and comparing new examples against them. Machine learning alternatives have also been used, with hand-crafted feature vectors with which the distinguishing features of the events of interest are assumed to be represented.

Automatic feature learning has been successfully applied in the past for road anomaly detection \citep{gonzalez2017}, and the usage of such techniques results very appealing for aggressive driving detection problem because of the added complications of having to work with two axes and the symmetry of event types.

The focus of this chapter is the study of automatic feature extraction as an alternative for the classification of aggressive maneuvers, removing the need for manual design of feature vectors or selection of base patterns. Symbolic aggregate approximation (SAX), Bag of Features (BoF), and the Bag of Words (BoW) representations are used to determine shapelets, with the intention of producing a feature vector useful in the identification of aggressive driving events.

\section{Aims and Objectives}

The objective of the study presented in this chapter is to find if automatic feature extraction techniques are effective for the tasks of aggressive driving detection and classification, as an alternative to the techniques conventionally used for this problem. 

The secondary objectives are the evaluation of several alternatives, in order to find the most effective, the determination of the range of parameters for these methods that are best suited for this problem, and an analysis of the performance of different machine learning algorithms to use along the automatically learned features.

\section{Methodology}

Two separate, but related, tasks are addressed in this chapter. The first is the detection of aggressive driving from regular driving, without considering what type of maneuver is performed. For the second, it is assumed that aggressive driving events are present in the data, and the problem consists on finding the specific kind of event represented by longitudinal and lateral acceleration time series. 

Three experiments are performed, using three different data sets: 1) the evaluation of a hand-crafted feature vector and the usage of automatic feature extraction with a bag of words methodology, using the data from \citet{zylius2017}; 2) the comparison of bag of words, bag of features, and SAX for automatic feature extraction to classify aggressive maneuvers, using our own  data set; 3) the evaluation of features learned with the bag of words methodology, on data from \citet{junior2017}.

As a reference for the performance of the proposed alternatives the results obtained with a robust hand-crafted feature vector, presented in \citet{zylius2017}, are considered for the task of aggressive driving detection, and a vector of features containing the statistical summarization features of acceleration samples, described in \citet{junior2017}, is used as baseline for the classification of aggressive driving events.

\subsection{Data Sets}

Three data sets were used in these experiments. Two originally consisting of long time series, that for which the method of segmentation is described, and one conformed with individual events captured at different times. 

In order to avoid introducing some bias because of the different lengths of the short time series with the proposed feature extraction methodologies, the signals were right-padded to an equal (200 acceleration samples, about four seconds) with  Gaussian noise, using the mean and standard deviation of the readings for each example.

\subsubsection{Žylius}

This data was presented in \citet{zylius2017} and kindly provided by the author in private communications. It consists of ten long time series (TS) reflecting about 2.6 hours of triaxial accelerometer readings, sampled at 17 Hz, oriented with respect to the vehicle's axes. Six of these TS were captured while driving in an aggressive manner, and the  rest reflect normal (not explicitly aggressive) driving. There is no information about the specific type of maneuvers, aggressive or not, performed during data captured.

The time series in this data set were preprocessed in the same way described in \citet{zylius2017}, removing the redundant signal captured when the vehicle is not in motion. This preprocessing was done as follows: windows of 1,000 acceleration readings were extracted, and then feature vectors were created considering the three axes. The transformations used to create the vector include the augmented Dickey-Fuller statistic \citep{dickey1979}, Akaike information criterion \citep{arnold2010}, and average logarithm of lower bound power spectral density coefficients. Principal Component Analysis was used for dimensional reduction, to produce feature vectors of three dimensions, which were clustered with Gaussian Mixture Models. The cluster of interest was manually selected by visual inspection of the original version of the grouped signals. Finally, a median-mean filter was applied to the selected time series, to eliminate outliers in acceleration. At the end of this process 15 TS containing normal driving were obtained, along with 65 containing aggressive driving events. 

\subsubsection{Ferreira et al.}

\citet{junior2017} collected accelerometer, linear acceleration, magnetometer, and gyroscope readings with a sampling rate of 50 Hz. Two different drivers completed four trips of approximately 13 minutes each, with a fixed smartphone aligned to the vehicle. The segments in which aggressive driving maneuvers were performed are specified along the data, and the individual events were manually extracted. 

In total, the data set formed includes 69 triaxial acceleration short time series, reflecting driving maneuvers in seven different categories, considering symmetric types of events. This data set is described in Table \ref{tbl:risky:data_ferreira}.

\begin{table}[!ht]
	\caption{Driving maneuvers included in the data set by Ferreira et al.}
	\label{tbl:risky:data_ferreira}
	\centering
	\begin{tabular}{cc}
    	\textbf{Category} & \textbf{Examples} \\
		\toprule
		Aggressive braking & 12 \\
		Aggressive acceleration & 12 \\
		Aggressive left turn & 11 \\
		Aggressive right turn & 11 \\
		Aggressive left lane change & 4 \\
		Aggressive right lane change & 5 \\
		Non-aggressive events & 14 \\
        \midrule
        & 69 \\
        \bottomrule        
	\end{tabular}
\end{table}

\subsubsection{Our Data Set}

Acceleration readings sampled at 50 Hz were collected for 450 individual events. Four types of events were considered, and the distribution of the examples is as follows: 75 examples for swerving left, 75 for swerving right, 150 for sudden braking, and 150 for sudden acceleration. Smartphone orientation was not known at the time of data collection, and were semi-automatically reoriented by performing vertical reorientation by using the methods described in \citet{carlos2016}, using principal component analysis to separate acceleration in the X and Y axes, and then manually inspecting and rotating these last two axes to match the expected behavior for the type of event registered at the moment the acceleration was collected; this method is similar to the one outlined in \citet{larsdotter2014}. 

\subsection{Feature Extraction}

The feature extraction methods used are now explained\footnote{It has to be taken into account that different alternatives exist when using these methods to work with data from several simultaneous sensors or axes: a) the time series into one (with the Euclidean norm or some other transformation) before feature extraction; b) treating each individual time series separately, using one model for each and concatenating all outputs; c) fitting one model for all of the signals, combining training data from all of the sources and then encoding each with this model, finally concatenating all the features produced.}, ordered from the most simple to the most complex. Commentary on the intuition behind these methods, their similarities and differences, is also presented.  

\subsubsection{Statistical Summarization}

This conventional feature extraction method was presented in \citet{junior2017}. It consists of concatenating statistical descriptors (mean, median, standard deviation, and trend) calculated over \textit{frames}, nested subsequences of different length. A sliding window is applied over the time series to achieve regularization, dividing the complete time series in $nf$ segments. These segments are grouped so that the statistical descriptors are calculated over shorter and shorter subsequences: first over $nf$ contiguous segments (the whole time series), then over $nf$ - 1 contiguous segments, $nf$ -2 contiguous segments, and so on until they are calculated over just one segment.

The main objectives of this technique are dimensional reduction and length regularization, like in most techniques for time series analysis. However, most techniques are atemporal, presenting information for the whole time series at once, and this scheme of overlapping subsequences (of different lengths) provides some temporal information. This happens because the subsequences containing the anomaly will most likely have different statistical measures when compared with the rest. The use of the slope of a fitted line as an indicator for the trend of a specific subsequence is useful to recognize which of the possible symmetric events are detected by the other statistics (e.g. sudden braking vs. sudden acceleration of the vehicle).

\subsubsection{SAX}

Symbolic aggregate approximation (SAX) \citep{lin2003} provides a way to transform time series into a sequence of standard values obtained by piecewise aggregate approximation. All time series are first standardized to zero mean and unit standard deviation, then reduced by grouping $n$ consecutive readings and taking their mean value to represent the group. The possible average values are regularized by restricting them to $m$ possible discrete values, which are commonly represented by letters (\textit{a} for the lowest, \textit{b} for the second lowest, and so on).

Like the previous method, SAX regularizes by  length (with the grouping of readings), but also regularizes the amplitude of the signal by using only a few possible discrete values. The regularized time series can be modeled as a string of symbols, on which patterns could be easier to find from the frequency of specific \textit{n-grams}. Some degree of temporal information is preserved if the n-gram is of length greater than one.

\subsubsection{Bag of Words}

This model attempts to find $K$ standard subsequences of equal length, that can be considered shapelets. These standard elements are supposed to be a generalization of the subsequences most commonly present in the type of time series that are being analyzed. A reconstruction of the original signal can be produced by concatenating these standard elements, however the goal is not necessarily to make this approximation very close to the original, but to find elements that allow a good enough reconstruction in which the distinctive features of the signals of interest can be identified.

There are two stages for the bag of words model, as presented in \citep{wang2013}. The first consists on the determination of the \textit{words} to be used in the model, while the second consists in the transformation of the original signals into histograms of these words.

In order to find the \textit{words} in the model the first step is the delimitation of subsequences of equal length $L$, with the possibility of having overlap (which help increasing the fidelity of the model by providing examples at different instants). Each time series designated for training is segmented, and each segment is then considered as vectors of $L$ dimensions; $K$ centroids are found by applying a clustering algorithm to the vectors, and these centroids  are now considered as codewords, collectively forming the codebook, i.e. the \textit{vocabulary} with which signals are represented. 

Once a codebook has been determined, the second stage begins by splitting each signal in segments, and each one is then represented by the nearest centroid found with the clustering algorithm. An approximation of the original signal can be produced by concatenating the codewords that correspond to the original segments, and a histogram is created by counting how many times each one is used to recreate the signal. Both stages of the creation of this feature extraction technique are shown in Figure \ref{fig:risky:bow}, and described as pseudocode in Algorithms \ref{alg:risky:fitting} and \ref{alg:risky:encoding}.

\begin{algorithm}[!ht]
	\caption{Fitting stage}
	\SetAlgoLined
	\SetKwInOut{Input}{Input}
	\Input{S: list of training signals of the same classes; K: number of codewords to find; L: length of sliding window; O: number of readings that overlap between consecutive windows.}
	\KwResult{$K$ codewords to represent the class.}

	W $\leftarrow$ \{\} // Segments from the original  signals. \\

	\ForEach{s in S}{
		W $\leftarrow$ W + SplitInWindows(s,L,O) \\
}

	codewords $\leftarrow$ {K-Means(W,K)}

	\Return $K$ codewords
	\label{alg:risky:fitting}	
\end{algorithm}

\begin{algorithm}[!ht]
	\caption{Encoding stage}
	\SetAlgoLined
	\SetKwInOut{Input}{Input}
	\Input{s: a signal accelerometer sample; L: length of sliding window; O: number of readings that overlap between consecutive windows; B: the codebook}
	\KwResult{A feature vector of the signal s.}

	F $\leftarrow$ initialize vector of size $|B|$ // Output feature vector \\
   	W $\leftarrow$ \{\} // Segments extracted from s \\
		W $\leftarrow$ SplitInWindows(s,L,O) \\
       \ForEach{w in W}{
        	i $\leftarrow$ Index in B of codeword closest to w\\ 
           F[i] $\leftarrow$ F[i]++ //Increase count of codeword $i$ 
        }
	\Return F
	\label{alg:risky:encoding}	
\end{algorithm}

\subsubsection{Bag of Features}

The bag of features model \citep{baydogan2013} also transforms time series into histograms of regularized features, but it fuses information obtained at different levels of resolution. 

Feature extraction starts with the selection of a few segments of the time series, which can be of different lengths and can be taken from any part of the complete time series (there is no need to include all of it when fitting the model). Each of these segments is then split in small intervals of a few points, and these are summarized with their mean, variance, and slope. The statistical descriptors, plus the mean and variance of the whole segment, are concatenated to form a feature vector that represents each segment. The newly calculated feature vectors are given the label of their parent time series, and are considered a new data set for which a Random Forest classifier is trained. This process is equivalent to the fitting stage of the above mentioned bag of words model.

The time series used for training and testing are converted into histograms by first being split in segments and obtaining the feature vector for them, and then using the Random Forest classifier to give the class probabilities for each segment. Those probabilities are then counted, and a histogram is formed. 

The main differences between this method and the bag of words are: a) for bag of features, segments of different lengths are considered; b) bag of words considers all segments of the time series while bag of of features can work with only some of them; c) bag of words regularizes samples directly, while bag of features regularizes based on the statistic information calculated from some subsequences of the time series; d) bag of features attempts to mitigate the influence of noise by means of random sampling, while bag of words relies on clustering.

\section{Experiments and Results}

\subsection{Detection of Aggressive Driving}

The bag of words model described above was used to learn a representation from which aggressive driving maneuvers could be separated from normal driving. Žylius' data set and results were considered as a benchmark. 

Given that acceleration in the longitudinal and lateral axes were considered, the bag of words model was applied to three different combinations of them: each axis considered independently, the concatenation of signals in both axes, and the Euclidean-norm of the two time series. Codebooks ranging from 2 to 250 words, considering book lengths from 5 to 300 samples, were evaluated. The application of 50\% overlapping, and the usage of TF-IDF (a weighting factor that highlights the most relevant codewords and penalizes the least relevant, regularizing the feature vectors and simplifying the classification task) \citep{Tfidf} were also considered in the grid search for parameter optimization. Table \ref{tbl:risky:BoWParameters} presents the parameters that were evaluated, creating 5,184 different combinations. 

\begin{table}
	\caption{Parameters considered for the bag of words model.}
	\label{tbl:risky:BoWParameters}
	\centering
	\begin{tabular}{cc}
		\toprule
		Parameter & Evaluated value\\
		\midrule
        Accelerometer axes & X, Y, $\|(X,Y)\|_2$, XY\\        
        Classifiers & MLP, RF, GNB, KNN\\        
        Codebook size & $K = \{ 2, 5, 10, 20, 50, 100, 150, 200, 250 \}$ \\        
        Event's length (time stamps) & $L = \{ 5, 10, 20, 50, 100, 150, 200, 250, 300 \}$ \\
        Overlapping (50\%) & Yes/No \\
        TF-IDF  & Yes/No \\
        \bottomrule        
	\end{tabular}
\end{table}

\begin{figure}
	\centering
	\includegraphics[width=8.5cm]{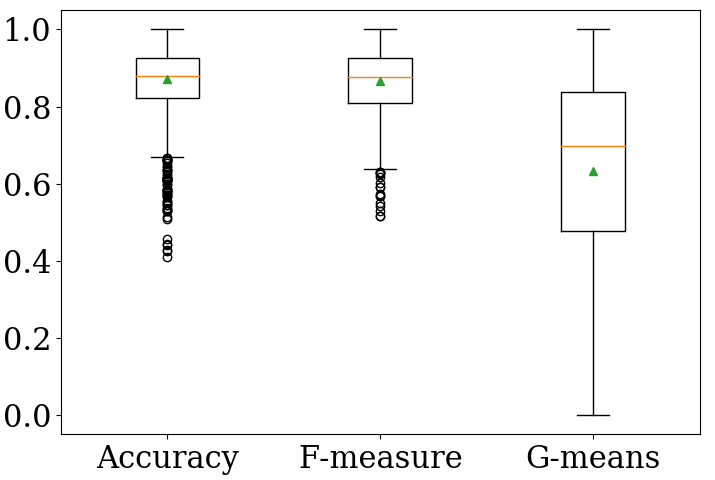}
	\caption{
        Box plots summarizing the distribution of classification results for the bag of words model.
	}
    \label{fig:risky:zylius_all}
\end{figure}

In total, 554 (10.6\%) out of the 5,184 total combinations of the parameters produced an accuracy of 95.5\% or more, that is, as good as or better results than those achieved in \citet{zylius2017}. Median values suggest that, in general, the model can be tuned to produce very good results for aggressive driving detection. The box plot in Figure \ref{fig:risky:zylius_all} presents the distributions of the results for the binary classification of normal and aggressive driving, in terms of accuracy, F-measure and G-means. It can be seen that there is a wide range in terms of the G-means metric, this indicates that many combinations of parameters managed to produce classification biased toward aggressive or normal driving, without being capable of making correct predictions for both classes in this unbalanced data set.

\begin{figure}
	\centering
	\subfigure[]{
		\includegraphics[width=8.5cm]{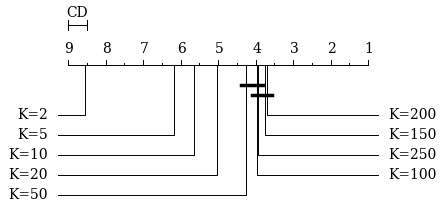}
	}
    \vfill
	\subfigure[]{
		\includegraphics[width=8.5cm]{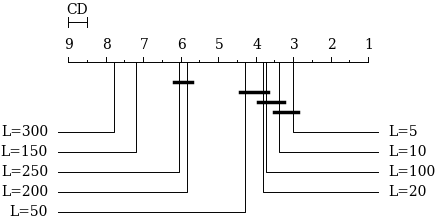}
	}      
	\caption[Critical difference diagrams for the parameters K and L.]{
		a) CDD for F-measure with respect to $K$.
        b) CDD for F-measure with respect to $L$.
	}
    \label{fig:risky:zylius_cdplot_k}
\end{figure}

Statistical testing was performed with the methodology described in \citet{demsar2006} (also used in the previous chapter). Figure \ref{fig:risky:zylius_cdplot_k} presents the critical difference diagrams for the parameters $K$ and $L$, showing that some values for the parameters can be considered equal. From this analysis, the following generalizations were made:

\begin{itemize}
    \item The usage of the norm of the vectors for longitudinal and lateral acceleration yields better results than the other evaluated axis combinations. This was expected since both longitudinal and lateral accelerations tend to be used to detect different types of driving events (braking/cornering, etc.), and is in agreement with the results presented in \citet{eboli2016}. Using only the axis associated to Lateral acceleration is the second best alternative.
	\item Larger codebooks ($K>50$) tend to perform better than smaller ones. 
    \item In general, smaller codeword lengths ($L$) achieve better results. This is related to the duration of the events of interest, and the lengths of the characteristic shapelets discovered.
    \item Overall, MLP was the best option to classify the feature vectors computed with BoW, followed by a GNB, but there's no significant difference between both classifiers in terms of G-means. However, the GNB classifier had a much lower correlation between F-measure and G-means than the MLP (0.76 and 0.96, respectively), therefore a MLP is preferred.
\end{itemize}

\subsection{Classification of Aggressive Driving Events}

The classification of individual aggressive driving events is now addressed, first evaluating the suitability of the bag of words methodology, and then comparing its results with other automatic feature learning methodologies.

\subsubsection{Bag of Words}

The bag of words feature extraction technique presented before was used to identify the specific type of aggressive maneuver reflected in a short acceleration time series in two axis (lateral and longitudinal) over our data set. The proposal of \citet{junior2017} was also applied to vertical acceleration as a benchmark. 

\begin{figure}
	\centering
		\includegraphics[width=6.5cm]{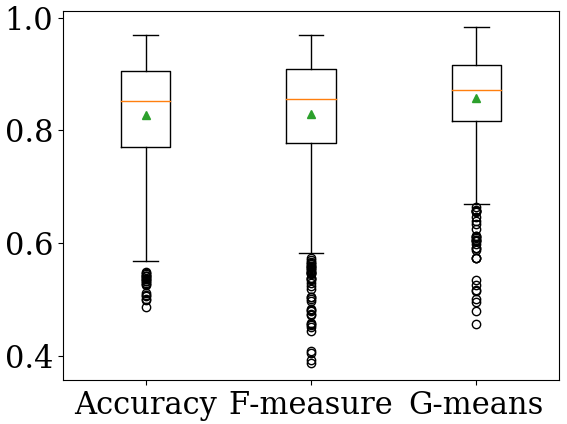}
	\caption[All results for individual aggressive event detection with the bag of words representation.]{
        Box plots summarizing the distribution of the average results for individual maneuver classification with all the parameters in the grid search for our data set.
	}
    \label{fig:ours_all}
\end{figure}

Figure \ref{fig:ours_all} presents the results in terms of accuracy, F-measure, and G-means, for individual aggressive maneuver classification, showing high median values in all three metrics. The best results, and the parameters that produced them are listed in Table \ref{tbl:risky:bow_four}. Automatic feature learning outperformed the feature vector proposed in \citet{junior2017}. 

\begin{table}
	\caption{Best results for aggressive event classification over our data set.}
	\label{tbl:risky:bow_four}
	\centering
	\resizebox{\columnwidth}{!}{%
	\begin{tabular}{cccccccccc}
		\toprule
		Methodology & Classifier & nf& K & L & Overlap & TF-IDF & F-measure & G-means & Acc.\\
		\midrule
        Author's &GNB & - & 150 & 200 & Yes & No  & 0.9702 & 0.9758 & 0.9688 \\
        Author's & MLP & - & 50  & 150 & Yes & No  & 0.9689 & 0.9807 & 0.9683 \\
        Author's & MLP & - & 150 & 200 & Yes & Yes & 0.9689 & 0.9828 & 0.9677 \\
        Author's & MLP & - & 100 & 200 & No  & Yes & 0.9669 & 0.9679 & 0.9661 \\
        Author's & MLP & - & 100 & 150 & Yes & Yes & 0.9665 & 0.9776 & 0.9655 \\
        \cite{junior2017}&  RF & 4 &- &- &- & -& 0.8738 & 0.9244 & 0.8725 \\        
        \cite{junior2017}& RF & 3 &- &- &- & -& 0.8370 & 0.8949 & 0.8351 \\
        \cite{junior2017}&RF & 2 &- &- &- & -& 0.6560 & 0.7701 & 0.6586 \\
        \cite{junior2017}& RF & 1 &- &- &- & -& 0.4662 & 0.6099 & 0.4781 \\
       
        \bottomrule        
	\end{tabular}
    }%
\end{table}

Both data representations were also tested in the data set released by \citet{junior2017}. Table \ref{tbl:risky:bowjuniors} presents the best performing combinations of parameters for the bag of words model, along with the results with statistical summarization. Again, the bag of words model has better results in terms of the three metrics.

\begin{table}
	\caption[Best results for aggressive event classification over the data set by Ferreira et al.]{Best results for aggressive event classification over the data set by \cite{junior2017}.}
	\label{tbl:risky:bowjuniors}
	\centering
	\resizebox{\columnwidth}{!}{%
	\begin{tabular}{cccccccccc}
		\toprule
		 Methodology & Classifier &nf& K & L & Overlapping & TF-IDF & F-measure & G-means & Acc.\\
		\midrule
        Authors' & RF &- & 10 & 5   & Yes & No  & 0.8987 & 0.9184 & 0.8928 \\
        Authors' & MLP &-& 5  & 100 & Yes & No  & 0.8806 & 0.8967 & 0.875 \\
        Authors' & RF  &-& 50 & 5   & Yes & Yes & 0.8787 & 0.9148 & 0.8785 \\
        Authors' & RF  &-& 10 & 150 & Yes & Yes & 0.8736 & 0.8960 & 0.8714 \\
        Authors' & MLP &-& 50 & 10  & Yes & Yes & 0.8601 & 0.9266 & 0.8607 \\
        \cite{junior2017} &  RF & 4 & - & - & -&-& 0.6862 & 0.8344 & 0.7013 \\        
        \cite{junior2017}&RF & 3 & - & - & -&-& 0.6381 & 0.8044 & 0.6473 \\
        \cite{junior2017}&RF & 2 & - & - & -&-& 0.4938 & 0.7181 & 0.5200 \\
        \cite{junior2017}&RF & 1 & - & - & -&-& 0.4564 & 0.6056 & 0.4709 \\
        \bottomrule        
	\end{tabular}
	}%
\end{table}

\subsubsection{SAX and Bag of Features}

Given the effectiveness of the bag of words model, two different automatic feature extraction alternatives were evaluated: a SAX \citep{senin2013} representation with n-grams, and bag of features \citep{baydogan2013}. Both techniques were applied over our data set and the one by \citet{junior2017}, to compare their suitability for the classification of aggressive driving events. The parameters for the SAX and bag of features models are shown in Tables \ref{tbl:risky:saxparameters} and \ref{tbl:risky:bof_params}. The best results obtained from the evaluation of the different combinations of parameters are summarized in Table \ref{tbl:risky:best_multi}.

\begin{table}
	\caption{Parameters considered for SAX-based classification.}
	\label{tbl:risky:saxparameters}
	\centering
	\begin{tabular}{cc}
		\toprule
		Parameter & Evaluated values\\
		\midrule
        Classifiers & MLP, RF, GNB, KNN\\        
        Word length & $L = \{ 25, 50, 100, 150, 200 \}$ \\        
        Alphabet length & $S = \{ 5, 10, 15, 20 \}$ \\        
        N-gram max. size & $N = \{ 1, 2, 3, 4, 5 \}$ \\
        TF-IDF  & Yes/No \\
        \bottomrule        
	\end{tabular}
\end{table}

\begin{table}
	\caption{Parameters considered for BoF-based classification.}
	\label{tbl:risky:bof_params}
	\centering
	\begin{tabular}{cc}
		\toprule
		Parameter & Evaluated value\\
		\midrule
        Classifiers & MLP, RF, GNB, KNN\\        
        Subsequence length factor  & $z = \{ 0.1, 0.25, 0.5, 0.75 \}$ \\   
        Histogram bins & $B = \{ 5, 10 \}$ \\
        Subsequence generation scheme & Totally random \\
        TF-IDF  & Yes/No \\
        \bottomrule        
	\end{tabular}
\end{table}

\begin{table}
  \caption{Comparison of the best results achieved with each feature extraction methods for both data sets.}
  \label{tbl:risky:best_multi}
  \centering  
  \begin{tabular}{ccccccc}
  	    \toprule
      & \multicolumn{3}{c}{Author's data set} & \multicolumn{3}{c}{Ferreira et al.'s data set} \\
		\midrule      
      & F-measure & G-means & \textcolor{black}{Accuracy} & F-measure & G-means & \textcolor{black}{Accuracy} \\
        \midrule            
  BoW & 0.9689 & 0.9807 & 0.9683 & 0.8987 & 0.9184  & 0.8928    \\
  BoF & 0.9581 & 0.9780 & 0.9566 & 0.7928 & 0.7162  & 0.7857    \\
  SAX & 0.9217 & 0.9325 & 0.9205 & 0.7395 & 0.8567  & 0.7357    \\
        \bottomrule    
  \end{tabular}
\end{table}

Although the three methodologies have different underlying principles, they all allow for very good classification results. The bag of words model achieves the highest scores, closely followed by bag of features. All three models have better performance when longer sequences of samples are considered for feature extraction, this is probably related to the length of the events in question and their low-frequency components, for which longer shapelets are more representative. 

\section{Contributions}

Two related driving style analysis problems were addressed by using automatic feature extraction techniques: aggressive driving detection, and the classification of common types of individual aggressive maneuvers. 

For the first problem, a grid search of parameters was applied to a bag of words model for time series. It was found that many configurations are capable of performing as well or better than the state of the art proposal by \citet{zylius2017}, based on hand-crafted features, achieving accuracy scores equal or greater than 95.5\%.

Automatic feature extraction  also managed to outperform the state of the art strategy by \citet{junior2017}, based on statistical summarization over sliding windows. Feature vectors produced by the bag of words methodology outperformed the hand-crafted descriptors by over 5\% in accuracy, F-measure, and G-means, on two different data sets.

In total, three different methods were evaluated for the automatic extraction of features and classification of individual aggressive driving events: the bag of words model, bag of features, and a SAX-based representation. These  techniques had not been  used in the context of driving style analysis but, as it was shown, are useful to produce highly classifiable representations without much direct human intervention. After the bag of words model, bag of features showed to be the second best alternative. The bag of features model requires the least adjustment in its parameters, and the default configuration in the original implementation manages to achieve very good results.

%% file: chapters/conclusions.tex
\chapter{In Conclusion: a General Pipeline for Smartphone-based Sensing of Roads and Driving Style
}

\label{conclusions}

\section{Motivation}

This study has addressed road anomalies and aggressive driving, two important problems in ITS which are very related because of their almost simultaneous occurrence in real life, the use of the same sensors for data collection, and the application of very similar algorithms and techniques for their study and the creation of solutions. Given these similarities, it is natural to think that both problems can, and perhaps should, be treated simultaneously, or that at least they would be highly intertwined in the literature. Surprisingly, that has not been the case.

Although researchers some times work on both problems, it is uncommon to do so in the same article\footnote{\citet{mohan2008} is a very important seminal work because of the exploration of multiple problems in ITS that have to be simultaneously addressed if robust systems are to be deployed in real life.}, and even less common to find an experiment that simultaneously treats road anomalies and aggressive driving\footnote{\citet{tecimer2015} presented a classification model that simultaneously deals with  road anomalies and aggressive driving events.}. This notorious division should be ended, so that the developments and experiences in one topic can more quickly benefit the work in the other. 

In this final chapter, the proposals found in the rest of the document are combined to create a pipeline for both road condition and driving style sensing and classification. This pipeline is based on machine learning techniques, does not assume knowledge about the characteristics of the vehicle, and does not rely on threshold heuristics. 

\section{Aims and Objectives}

The objective of this study is to produce a pipeline that can perform road anomaly detection, classification, and characterization, alongside  aggressive driving detection and the classification of individual aggressive maneuvers, presenting a unified framework for acceleration time series analysis for ITS applications.

As a secondary objective, a feature vector is proposed for the detection of events of interest in both road and driving analysis, becoming the main routing element for the proposed pipeline. This feature vector includes statistical descriptors calculated over triaxial acceleration readings and the pitch and roll angles.

\section{Our Proposal}

\begin{figure}[!ht]
    \centering
    \includegraphics[width=14cm]{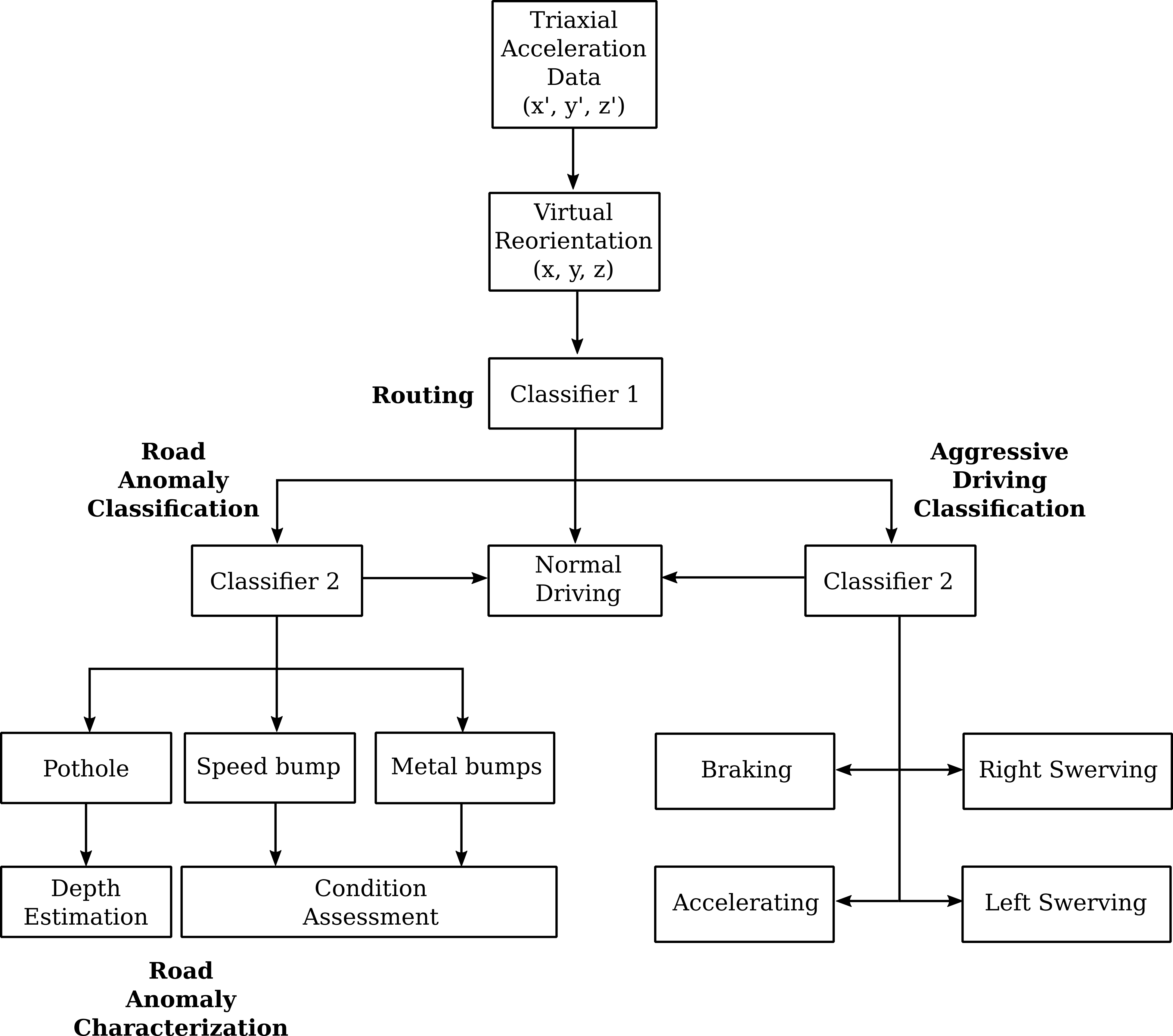}
    \caption[Diagram of the proposed pipeline.]{Diagram of the proposed pipeline. It takes as input acceleration in three dimensions, which might need to be reoriented. Routing is the first step, in which a classifier decides if the acceleration readings reflect normal driving (in which case no further action is required), a road anomaly, or an aggressive driving event. After this decision is made, the tasks of road anomaly classification, and characterization, or aggressive driving event classification are performed, accordingly.}  
    \label{fig:conclusion:pipeline}   
\end{figure}

A general overview of the pipeline is presented in Figure \ref{fig:conclusion:pipeline}. It starts with the reorientation of the acceleration signals addressed in Chapter \ref{reorientation}, if needed, which is followed by the calculation of a feature vector and the classification of signals in three classes: normal driving (for which no further action is needed), road anomaly, and aggressive driving event. For the case of road anomalies, the procedure described in \citet{gonzalez2017} is followed, in order to determine the type or road anomaly found. That process can be optionally followed by the estimation of depth, for potholes, and functional status, for speed bumps, as described in Chapter \ref{characterization}. Aggressive driving maneuvers are classified with the methods described in Chapter \ref{risky}. Both the road anomalies and the aggressive maneuver classifiers also manage the normal driving category, as to be able to handle those samples erroneously sent by the first classification.

The new component presented in this chapter is the feature vector used in the first classification, for the determination of which of the three main paths has to be followed for a specific signal. It combines the insights gained from the analysis of the literature and the experience acquired while modeling acceleration signals. 

This feature vector can be calculated over time series of variable length (from one to five seconds, the length of most of the events of interest addressed in this dissertation). It uses some statistical features calculated over the complete time series, but also divides it in subsequences with length of one second (with 50\% overlapping), and considers the maximum value calculated over each of these subsequences. Features are calculated from three time series: vertical acceleration (Z), the Euclidean norm of longitudinal and lateral acceleration (H), and the Euclidean norm of pitch and roll angles, calculated with Equation \ref{eq:intro:a_b_linear} or an equivalent (A). The first derivative (deriv) of vertical acceleration and the Euclidean norm of the other two axes are also used to calculate features. The complete feature vector is described in Table \ref{tbl:conclusion:features}, indicating the descriptor, time series for which it is calculated, and if it the global value for the time series or the maximum value of all its subsequences.

\begin{table}
	\caption{Features used for the separation of normal driving, road anomaly, and aggressive driving.}
	\label{tbl:conclusion:features}
	\centering
	\begin{tabular}{lccc}
		\toprule
		\textbf{Descriptor} & \textbf{Time series} & \textbf{Global value} & \textbf{Max. from subsequences} \\
		\midrule
        Stdev & A & X & \\
        Stdev & Z & X & \\
        Stdev & Z &  & X\\
        Stdev & deriv(Z) &  & X\\
        Crossing rate & Z & X & \\
        Mean & H & X & \\
        Max & deriv(H) & X & \\
        Stdev & H &  &X \\
        \bottomrule        
	\end{tabular}
\end{table}

Using the maximum value of the one overlapping second long subsequences attempts to focus on the most probable segment in which most of the energy of the event of interest is present. The descriptors obtained from the complete time series attempt to provide a more ample context, giving the classifier a reference of the general conditions of the road and the driving maneuvers.

\section{Preliminary Results}

The pipeline was evaluated over a data set with 1,477 time series representing seven classes: normal driving, aggressive acceleration, aggressive braking, aggressive swerving, potholes, speed bumps, and metal bumps. The distribution of the time series for all events contained in the data set is detailed in Table \ref{tbl:conclusion:dataset}.

\begin{table}
	\caption{Number of examples for each category in the data set used to evaluate the proposed pipeline}
	\label{tbl:conclusion:dataset}
	\centering
	\begin{tabular}{lc}
		\toprule
		\textbf{Class} & \textbf{Examples}\\
		\midrule
		Normal driving & 462 \\
        Aggressive acceleration & 146 \\
        Aggressive braking & 133 \\
        Aggressive swerving & 286 \\
        Potholes & 150 \\
        Speed bumps & 150 \\
        Metal bumps & 150 \\
        & 1,477 \\
        \bottomrule        
	\end{tabular}
\end{table}

A SVM machine with RBF kernel was used to classify the feature vector described in this chapter. The evaluation for the complete pipeline, and the individual classifiers, was performed using a stratified shuffle split cross-validation scheme with a 80\% of data for training and 20\% for testing, repeated 30 times.

The average F-score for the routing classifier using the feature vector proposed above was 0.94. The average F-score for all categories, after using the classifiers for road anomalies and aggressive maneuvers, was 0.87, with most of the classes having values over 0.9 but being lowered by the mistakes made in the classification of road anomalies in some batches. 

\section{A Take-home Message of this Dissertation}

Several remarks can be made as general recommendations and conclusions from the results presented in this dissertation, and the experience gained while working on the experiments described above. These are meant to simplify the development of new systems that apply this knowledge, and to save time to new researchers who wish to work on the problems addressed in this text.

The first observation is that a few minutes of data should be considered for sensor reorientation. There is a higher probability of having a smartphone changing its orientation during longer periods of time, but if there is no indication of user interaction with the device, these longer periods will produce better results. The extraction of acceleration readings in the vertical axis is almost always possible, and the separation of the two other axes is simple once vertical reorientation has been performed. However, individually identifying lateral and longitudinal components is more difficult, and the use of algorithms that rely on their Euclidean norm can both simplify the process and improve the results obtained.

Although data collection is an expensive process for ITS research, it is very important to try to assemble a robust and diverse data set. Even if dedicated sensors are installed in vehicles, there are many sources of noise and unforeseeable changes that complicate the analysis of the signals. Data augmentation techniques should be explored to address this situation. The use of semi-supervised learning can be another worthy pursuit, to apply the available techniques and data to tag signals obtained by means of crowdsourcing and thus increase the size of the data sets for future research. 

The reliance on threshold-based algorithms should be avoided, except when a computationally inexpensive basic detector for a whole family of patterns is desired. As it has been showed, machine learning detectors are capable of outperforming them, and alternatives such as the first classifier presented in this chapter can be more versatile.  

Hand-crafted feature vectors, based on time and frequency domain descriptors, are useful for the identification of anomalies in time series, but they do not perform as well for the classification of more specific patterns. The information required for more fine-grained classification is easier to detect by means of the comparison of sequences, for which some level of temporal information has to be preserved. Most of the descriptors used in the feature vectors described in the literature discard this temporal information and present an atemporal summary of a sequence of readings. Techniques such as those presented in Chapter \ref{risky}, or others capable of grouping individual readings and treat them as basic elements (such as those relying on shapelets, or convolutional neural networks) should be tried for high level pattern classification, to save time (since manual feature engineering is more time consuming) and achieve better performance. 

Smartphones are a simple, widely deployed, and inexpensive sensing platform for ITS, both as an opportunistic sensing alternative for crowdsourcing campaigns and as dedicated sensors for controlled experiments. Their ease of use and low cost have truly removed many of the barriers to entry for many research topics in ITS, smart cities, civionics, civil engineering, and related areas. The challenge now resides in the software components of the computer systems, making sense out of the great volumes of data that are starting to become available, to create more advanced systems capable of dealing with higher-level problems, and having them deployed at large scales.